# Context in object detection: a systematic literature review

Mahtab Jamali[1] · Paul Davidsson[1] · Reza Khoshkangini[1] · Martin Georg Ljungqvist[2] · Radu-Casian Mihailescu[1]

## Abstract

Context is an important factor in computer vision as it offers valuable information to clarify and analyze visual data. Utilizing the contextual information inherent in an image or a video can improve the precision and effectiveness of object detectors. For example, where recognizing an isolated object might be challenging, context information can improve comprehension of the scene. This study explores the impact of various context-based approaches to object detection. Initially, we investigate the role of context in object detection and survey it from several perspectives. We then review and discuss the most recent context-based object detection approaches and compare them. Finally, we conclude by addressing research questions and identifying gaps for further studies. More than 265 publications are included in this survey, covering different aspects of context in different categories of object detection, including general object detection, video object detection, small object detection, camouflaged object detection, zero-shot, one-shot, and few-shot object detection. This literature review presents a comprehensive overview of the latest advancements in context-based object detection, providing valuable contributions such as a thorough understanding of contextual information and effective methods for integrating various context types into object detection, thus benefiting researchers.

**Keywords** Context · Contextual information · Object detection · Object recognition · Computer vision

## 1 Introduction

Object detection is a fundamental computer vision task to identify and locate objects within images or videos. It is a foundation in other computer vision tasks, such as object tracking, scene understanding, and image captioning. Object detectors can be classified into two distinct categories: traditional object detectors, and deep learning object detectors, which have emerged since 2012. The traditional methods have limits in terms of robustness and speed when dealing with large datasets. The introduction of Convolutional Neural Networks (CNNs) by AlexNet (Krizhevsky et al. 2012) in 2012, sparked a profound revolution in the

---

Extended author information available on the last page of the article field of object detection. The timeline of some of the most important object detectors is depicted in Fig. 1, illustrating the historical development of these methods over time.

Despite significant progress in object detection, finding all objects in visual scenes remains a challenging topic for object detectors due to a multitude of factors. Some of the factors are as follows:

- Inter-class similarity and intra-class variations: Inter-class similarity is high when objects from two different classes are extremely similar to one another; intra-class variations are high when the exterior perspectives of something, such as a school, can vary so drastically across various images (Venkataramanan et al. 2021). Both have the potential to impair the networks' ability to comprehend the scenarios accurately.
- Adverse environmental or imaging conditions: Fluctuations in images caused by occlusion, blur, weather and lighting conditions, small objects, deformation, and variations in object orientation are additional obstacles for detecting objects.
- Objects out of context: Real entities have a tendency to appear in a spatial arrangement that facilitates their identification and localization. However, this can complicate object detection when objects are presented out of the correct context.

Several of these challenges are depicted in Fig. 2.

To address the aforementioned challenges, using context is one of the effective approaches that significantly enhances the accuracy and robustness of object detectors. Object detection is more accurate when contextual information is taken into account (Shrivastava and Gupta 2016; Gong et al. 2019). Context refers to any information that can be used in accurate semantic understanding of a scene and recognition of its element (Zolghadr and Furht 2016). Contextual information encompasses a wide range of information, including environmental information, lighting conditions, objects' position and orientation, relationships

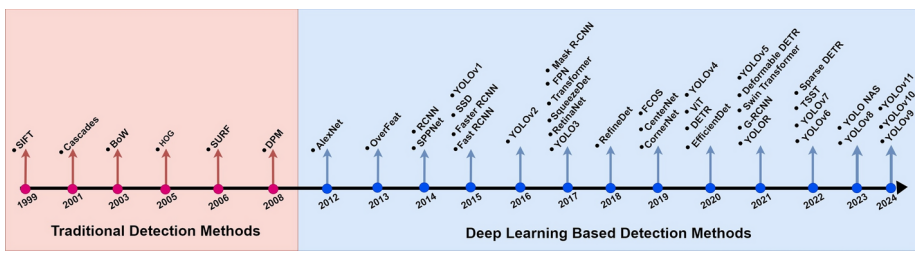

**Fig. 1** Milestones of object detection, SIFT (Lowe 1999), Cascades (Viola and Jones 2001), BoW (Sivic and Zisserman 2003), HOG (Dalal and Triggs 2005), SURF (Bay et al. 2006), DPM (Felzenszwalb et al. 2008), AlexNet (Krizhevsky et al. 2012), OverFeat (Sermanet et al. 2013), RCNN (Girshick et al. 2014), SPPNet (Kaiming et al. 2014), Fast RCNN (Girshick 2015), Faster RCNN (Ren et al. 2015), YOLOv1 (Redmon et al. 2016), SSD (Liu et al. 2016), YOLOv2 (Redmon and Farhadi 2017), YOLOv3 (Redmon and Farhadi 2018), Mask R-CNN (He et al. 2017), FPN (Lin, Dollár, et al. 2017), RetinaNet (Lin, Goyal, et al. 2017), SqueezeDet (Wu et al. 2017), Transformer (Vaswani et al. 2017), RefineDet (Zhang et al. 2018), CornerNet (Law and Deng 2018), CenterNet (Duan et al. 2019), FCOS (Tian et al. 2019), EfficientDet (Tan et al. 2020), DETR (Carion et al. 2020), ViT (Dosovitskiy et al. 2020), YOLOv4 (Bochkovskiy et al. 2020), YOLOv5 (Jocher 2020),YOLOR (Wang et al. 2021), G-RCNN (Wang and Hu 2021), Swin Transformer (Liu et al. 2021), Deformable DETR (Zhu et al. 2020), YOLOv6 (Li et al. 2022), YOLOv7 (Wang et al. 2023), TSST (Lee 2022), Sparse DETR (Roh et al. 2021), YOLOv8 (Jocher et al. 2023), YOLO NAS (Aharon et al. 2021), YOLOv9 (Wang and Liao 2024), YOLOv10 (Ao Wang 2024), YOLOv11 (Jocher and Qiu 2024)

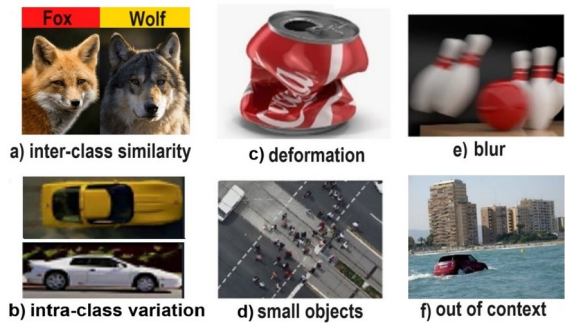

**Fig. 2** Object detection challenges, **a**, **b** inter-class similarity, intra-class variation, **c**–**e** imaging condition, **f** object out of context

between objects, time, location, and other visual or non-visual information that can provide additional information for object detectors. Non-context-based approaches perform object detection without considering contextual information in scenes. They mostly use low-level visual features such as color, texture, and shape to detect and locate objects. However, such appearance features often do not provide sufficient information to identify objects when environmental conditions change. In complex and challenging scenes, context-based approaches can provide more accurate results when dealing with adverse conditions. Context enables object detectors to comprehend their visual inputs by supplying information about their environments. A wide range of domains, from autonomous vehicles (Ibañez-Guzman et al. 2012), surveillance systems (Nazir et al. 2022), robotics (Dimitropoulos and Hatzilygeroudis 2022), medical image processing (Girum et al. 2021), benefit from context in terms of having better performance in detecting objects.

Based on Fig. 29, due to the increasing number of publications exploring context in object detection, it is important to determine how to make use of context to enhance performance and address the mentioned challenges. We argue that a comprehensive analysis of context in various types of object detection is desirable and can bring significant insights to the research community. To this end, we have conducted a comprehensive and systematic investigation of recent context-based object detection methods, including a detailed analysis and comparison. Compared to the existing surveys ((Galleguillos and Belongie 2010), (Marques et al. 2011), (Liu et al. 2020), (Gong et al. 2019)), our literature review is systematic and covers the recent developments in different categories of context-based object detection, including general object detection (GOD), video object detection (VOD), small object detection (SOD), zero-shot object detection (ZSD), one-shot object detection (OSOD), few-shot object detection (FSOD), and camouflaged object detection (COD). Galleguillos and Belongie (Galleguillos and Belongie 2010) focused on the role of context in object categorization. Marques et al. (Marques et al. 2011) discussed the importance and different types of context. In one section of another survey, Liu et al. (Liu et al. 2020) mentioned several papers that employed local and global context to detect objects (without specifying the type of context utilized). Recently, Wanga and Zhua (Gong et al. 2019), published a survey about applications of context in computer vision. Unlike these previous surveys that narrowly focus on certain aspects of context, such as local or global considerations, or limit their examination to general object detection, our analysis encompasses context from various perspectives across seven categories of object detection. Moreover, one of the goals of this study is to explore the use of context in recognizing objects in unfavorable scenarios, such as small object detection, zero-shot object detection, one-shot object detection, and few-shot object detection that have not yet been explored, and no literature review has been published on these topics.

After thoroughly reviewing the available literature and identifying the gaps therein, we address the following research questions in this literature review.

- RQ1. Which context types have been predominantly used in different categories of object detection?
- RQ2. What approaches are applicable for integrating context in object detection?
- RQ3. Why are certain backbone networks and architectures most commonly used in recent context-based object detectors?
- RQ4. What are the best performing context-based methods on the most widely used datasets, including COCO and PASCAL VOC? What about for one-stage and two-stage object detectors?
- RQ5. To what extent can context improve object detection in scenarios where the number of training samples is very limited, such as in few-shot object detection, or when objects are indistinguishable from the background, as in camouflage object detection?

To answer these research questions and provide a comprehensive view of context, we reviewed more than 240 papers and conducted a thorough literature review. The main contributions of this work can be summarized as follows:

- A comprehensive review of context from different perspectives, including context in human vision and computer vision, pairwise and higher-order relations, context levels: prior knowledge - global and local, contextual interactions in global and local considerations, and different types of context such as spatial, scale, temporal, spectral, thermal, etc.
- An analysis of the recent context-based object detection approaches in seven categories, including general object detection, video object detection, small object detection, zero-shot object detection, one-shot object detection, few-shot object detection, and camouflaged object detection. All approaches are investigated based on context type, context level, backbone and architecture, mechanism and module, dataset, and mean average precision (mAP), and other evaluation metrics.
- Highlighting the problems addressed in object detection that can be addressed by the utilization of context.
- Identifying research gaps for future studies.

The structure of this literature review is organized as follows. In Sect. 2, general concept and various definitions of context are investigated. The importance of context in computer vision and human vision is described in Sect. 2.1. Section 2.2 covers different levels of context, as well as contextual interactions. Higher-order and pairwise relations are subsequently examined in Sect. 2.3. Section 2.4 gives a comprehensive analysis of different types of contextual information. Section 2.4 covers research method, including bibliographical databases 3.1, selection of studies 3.2, inclusion and exclusion criteria 3.3, data extraction and validity control 3.4, classification of the papers 3.5, context in categories of object detection 3.6, and data extraction and synthesis 3.7. In Sect. 4, employed datasets 4.1, papers in seven categories, including general object detection 4.2, small object detection 4.3, video object detection 4.4, zero-shot, one-shot, few-shot object detection 4.5, and camouflaged object detection 4.6, are reviewed, and key points, as well as the best models, are discussed. Finally, Sect. 5 provides a conclusion by addressing research questions and identifying research gaps.

## 2 Context

Different definitions of context have been proposed in different papers. Brown et al. (1997) describe context as the locations and identities of objects in different scenes, the time of day, season, temperature, etc. Galleguillos and Belongie (2010) define context in computer vision as available information beyond the local visual features of an image or video, which can help in resolving ambiguities and enhancing the accuracy of recognition and detection tasks. Felzenszwalb and Huttenlocher (2005) define context as a set of objects, scenes, and events that are relevant to a specific task that can be used to guide visual attention, reasoning, and decision-making. In general, context or contextual information can be any visual or non-visual information, including appearance information, such as color and texture; location information consisting of a kitchen or a library; time-related details, like the precise hour of a day or a month; semantic information, including the relationship between objects or different regions of a visual scene, and any other information that aids in better understanding the environment (Wang and Zhu 2023). Based on Fig. 6, context in computer vision refers to the information inside and surrounding an object or region of interest, providing essential cues for understanding its identity, location, and function. In order to have a deeper comprehension of context, we have thoroughly investigated and categorized it from different aspects, as shown in Fig. 3. This categorization is not merely our opinion but is derived from the extensive analysis of existing research.

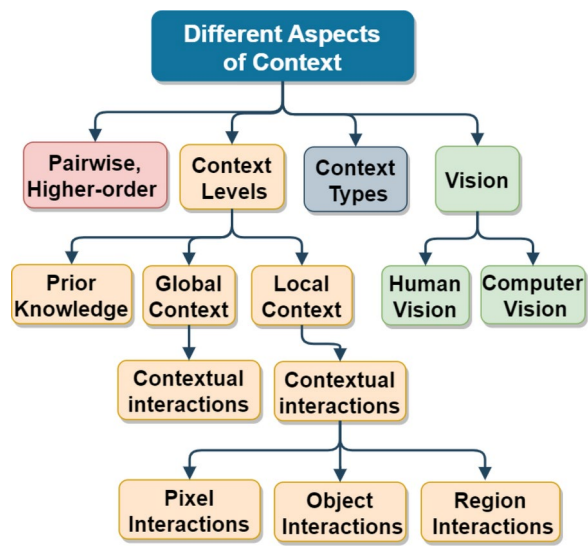

**Fig. 3** Context from different aspects

## 2.1 The role of context in human vision and computer vision

The notion of context in visual situations has been explored for many years, but it is unclear who originally suggested it. However, the origins of some of the earliest research on visual perception can go back to Gestalt psychology in the early 20th century, which theorized that the human's brain groups all environmental information into meaningful structures and patterns based upon the context, such as proximity, similarity, closure, continuation, figure, and ground (Wertheimer 2017). According to Gestalt psychology, humans, animals, and other organisms understand complete patterns or combinations, not just isolated objects, regions, or components. This striking ability helps humans to comprehend visual scenes quickly and intuitively. This process is called human vision. The human visual system is extraordinarily adept at detecting, categorizing, and naming objects embedded in natural scenes (Munneke et al. 2013). Complex visual information, such as objects, layouts, scenes, actions, etc., can be seamlessly incorporated by our brains into coherent perceptions of the world. Despite the constant changes in environments, obstacles, blur, changes in weather and light, etc., humans are still able to understand the environments by using contextual information. Through context-based understanding, humans are able to recognize objects faster and more accurately, as well as make better decisions, even if they are incomplete or ambiguous. For instance, if humans see a partially obscured object in a scene, surrounding context, such as other objects, the background, and the relationship between objects can help their brains to infer what the object might be. Based on Fig. 4, identifying a keyboard in isolation is challenging, especially when the quality of the image is low. In an office with a monitor and a mouse on the desk, it is probable that the item positioned in front of the monitor is a keyboard.

There has been a noticeable increase in evidence suggesting that object recognition does not happen in isolation (Oliva and Torralba 2007; Marques et al. 2011). It is influenced by the presence of other objects as well as by the overall context of the scene. Real-world objects co-occur with other objects and particular environments, which allows visual systems to extract rich contextual clues (Perko and Leonardis 2010). Computer vision tries to imitate human vision to use contextual information for a better understanding of environments. The idea of utilizing context to help computers recognize patterns, objects, and scenes goes back to the 1970 s and 1980 s. Marr and Nishihara (Marr and Nishihara 1978) investigated the use of contextual information to segment images (Marr and Nishihara 1978), and Ballard and Brown (Dana and Christopher 1982) used context in object recognition. Later, (Oliva and Torralba 2007) suggested that visual systems can exploit contextual associations between objects and environments to guide attention and eyes to regions of interest in natural scenes. Overall, the studies suggest incorporating contextual information can lead to more effective computer vision systems. For example, an object can have various applications in different scenes: a tree can exist in a jungle, serve as a symbol in a book, or be used as an artificial decoration in a room. Therefore, it is imperative for a machine to identify the relationships between objects, their co-occurrence, and consider additional information to differentiate objects and accurately comprehend the entire scene. Furthermore, context can decrease the search space in computer vision tasks (Jain and Sinha 2010). When a machine tries to detect objects in an image or video, it has to look through several options or "search space" to figure out what it is seeing. For each object, a machine needs to search over all possible locations and scales and this process can take a lot of time and computing power. Given knowledge of the appearance and potential location of an object, the machine can reduce the search space, ignore heaps of objects that are unlikely to be what it is looking for, and concentrate on the regions where the object is expected to be located. This makes the process faster and more accurate. Additionally, where local features, such as edges or corners, are not sufficient for detecting objects due to different conditions, such as occlusion or rotation, contextual information can improve the efficiency and accuracy of systems. As shown in Fig. 5, it is impossible to detect the wheel of a car based solely on local features. However, by considering other parts of the car and the overall scene, it becomes possible to identify the wheel. In general, by using contextual information, computer vision systems can better identify objects and reduce the chances of false positives or false negatives.

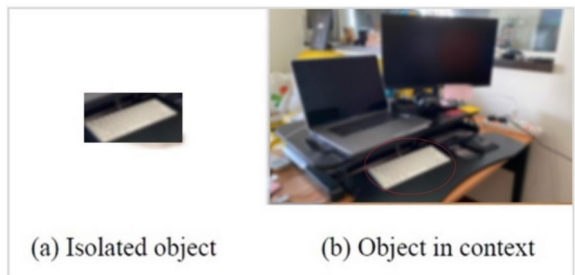

**Fig. 4** Differences between an object in isolation and the same object in context

## 2.2 Context levels

Context can be classified into three main levels: prior knowledge, local context, and global context (Galleguillos and Belongie 2010).

The prior knowledge pertains to the information that can be obtained prior to observing environments. It provides insights into environmental variables, such as time, weather, location, and action, that can be utilized to forecast the occurrence of specific events or the visibility of particular objects. For example, if we know that the location of an image is a city street, the likelihood of encountering a lion is very low.

Local context refers to the information about the object itself and available information in the neighborhood of a pixel or an area. For example, in Fig. 6, for detecting the monitor, various elements can serve as local context, including nearby objects like the keyboard and cup, the overall appearance of the monitor, its different parts, and the surrounding pixels. The local context within bounding boxes around the objects can aid in distinguishing objects that may have significant visual and structural similarities. (Gidaris and Komodakis 2015) and (Zagoruyko et al. 2016) are examples of using local context in object detection.

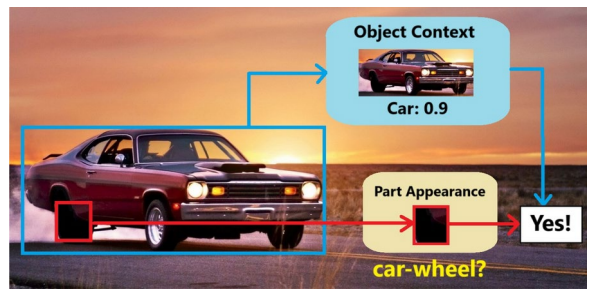

Fig. 5 Utilizing context to better comprehend an object's components

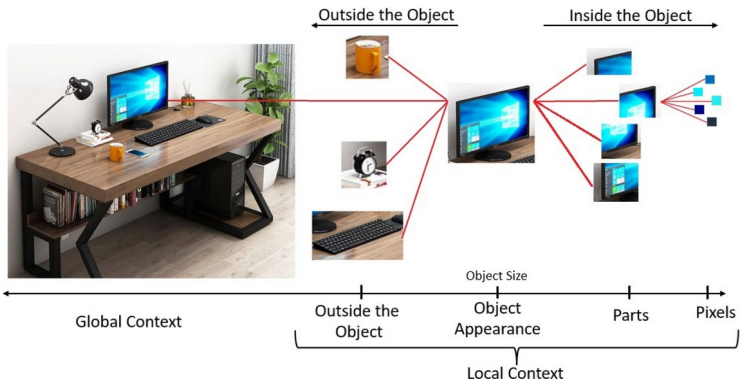

**Fig. 6** Global and local context

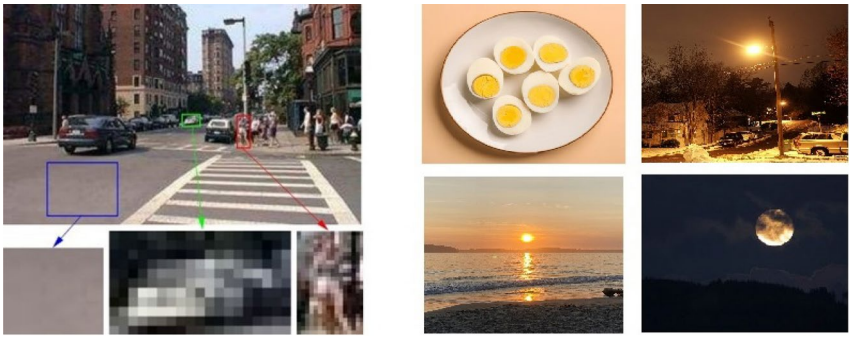

**Fig. 7** Local context is sometimes insufficient for object recognition, as it necessitates the interpretation of the full image. On the left side, the vehicle, the person, and the road are nearly undetectable individually; however, when viewed collectively, they create a logical visual narrative (Derek 2006). On the right side, there are some examples of the importance of global context in object detection (objects, shapes, colors, and textures are similar). Capturing object characteristics may pose a challenge for feature extractors when merely considering local context

When objects are very small in size, relying on local context alone may not be an efficient method, as detecting one object by considering itself and its surrounding pixels poses challenges. In this situation, global context can provide more valuable information. Global context refers to the information that is available over a wider area, such as the entire image or a larger region within the image. For example, to detect a car within an image, global context can be used to leverage the information in the entire image, such as the presence of other cars, buildings, or roads, which can help to provide additional cues for identifying the object. Moreover, global context can be utilized to provide prior knowledge about the typical spatial arrangement of objects within a scene, which can help to guide the interpretation of visual information and reduce the ambiguity and uncertainty of the data. Gist descriptor (Oliva and Torralba 2001) is one of the global image representations that captures the essential global spatial information about the entire scene rather than local information. Figure 7 shows the importance of global context in detecting objects. It should be noted that using global context alone to detect objects may cause confusion because objects can be detected differently in various environments (e.g., a dog in a park or a bedroom). Since each method has its own advantages and disadvantages, combining local context and global context, as demonstrated in (Li et al. 2016), is a practical method for enhancing object detection.

### 2.2.1 Contextual interactions in local context

Three types of contextual interactions exist in the local context, as shown in Fig. 8: pixel, region, and object.

(1) Pixel Interactions: Interactions at the pixel level are based on the assumption that nearby pixels have identical labels. Pixel-level analysis involves analyzing individual pixels in an image or video. Further application of the data obtained from pixel-level interactions is the delineation of object boundaries, which enables automated object segmentation and enhanced object localization. This is often done using techniques such as edge detection, corner detection, and color histograms. The advantage of pixel-level analysis is that it can provide very fine-grained information about an image, such as the location of edges or the distribution of colors. However, pixel-level analysis can be very computationally expensive. It is computationally more complex and time-consuming to obtain interactions at the pixel level, as the network is required to analyze many combinations of small windows from the image in order to arrive at a consistent result.

(2) Region Interactions: Region interaction can be classified into two categories: interaction between different parts of one object, and interactions between image patches or segments. Interactions between different parts of one object can be used to obtain a complete picture of an object. This is often done using techniques such as segmentation. In interactions between image patches, image partitioning methods are usually used for dividing one image into several patches or segments (Carbonetto et al. 2004). The advantage of region or part-level analysis is that it can provide more detailed information about an image than pixel-level analysis, while still being computationally feasible.

(3) Object Interactions: Object interactions are the most intuitive type of contextual interactions for humans, and have been widely analyzed in cognitive sciences (Bar and Ullman 1996; Biederman et al. 1982). Object-level analysis involves identifying and analyzing individual objects within an image or video. In object interactions, since the number of regions that need to be processed by the network is equal to the number of objects, extracting information is computationally less complex and time-consuming compared to pixel and region interactions.

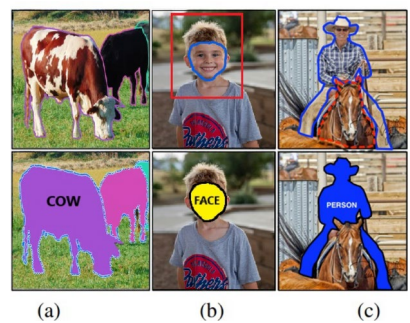

**Fig. 8** Local contextual interactions. **a** Pixel interactions capture information such as grass pixels around the cow's boundary. **b** Region interactions are represented by relations between the face and the upper region of the body. **c** Object relationships capture interactions between the person and the horse

### 2.2.2 Contextual interactions in global context

For incorporating global context a top-down processing is essential for detecting objects. In global context, contextual interactions are analyzed between different objects and the environment. For example, in Fig. 9, the network searches for images with similar scene configurations, as objects typically maintain fixed spatial relationships to each other within a global context. In fact, the background information is used to guide the detection process. Object-scene interactions investigate the possibility of an object in a certain environment to reduce the amount of error.

## 2.3 Pairwise and higher-order relations

Pairwise relations are an essential aspect of spatial context analysis, as they provide valuable information for understanding the spatial relationships between objects. Pairwise relations describe the spatial relationship between two objects or regions of interest (ROIs) within an image. These relationships can be measured using geometric properties such as distance, angle, orientation, overlap, or adjacency between pairs of objects in an image. For example, the distance between two objects can provide information about their proximity and suggest that they belong to the same group or category. The orientation between two objects can provide information about their relative position and suggest that they belong to a specific arrangement or layout. Pairwise orders such as "above", "below", "inside", "around", "left", "right" "next to", "behind" or "overlapping" can help identify objects. For example, the building is "on the left side" of the car. The model presented by Gkioxari et al. (2018) underscores the significance of pairwise relations in a human-centric context by using interactions between humans and objects to enhance object detection. Their model leverages a person's appearance and pose as cues to predict both the location and type of object they are likely interacting with. For instance, if a person is performing an action like "cutting," the model infers that a nearby knife or cutting tool is likely involved, improving detection accuracy by analyzing the spatial relationship between the person and object. This approach highlights how human-object interactions serve as contextual anchors for more precise detection in complex scenes. Higher-order relations refer to the relationships between several objects or ROIs within an image, as shown in Fig. 10. Despite being more complex than pairwise relations, they can provide a more comprehensive understanding of spatial relationships. Pairwise relations and higher-order relations can be modeled using different approaches such as graph-based approaches(Georgousis et al. 2021), tensor-based approaches(Panagakis et al. 2021), and Markov random fields (MRFs)(Blake et al. 2011).

## 2.4 Context types

Depending on the application, different types of context are employed in computer vision. Biederman et al. (1982) suggested five categories of relations between objects and their sur-roundings, including probability (objects tend to be found in some scenes but not others), position (given an object is probable in a scene, it often is found in one position but not others), interposition (object interrupts their background), support (objects tend to rest on surfaces), and familiar size (objects have a limited set of size relations with other objects). Galleguillos and Belongie (Belongie 2007) divided contextual information into 3 main categories: probability (semantic), position (spatial), and size (scale).

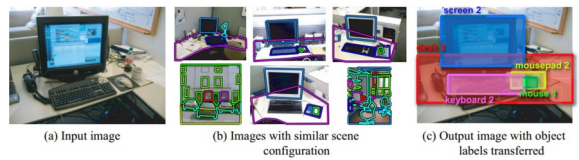

**Fig. 9** Contextual interactions in global context (Russell et al. 2007)

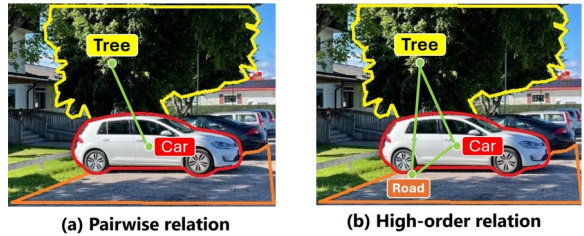

**Fig. 10** Pairwise and high-order relations

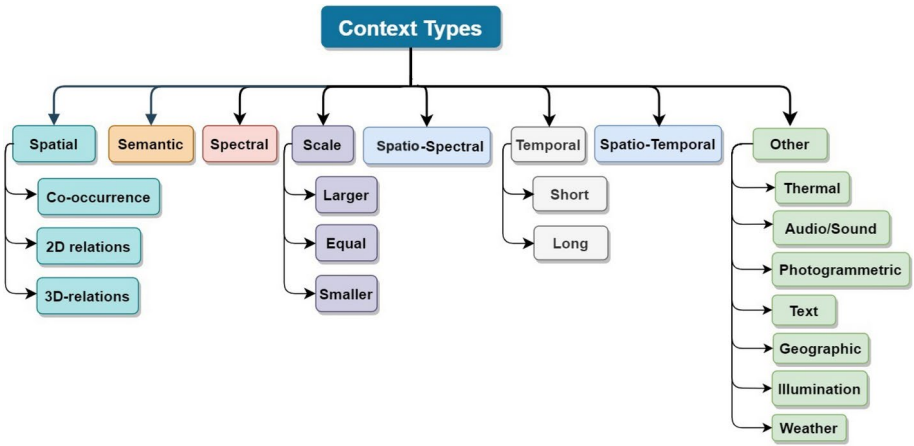

**Fig. 11** The ontology of context types

Furthermore, types and significance of context have been discussed in the literature (Oliva and Torralba 2007; Marques et al. 2011), and recently Wang et al. have categorized context into spatial, temporal, and other types (Wang and Zhu 2023). Despite all these efforts, there is still no comprehensive literature on context types in computer vision. In this section, we provide a comprehensive view of the different types of context that can be used in object detection, illustrated in Fig. 11. Five types of contextual information that are mostly used in computer vision are semantic (probability), spatial (position), temporal (time), scale (size), and spectral. Context types that have received less attention in articles are categorized under the 'Other' classification. Each will be elaborated upon in the subsequent sections.

### 2.4.1 Semantic context (probability)

As shown in Fig. 12, objects are more likely to be seen in some scenes than in others based on semantic context. In fact, the existence of one object increases the probability of the existence of another object appearing in a specific scene. For example, a computer vision system is able to identify a car in an image, but by comprehending the semantic context, it may also infer that the car is most likely on a road, thereby recognizing the scene as a street. Another example is that one sofa in the middle of a jungle does not make sense, but it can be in a living room with a high percentage. A multitude of studies, including (Rabinovich et al. 2007; Katti et al. 2019; Leroy et al. 2020; Ladicky et al. 2010), and (Mensink et al. 2014), have examined and implemented the semantic relationship.

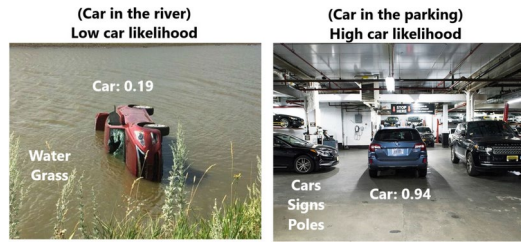

**Fig. 12** Inconsistent and consistent semantic relationships between objects (Katti et al. 2019)

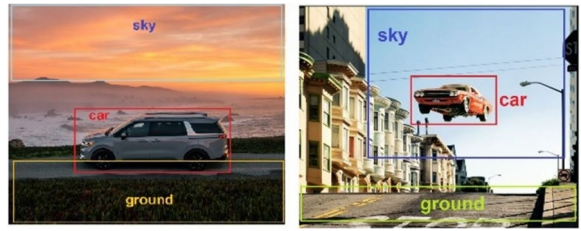

**Fig. 13** Spatial context. Left: car on the road (correct spatial context). Right: car is flying in the sky (incorrect spatial context)

### 2.4.2 Spatial context (position)

Spatial context refers to the physical relationships and arrangements of objects, such as their relative position, orientation, and distance within a specific scene. As a result of the spatial context of a scene, an object is more likely to be found in certain positions than others. In fact, it tries to find reasonable physical properties and relationships between objects and how they interact with each other to understand the structure of the scene. For example, as shown in Fig. 13, a car cannot exist in the sky because it is supposed to be on a road, or if the sky appears above the sand and water, the likelihood of a beach image is strengthened (Singhal et al. 2003).

As depicted in Fig. 14, spatial context can be divided into three main categories: co-occurrence, 2D relations, and 3D relations.

The term 'co-occurrence' is used in both semantic context and spatial context. In the spatial context, co-occurence means focusing on the physical distribution of objects rather than their semantic meaning. Relationships between objects in a visual scene can be introduced simply by co-occurrence (Galleguillos et al. 2008; Rabinovich et al. 2007; Zheng et al. 2011). Spatial context implicitly encodes the physical co-occurrence of objects in an environment (Belongie 2007). In Fig. 4, distinguishing the keyboard without the monitor is difficult because the image is so blurred, but since these two objects are usually seen together, the presence of one object increases the probability of the other object. Co-occurrence analysis identifies patterns or relationships between objects that often occur together in one image. These patterns can be quantified using statistical methods such as co-occurrence matrices, which count the number of times that objects occur together in an image. For example, knives, plates, and a refrigerator are frequently seen in the kitchen, but the bed, closet, and mirror are seen in the bedroom. Co-occurrence matrices can be used as pairwise features in machine learning models like CRFs (Condition Random Fields) or HMMs (Hidden Markov Models) to capture relationships between adjacent observations. For example, Rabinovich (Rabinovich et al. 2007) invented interaction potentials for CRFs to calculate contextual information between different objects. The use of co-occurrence patterns for object detection has been explored by several researchers over the years (Galleguillos and Belongie 2010; Galleguillos et al. 2008; Mensink et al. 2014; Shrivastava and Gupta 2016; Rabinovich et al. 2007).

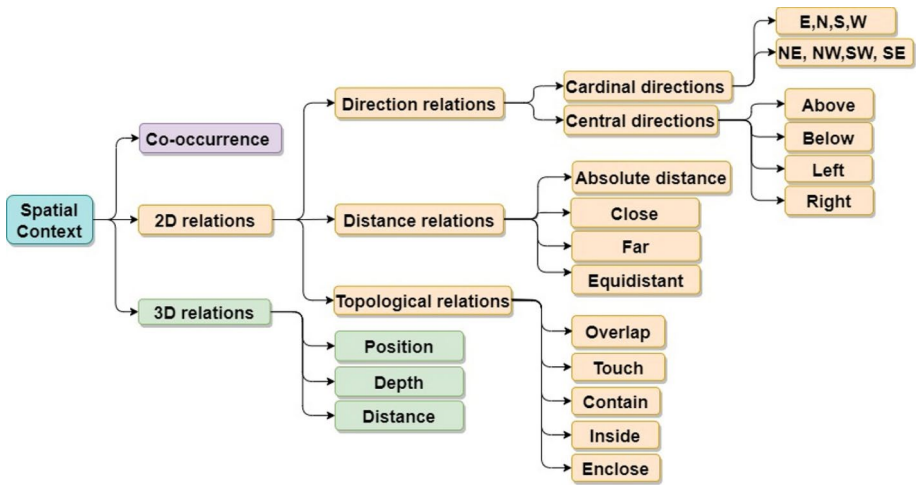

**Fig. 14** The classification of spatial context

The 2D spatial relations are direction relations, distance relations, and topological relations (Marques et al. 2011). In direction relations, objects are oriented in relation to one another. Cardinal directions (E,N,S,W), (NE, NW, SW, SE), and relative vertical positions like "above", "below", horizontal positions like "left", and "right" are samples of direction relations. For example, a book is below a table. Distance relationships are based on the Euclidean distance between two spatial features, as shown in Fig. 15, that measure the distance between objects. Terms like "absolute distance", "close", "far", or "equidistant" are used for distance relations. For example, the car is close to the building. The topological relations describe an object's relationship with its neighbors. In fact, topological relationships describe concepts of adjacency, containment, and intersection between two spatial features (Bogorny et al. 2009). Touch, overlap, contain, inside, and encloses are examples of topological relations. For instance, the car touches the road. Some scholarly works such as (Heitz and Koller 2008; Zheng et al. 2011; Choi et al. 2011) have used this type of spatial context for object detection. 3D spatial relations are another type of spatial context (Chen et al. 2022). In 3D spatial relations, objects are analyzed in three-dimensional space, considering not only their positions but also their depth or distance from the observer. This involves understanding the spatial layout, relative positions, and orientations of objects in a 3D scene. For example, determining if one object is in front of or behind another, or if they are located at different heights or depths within the scene. (Bao et al. 2011; Sun et al. 2010; Pan and Kanade 2013; Hoiem et al. 2005) are some papers that have employed 3D spatial relations in object detection.

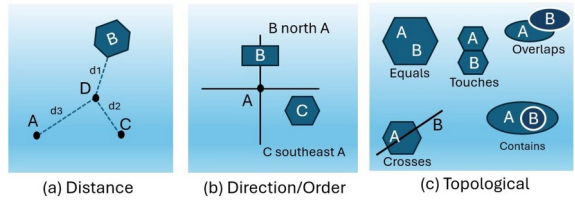

**Fig. 15** Distance, direction/order, and topological 2D spatial relations between objects

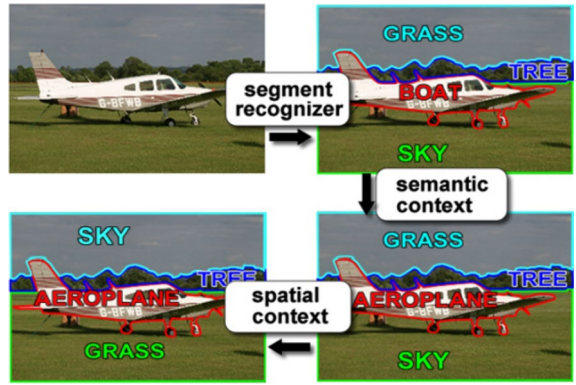

**Fig. 16** Integration of semantic and spatial context. Initially, the input is segmented, resulting in the labeling of each segment. Furthermore, semantic context is employed to rectify some labels by relying on object co-occurrence. Ultimately, spatial context is used to offer more clarification by considering the relative locations of objects (Rabinovich et al. 2007)

It is feasible to create combinations of all these spatial arrangements. For instance, in things and stuff model (TAS) (Heitz and Koller 2008), Geremy Heitz et al. combined all categories of relations (including eight directional relations, two different distances, and a topological relation) in order to generate a large number of candidate relationships. Some methods combine spatial context and semantic context to enhance the accuracy of a network. Figure 16 depicts the integration of semantic content and spatial content inside a network.

### 2.4.3 Scale context (size)

Scale context, originally defined by Biderman (Biederman et al. 1982) as "familiar size," refers to the relationship between objects based on their relative sizes within a scene. In computer vision, scale context helps detectors manage the variability of object sizes, enabling more accurate detection at different scales. By understanding the expected sizes of objects within a scene, a detector can reduce the need for exhaustive multi-scale searches. Objects within a scene often follow a limited set of size relations. For instance, a chair typically appears smaller than a person and not larger. Semantic context plays a role in both spatial and scale contexts. For example, in a scene depicting an office, the presence of a desk, keyboard, and monitor is dictated by semantic context (office setting). Semantic context implies spatial context, as the keyboard and monitor are expected to be placed close to each other, and it also implies scale context, as the desk is expected to be larger than both.

Both spatial and scale contexts benefit from semantic context, as the semantic coherence of a scene determines the likelihood of certain objects and their relationships appearing together. These interconnected relationships enhance the overall understanding and detection of objects within a scene. However, scale relationships can also vary depending on the relative depth of objects within the scene. For example, as shown in Fig. 17, the apparent scale relationship between a tree and an elephant is influenced by their similar depth in the image. If the tree were further back, it would naturally appear smaller, highlighting that depth affects perceived scale. Therefore, scale context in object detection should consider not only object sizes but also their positions in depth, as this affects how they are perceived and how bounding boxes are generated. This relationship between scale context and depth could be incorporated by using depth-aware methods, which adjust object size predictions based on their distance from the camera. Models like Feature Pyramid Networks Lin, Dollár, et al. (2017) and multi-scale convolutional networks (Zhao et al. 2017) provide examples of techniques that capture scale context; however, further advancements could involve depth-sensitive context to enhance robustness in varied scenes.

### 2.4.4 Spectral context

Spectral context refers to the relationships and patterns that exist between different colors or spectral bands in an image. This type of context is often used in remote sensing image analysis (Shaw and Burke 2003) and food safety control (Feng and Sun 2012). In spectral image analysis, an image is often composed of multiple spectral bands, each representing a different wavelength or color. In remote sensing and also other applications, there are two main bands, Near-infrared (NIR) bands and color or visual bands (VIS) such as red, green, and blue. NIR and color bands are often used to capture information about the reflectance properties of the scene or Earth's surface. The main differences between NIR and the color bands are their wavelength and the information they capture. NIR is not visible to the human eye because of having a longer wavelength than visible light, but it can provide valuable information about environments, while the color bands capture information about the visible features in the scene. Spectral context refers to the fact that the colors in an image are not independent, but are instead related to one another in some way. Certain colors might be more strongly related to each other than to other colors in the image. For instance, in a satellite image of vegetation as shown in Fig. 18, the reflectance values in the near-infrared band may be highly correlated with the reflectance values in the red band, since vegetation reflects strongly in the near-infrared and absorbs strongly in the red. This relationship between the near-infrared and red bands is an example of spectral context. Generally, spectral context pertains to the utilization of distinct "colors" of light in order to gain further insights into distant observations. By considering the spectral context of an image, a machine can gain insights into the underlying properties of the scene, such as the presence of objects, vegetation, water, or other materials.

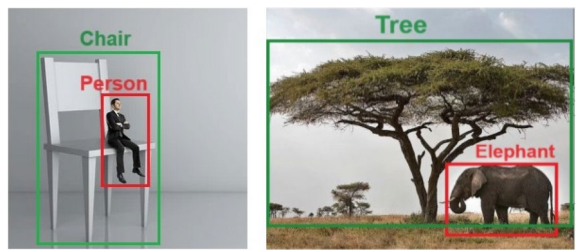

**Fig. 17** Scale context. Left: large chair compared to a person. Right: normal scaled elements

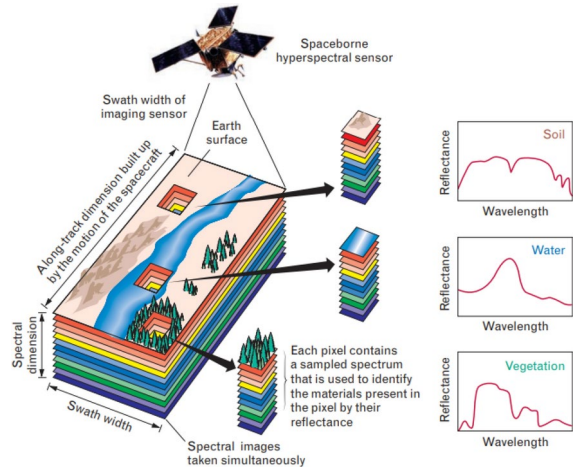

**Fig. 18** Multiple spectral wavebands spread across a vast region. The graphs show the spectrum variation in the reflectance of soil, water, and vegetation. The spectral information can be used to create a visual representation of the scene at different wavelengths (Shaw and Burke 2003)

### 2.4.5 Spatial-spectral context

Spatial-spectral context refers to the idea that the spectral properties of an object in an image are not only determined by its intrinsic reflectance properties but also by its spatial relationships to other objects in the scene. In remote sensing, spatial-spectral context can be used to improve the classification of objects in an image by considering the spatial relationships between objects as well as their spectral properties. A system that considers both spectral and spatial context can discern objects inside an image, as well as different properties like vegetation, roads, and buildings. Figure 19 is an example of a two-stream spectral-spatial feature aggregation approach titled S2ADet (He et al. 2023) that leverages complementary spatial and semantic information to learn better semantic features of objects. Moreover, spatial-spectral context is also used in image classification with impressive results. (Fauvel et al. 2012) and (Li et al. 2021) are examples of using spatial-spectral context in image classification.

### 2.4.6 Temporal context (time)

Time can be defined as a measure in which events can be ordered from the past through the present into the future, and also, as the measure of the durations of events and the intervals between them (Lim et al. 2019). Using information from previous time frames in order to enhance understanding of the present frame is defined as the concept of "temporal context". Temporal context is mostly used for dynamic data like videos. Schilder et al., (Oliva and Torralba 2007) proposed three types of temporal expression: explicit references (e.g., December 4), indexical references (e.g., yesterday), and vague references (e.g., about two years ago). Recently, (Wang and Zhu 2023) presented two forms of temporal context, including short-term temporal context (relations between frames in a short video) and long-term temporal context (relations between frames in a long video). In this review, we follow the proposed forms by (Wang and Zhu 2023). Through the investigation of disparities and resemblances across neighboring frames, short-term temporal context can be employed to identify and monitor objects, as well as activities. When recognizing an item necessitates evaluating a large number of frames over a lengthy period of time, short-term temporal context cannot give very valuable information; thus, long-term temporal context plays an important role (Beery et al. 2020). Long-term temporal context pertains to the interconnections that exist among frames within an extended video sequence. Long-term temporal context can be used to detect and track objects over longer periods of time, as well as recognize more complex actions and events, by examining the overall structure and patterns of motion in a video clip. In video object detection, temporal context is undeniably significant. Objects may frequently change appearance or motion between frames in a video sequence, making it challenging to detect and track them precisely using spatial information alone. By integrating temporal context into object detectors, researchers can leverage the interdependencies across frames in a video sequence to enhance the precision and robustness of object detection. Figure 20 is one example of leveraging long-term temporal context for detecting one wildebeest in a wild, which needs to analyze many frames over a long time (one month).

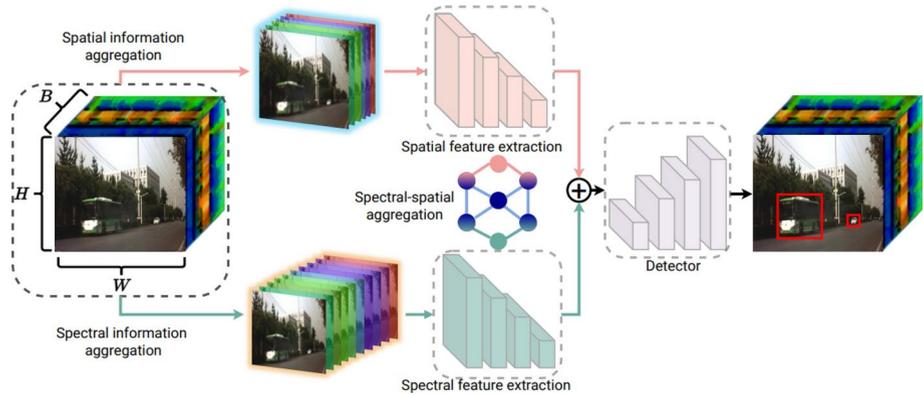

**Fig. 19** Combining spatial and spectral context for detecting objects (He et al. 2023)

### 2.4.7 Spatial-temporal context

Spatio-temporal or spatial-temporal context refers to the combination of spatial and temporal information. It involves analyzing not only the spatial relationships between objects in a scene but also their temporal relationships over time. By incorporating both spatial and temporal information, spatio-temporal context can provide a richer understanding of dynamic and complex scenes. This context can be used to capture both short-term and long-term temporal dependencies, such as the motion of objects or the evolution of a scene. For example, in action recognition, spatio-temporal context can be utilized to capture the dynamic motion patterns of a particular activity, such as walking or running, or in video analysis, spatio-temporal context is used to detect and track objects over time. Berg et al. (2014) introduced a spatio-temporal prior for improving the accuracy of bird species classification. In their approach, location and time are discretized into spatio-temporal cubes, and a kernel density estimate is used to determine the distribution of each species on an individual basis. In another paper (Mac Aodha et al. 2019), one spatio-temporal model for integrating all spatial context, long term temporal context (months to years), and semantic context together was proposed. Hao Luo et al. (Luo et al. 2019), as shown in Fig. 21, used spatio-temporal context to enhance the network for detecting objects in difficult conditions such as video defocus, motion blur, occlusion, etc.

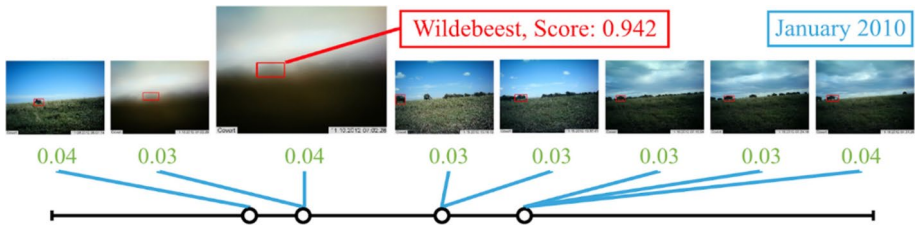

**Fig. 20** Using previous and next frames of one video for detecting a wildebeest in a blurry frame (Beery et al. 2020)

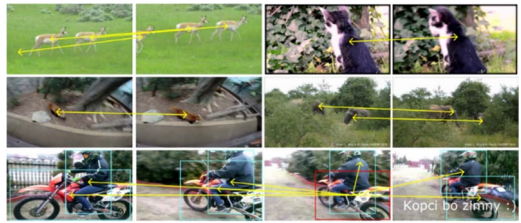

**Fig. 21** Distinctions between pixel/instance level feature correspondence in the top and middle rows, and the spatio-temporal context in the bottom row. The former is susceptible to issues such as the appearance of new objects, and occlusion, whereas the latter depicts the interdependence between intra-frame and inter-frame proposals (Luo et al. 2019)

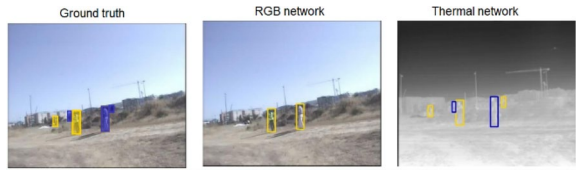

**Fig. 22** Comparison between thermal network and RGB network for finding objects (Banuls et al. 2020)

### 2.4.8 Thermal context

Thermal context in computer vision refers to the use of thermal imaging data to improve the performance of computer vision algorithms. Thermal images are visual displays of measured emitted, reflected, and transmitted thermal radiation by objects within an environment (Berg 2016). For example, in surveillance systems, thermal imaging can be used to detect people in low light or complete darkness, where visible light cameras would not be effective Fig. 22 shows one example of using thermal context to detect people in darkness. To incorporate thermal context into computer vision algorithms, researchers often use specialized thermal cameras to capture thermal images of the environment (Krišto et al. 2020). They also use fusion techniques to combine thermal and visible light images to gain a more complete understanding of the environment (Zhong et al. 2017).

### 2.4.9 Photogrammetric context

Photogrammetry is the science of extracting 3D information about objects and structures from photographs. It involves taking photographs of an object or scene from different angles and then using specialized software to analyze the images and create a 3D model of the object or scene. Photogrammetric context refers to as intrinsic and extrinsic camera parameters of image capturing. Intrinsic camera parameters refer to the internal properties of the camera, such as focal length, radiometric response, lens distortion, affinity, and shear (Lin et al. 2022). These parameters are used to calculate the relationship between the 3D world and the 2D image captured by the camera, which is important for accurate object detection and recognition. For example, knowing the focal length of the camera lens can help in estimating the size and distance of objects in the scene. Extrinsic camera parameters refer to the position, camera height, and orientation of the camera in the space (Hoiem et al. 2008). For instance, in the case of aerial photography, the height of the camera above the ground can be used to estimate the size and location of objects in the scene. By combining the intrinsic and extrinsic parameters of the camera, it is possible to create a more accurate and complete model of the environment. For example, knowing the camera height and lens distortion can help in accurately detecting objects in a scene, even if they are partially occluded or have irregular shapes. Figure 23 is one example of using photogrammetric context to enhance the performance of the object detector.

### 2.4.10 Geographic context

Geographic context refers to the use of location-based information to improve the performance of computer vision algorithms. Geographic context takes into account the fact that images and videos are often captured in specific geographic locations, which can provide valuable contextual information about the scene. As shown in Fig. 24, it can specify the actual location of the image like GPS, or it can indicate a more generic land type such as desert, ocean, urban, or agricultural areas. For example, knowing that an image was captured in a forest area could increase the probability of detecting wildlife rather than vehicles. (Ardeshir et al. 2014), (Groenen et al. 2023), (Wang et al. 2015), and (Wang et al. 2017) are examples of using geospatial context for detecting objects.

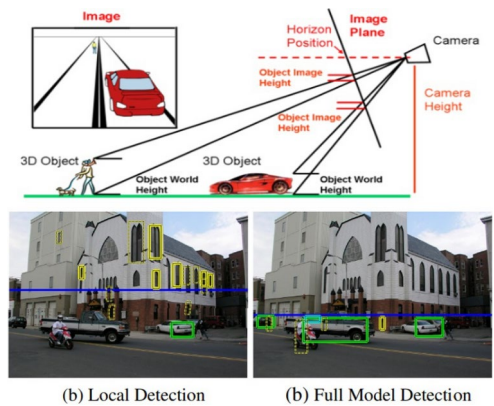

**Fig. 23** The top image shows the effect of camera height, horizon position on detecting objects in the environment. In the bottom image, the blue lines represent the estimated horizon. The green boxes indicate true cars, while the yellow boxes represent false positives for pedestrians. With the integration of photogrammetric context in the full model detection, the number of false detections has decreased, while the number of true detections has increased (Hoiem et al. 2008)

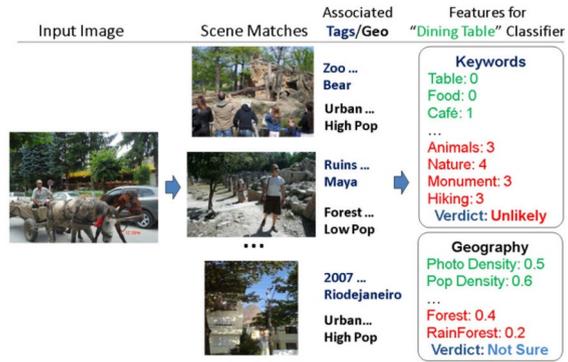

**Fig. 24** By leveraging semantic context, geographic context, and keywords linked to the scene, it is possible to forecast the presence of objects in an image (Divvala et al. 2009)

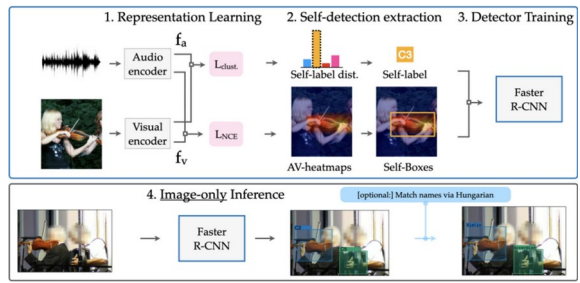

**Fig. 25** Utilizing a combination of noise-contrastive and clustering-based self-supervised learning to create self-detections (boxes and labels) and then utilizing those as targets to train a detector (Afouras et al. 2022)

### 2.4.11 Audio context

Audio context can provide information about environments that visual data alone cannot capture. For example, in surveillance systems, sound context can be used to detect gunshots or breaking glass, which can indicate potential security threats. In autonomous driving systems, sound context can be used to detect and recognize different types of vehicles or to detect and avoid obstacles that are not visible in the visual data. The integration of audio context into multimodal models, using techniques such as multimodal fusion (Gao et al. 2020) and attention mechanisms (Lieskovská et al. 2021) enables a comprehensive comprehension of the scene by capturing the relationships between visual and audio data. Figure 25 provides an example of leveraging audio context to detect objects alongside visual feature maps.

In addition to the presence of objects, audio also determines distances and even directions. For example, by hearing the sound of a bird, in addition to identifying the bird as an object, the approximate distance and direction of the sound can also be identified. Figure 26 is an example of leveraging sound context with visual features to create an audio-visual event localization framework in unconstrained videos.

### 2.4.12 Text context

Text context can provide important information about a scene or object. Text context can be used to enhance the performance of computer vision tasks such as object recognition, scene understanding, and image retrieval (Mishra et al. 2013). Figure 27, for example, is an image-text retrieval system, and text context can be used to search for images with specific labels or descriptions. In a scene understanding system, text context can be used to identify specific objects within the scene or to infer relationships between objects based on their labels or descriptions. To incorporate text context into computer vision algorithms, natural language processing (NLP) techniques are effective approaches to analyze and extract textual information from various sources, such as image captions, object labels, or product descriptions. Machine learning algorithms can then be trained on this data to recognize patterns and identify specific objects or scenes based on their textual context.

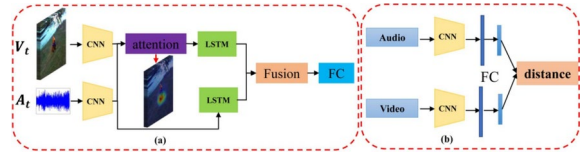

**Fig. 26** Audio-visual event localization framework with audio-guided visual attention and multimodal fusion (Tian et al. 2018)

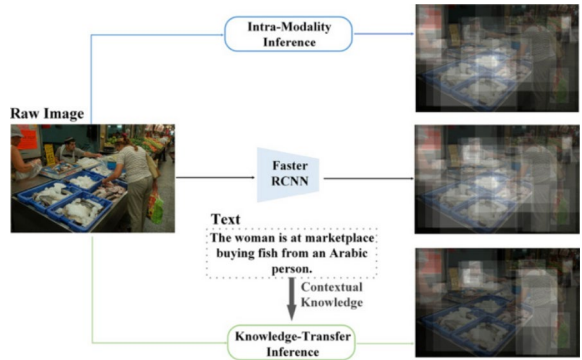

**Fig. 27** Visual context learning based on textual knowledge for image-text retrieval (Qin et al. 2022)

### 2.4.13 Illumination context

Illumination context refers to the study of how lighting conditions affect the appearance of objects in a scene and how this information can be used to improve computer vision algorithms. Illumination context involves information such as sun direction (Lalonde et al. 2008), sky color, shadow contrast, and covered by clouds. Illumination context is important because the way light interacts with objects in a scene can dramatically affect their appearance. By analyzing the illumination context of a scene, computer vision algorithms can adjust for variations in lighting conditions and make more accurate predictions about the appearance and behavior of objects in that scene.

### 2.4.14 Weather context

In outdoor applications, there is no escape from "bad" weather. Ultimately, computer vision systems need to leverage mechanisms that enable them to function in the presence of haze, fog, rain, hail, and snow (Narasimhan and Nayar 2002). The bad weather, however, turns out to have a positive side since it could serve as a powerful means for coding and conveying scene structure (Narasimhan and Nayar 2002). Weather context would describe meteo-rological conditions such as temperature, wind speed, or direction, and weather conditions such as rain, snow, mist, and different seasons.

## 3 Research method

A review protocol is developed to guide the conduct of the literature survey. Research questions (RQs) as mentioned in Sect. 1, and a set of criteria determine the topic and the objective of this literature review.

### 3.1 Identification of bibliographical databases

To conduct this literature review, three major and well-known bibliographical databases with good coverage of computer science were selected: IEEE Xplore, Web of Science (WoS), and Scopus. According to (Stapic et al. 2012), it is important to determine the starting and ending dates of a literature review; thus, in order to focus on the state-of-the-art methods, the period from 2018 to 2023 is selected as the date range.

### 3.2 Searching and selection of primary studies

A Boolean search criterion was utilized to search the databases. "(Title ((Context) AND (Object detection)) OR Abstract ((Context) AND (object detection)))". Duplicate papers in different databases have been removed. The results are shown in Table 1.

### 3.3 Inclusion and exclusion criteria

According to (Kitchenham 2004), inclusion and exclusion criteria should be based on research questions. The inclusion and exclusion criteria are shown in Table 2. It should be noted that categories such as salient, RGB-D, remote sensing, 3D, moving, and adversarial object detection were excluded due to their specialized nature, lower recent research volume, and narrower practical applications. This allows for a more focused and relevant survey, ensuring thorough and detailed analysis of the most active and impactful areas in object detection.

### 3.4 Data extraction and validity control

An overview of the data extraction strategy is shown in Fig. 28. A total of 265 papers were retrieved from three academic databases. Papers that are duplicated in different databases are considered in only one of the sources. Out of 265 papers, 2 papers were rejected for being either a survey or a review. Additionally, 2 papers were excluded because they were not written in English, and 126 papers were excluded for emphasizing context in salient

Table 1  Papers retrieved from databases (Date Range: 17 April 2023 - 1 October 2024)

| Database | Papers | After removing duplicates |
| --- | --- | --- |
| IEEE Xplore | 172 | 172 |
| Scopus | 68 | 61 |
| Web Of Science | 78 | 32 |
| Total | 318 | 265 |

Table 2  Inclusion and exclusion criteria used in the systematic literature review

*Inclusion criteria*
- Full papers published between January 2018 and April 2023
- Papers with a focus on context in object detection

*Exclusion criteria*
- Papers with a focus on salient object detection, RGB-D object detection, remote sensing object detection, 3D object detection, moving object detection, and adversarial object detection
- Papers not written in the English language
- Papers which are not published in a peer-reviewed journal or conference
- Being a survey or review
- Lack of a final mAP and using other metrics

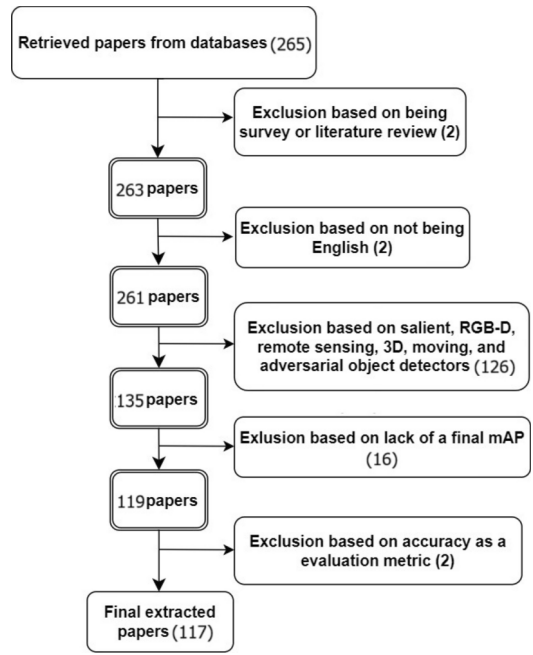

**Fig. 28** Data extraction strategy

object detection, RGB-D object detection, remote sensing object detection, 3D object detection, moving object detection, and adversarial object detection that are not covered in this literature review. Then, due to the absence of a final mAP in 16 papers and the use of alternative evaluation metrics in 2 papers, they were eliminated. Finally, 117 papers qualify for this systematic literature review. All these 89 papers were reviewed and classified into different application areas: context in general object detection, context in small object detection, context in video object detection, context in few-shot object detection, context in one-shot object detection, context in zero-shot object detection, and context in camouflaged object detection. In all categories, method or model, level and type of context, backbone and architecture, mechanism or module for exploiting and integrating contextual information, dataset, and mAP as the evaluation metrics were investigated.

**Fig. 29** Distribution based on years of the publications

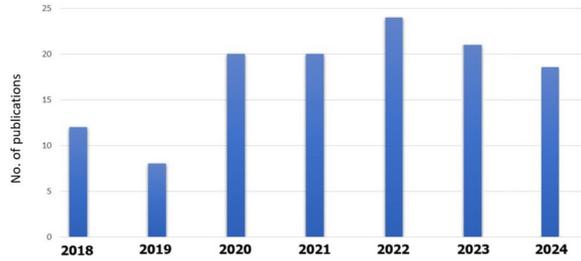

**Fig. 30** The number of context-based papers in different categories of object detection

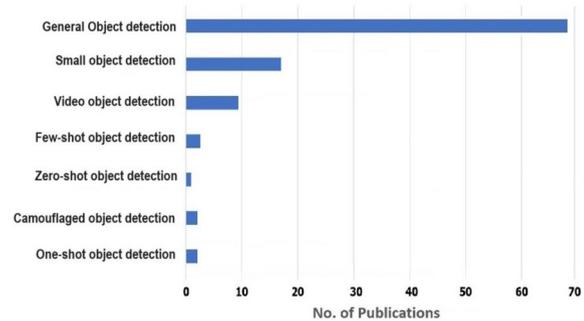

## 3.5 Classification of the reviewed papers over time

Context has gotten greater attention in recent years than in the past. According to Fig. 29, the number of publications that have employed contextual information to improve object detection is increasing. The Fig. 29 demonstrates that the utilization of contextual information in computer vision is an active area, drawing the attention of more researchers and practitioners. Please note that the statistics for 2023 only include papers published during January to April.

## 3.6 Context in categories of object detection

As shown in Fig. 30, in this literature review, we analyze seven distinct categories of object detection. The rationale behind the selection of these seven categories is as follows. Firstly, both video object detection and general object detection are extremely prevalent and currently garnering significant attention. Secondly, detecting small objects continues to pose a substantial obstacle in the field of object detection. Thirdly, no comprehensive research has been conducted on object detection using context in zero-shot object detection, one-shot object detection, few-shot object detection, and camouflaged object detection. Most of the papers under review have focused on using context in general object detection. The limited number of papers in zero-shot, one-shot, few-shot, and camouflaged object detection indicates that these areas have not yet fully benefited from contextual information and still have considerable room for growth.

### 3.7 Data extraction and synthesis

In the following steps, we extracted relevant information regarding RQs from the papers. For each paper, we extracted the following information: (1) Detector name (2) Employed context types (3) Employed context levels (4) Backbone and architectures (5) Mechanisms or modules to integrate contextual information (6) Evaluation metrics: We use mean Average Precision (mAP) to evaluate and compare the effectiveness of methods across studies, focusing on mAP averaged over Intersection over Union (IoU) thresholds from 50% to 95% to provide a balanced and comprehensive comparison. For methods evaluated on the PASCAL VOC (2007 and 2012) datasets, we used the mAP metric based on the 50% IoU threshold (mAP@50) with 11-point interpolation, consistent with the VOC benchmark. For COCO, we adopted the COCO-standard mAP at IoU thresholds from 50% to 95% in 5% increments (mAP@50-95) with 101-point interpolation, and included COCO-specific metrics like AP for small (APs), medium (APm), and large (APl) object scales. For other datasets, we used mAP as reported in the referenced papers, to allow for consistent evaluation across different methods. Additionally, Mean Absolute Error (MAE), F-measure, S-measure, and E-measure were included for camouflaged object detection, as these metrics are commonly used in this domain. For Zero-shot, One-shot, and Few-shot object detection, apart from mAP, the comparison focuses on the unique learning approaches within each approach, as they play critical roles in evaluating models. (7) Dataset used for evaluation.

## 4 Analysis and discussion

### 4.1 Datasets

In object detection research, several datasets have become benchmarks due to their diversity, complexity, and real-world relevance. Among the most prominent are MS COCO (Lin et al. 2014) and PASCAL VOC2007 (Everingham et al. n.d.) and 2012 (Everingham and Winn 2012), widely used across various object detection tasks. MS COCO contains over 200k annotated images and 80 object categories, presenting challenges such as detecting small, medium, and large objects, managing cluttered scenes, and understanding complex contextual interactions. Similarly, PASCAL VOC2007 and PASCAL VOC2012 are essential benchmarks with 20 object classes, where models must contend with occlusions, variations in object viewpoints, and the presence of small instances.

In specific domains, datasets such as Cityscapes (Cordts et al. 2016) are critical for urban scene understanding, challenging models with crowded environments, diverse lighting conditions, and occlusions. BDD100K (Yu et al. 2020) is a large-scale dataset designed for autonomous driving that introduces challenges in detecting objects across diverse conditions, including nighttime driving, rain, and heavy traffic, requiring models to handle both small and large objects in dynamic street scenes. DOTA (Xia et al. 2018) addresses object detection in aerial imagery, presenting challenges like rotated objects and significant scale variations, particularly for detecting airplanes, vehicles, and ships. In face detection, WIDER FACE (Yang et al. 2016) is a well-known dataset that presents extreme variations in occlusion, pose, and lighting conditions, making it highly challenging for models to generalize across real-world crowded scenarios. For wildlife detection, Caltech Camera Traps (Beery et al. 2018) introduces the challenge of detecting animals in natural environments, where factors like camouflage and low-contrast conditions complicate detection tasks. S-UODAC2020 (Chen et al. 2023), designed for underwater object detection, tests models' adaptability to low visibility, distortion, and dynamic marine environments.

While these datasets are widely used for benchmarking, each has certain limitations and biases. For instance, MS COCO, with its rich variety, predominantly features objects common in western urban environments, which can bias models toward specific cultural contexts and affect generalizability in diverse settings. Similarly, PASCAL VOC datasets, while influential, lack the breadth of modern datasets, with fewer object classes and limited data diversity. The Cityscapes and BDD100K datasets focus on urban driving scenarios, which may bias models toward detecting road-related objects, potentially limiting performance in rural or off-road contexts. DOTA, in aerial imagery, presents unique challenges like scale variance but also emphasizes certain object types, such as vehicles and buildings, potentially reducing the robustness of models in other aerial domains. WIDER FACE and Caltech Camera Traps datasets, though comprehensive within their domains, are biased by their respective collection methods: WIDER FACE images come from news and media, possibly skewing model performance in casual or non-professional images, while Caltech Camera Traps may bias models toward recognizing specific animal species or environments. Lastly, underwater datasets like S-UODAC2020 are limited by the constraints of underwater imaging technology, which can reduce detection accuracy in diverse marine environments. Addressing these biases is crucial for developing robust object detection models that generalize well across varied real-world conditions. More comprehensive details about different datasets, along with their categories and attributes, can be found in Table 3, which organizes them by task, data type, number of classes, and dataset size.

### 4.2 General object detection (GOD)

General object detection entails the identification and localization of various objects within images or videos, belonging to different categories. This task is essential for different applications, such as autonomous vehicles, surveillance, image retrieval, augmented reality, etc. The papers have used different approaches to integrate context into networks. We have categorized them into seven approaches: (1) graph-based approaches 4.2.1, (2) hierarchical approaches 4.2.2, (3) context data augmentation 4.2.3, (4) multi-scale approaches 4.2.4, (5) RPN-based approaches 4.2.5, (6) attention-based approaches 4.2.6, and (7) other approaches 4.2.7. All approaches are shown in Fig. 31. Approaches highlighted in red are exclusive to two-stage models, those highlighted in blue have been designed merely for one-stage models, and green-highlighted approaches are weakly supervised object detectors. Extracted information and mAPs of the papers have been compared in three different tables: (1) papers that utilized the COCO dataset in Table 5, (2) papers that utilized the Pascal dataset in Table 6, and (3) papers that employed other datasets in Table 7.

#### 4.2.1 Graph-based approaches

Graph-based methods enhance object detection by encoding relationships between objects as a graph structure. In this setup, nodes represent objects, and edges denote relationships, providing a structured way to model spatial and semantic context. This approach allows

Table 3 Overview of datasets utilized by context-based object detection approaches

| Dataset | Task | Data type | Classes | Size |
| --- | --- | --- | --- | --- |
| MSCOCO Lin et al. (2014) | General Object Detection | Image | 80 | 200k |
| PASCAL VOC2007 Everingham et al. (n.d.) | General Object Detection | Image | 20 | 9,963 |
| PASCAL VOC2012 Everingham and Winn (2012) | General Object Detection | Image | 20 | 11k |
| LVIS Gupta et al. (2019) | General Object Detection | Image | 1203 | 164k |
| ILSVRC Russakovsky et al. (2015) | General Object Detection | Image | 1000 | 1,431,167 |
| WIDER FACE Yang et al. (2016) | Face Detection | Image | 61 | 32k |
| FDDB Jain and Learned-Miller (2010) | Face Detection | Image | 1 | 2,845 |
| CelebA Liu et al. (2015) | Face Detection | Image | 40 | 202,599 |
| PASCAL3D+ Xiang et al. (2014) | 3D Object Detection | Image | 12 | 20k |
| VisDrone Zhue t al. (2021) | Aerial and Drone Detection | Image, Video | 14 | 10k images, 263 videos |
| WPAFB 2009 AFRL (2009) | Aerial and Remote Sensing | Image | 6 | 1,537 |
| UCAS-AOD Zhu et al. (2015) | Aerial and Remote Sensing | Image | 2 | 1,510 |
| RSOD Long et al. (2017) | Aerial and Remote Sensing | Image | 4 | 976 |
| Airport Cui et al. (2020) | Aerial Imagery | Image | 1 | 1,111 |
| NWPU VHR-10 Zhong et al. (2018) | Aerial Remote Sensing | Image | 10 | 650 |
| DOTA Xia et al. (2018) | Aerial Remote Sensing | Image | 15 | 2,806 |
| TJ-LDRO Chen et al. (2021) | Small Road Object Detection | Image | 18 | 109,337 |
| CAMO Le et al. (2019) | Camouflaged Object Detection | Image | 8 | 1,250 |
| Chameleon Skurowski et al. (2018) | Camouflaged Object Detection | Image | 5 | 76 |
| COD10K Fan et al. (2021) | Camouflaged Object Detection | Image | 5+69 | 5,066 |
| NC4K Lv et al. (2021) | Camouflaged Object Detection | Image | – | 4,121 |
| Cityscapes Cordts et al. (2016) | Urban Scene Understanding | Image | 30 | 25k |
| CityCam Zhang et al. (2017) | Urban scene understanding | Video | 19 | 60k frames |
| Tsinghua-Tencent Zhu et al. (2016) | Traffic Sign Detection | Image | 45 | 100k |
| Bosch traffic lights Behrendt and Novak (2017) | Traffic Light Detection | Image | 15 | 13,427 |
| NM10k Li et al. (2022) | Autonomous driving | Image | 7 | 14k |
| BDD100k Yu et al. (2020) | Autonomous driving | Image | 10 | 80,000 |
| KITTI Geiger et al. (2012) | Autonomous driving | Image | 3 | 14k |
| Aerial Vehicle Zhao et al. (2022) | Aerial Vehicle Detection | Image | 7 | 2,950 |
| OCS driving Lu et al. (2020) | Vehicle Detection | Image | 10 | 2,597 |
| MIO-TCD Luo et al. (2018) | Vehicle Detection | Image | 11 | 786,702 |
| OTB50 Wu et al. (2013) | Vehicle Detection | Video | - | 49 Videos |
| SIXray Miao et al. (2019) | X-ray Object Detection | Image | 6 | 1,059,321 |
| OPIXray Wei et al. (2020) | X-ray Object Detection | Image | 5 | 8,885 |
| WIXray Qiu et al. (2022) | X-ray Object Detection | Image | 12 | 5,038 |
| ALPR Hossain et al. (2022) | License plate detection | Image | 12 | 6,800 |
| Underwater Yang et al. (2023) | Underwater object detection | Image | 4 | – |
| S-UODAC2020 Chen et al. (2023) | Underwater object detection | Image | 4 | 5,537 |
| UTDAC2020 SRMIST (2023) | Underwater object detection | Image | 4 | 6,461 |

**Table 3** (continued)

| Dataset | Task | Data type | Classes | Size |
|---|---|---|---|---|
| FSVOD Fan et al. (2022) | few-shot video object detection | Video | 500 | 1,500 |
| FSYTV Fan et al. (2022) | few-shot video object detection | Video | 40 | 1,800 |
| ImageNet VID Russakovsky et al. (2015) | Video Object Detection | Image | 30 | 1,100k |
| UA-DETRAC Wen et al. (2020) | Video Object Detection | Video | 4 | 140k frames |
| UAVDT Du et al. (2018) | Video Object Detection | Video | 14 attributes | 80k frames |
| Caltech Camera Traps Beery et al. (2018) | Wildlife detection | Image | 21 | 196k |
| GDD Mei et al. (2020) | Glass Surface Detection | Image | 6 | 3,916 |

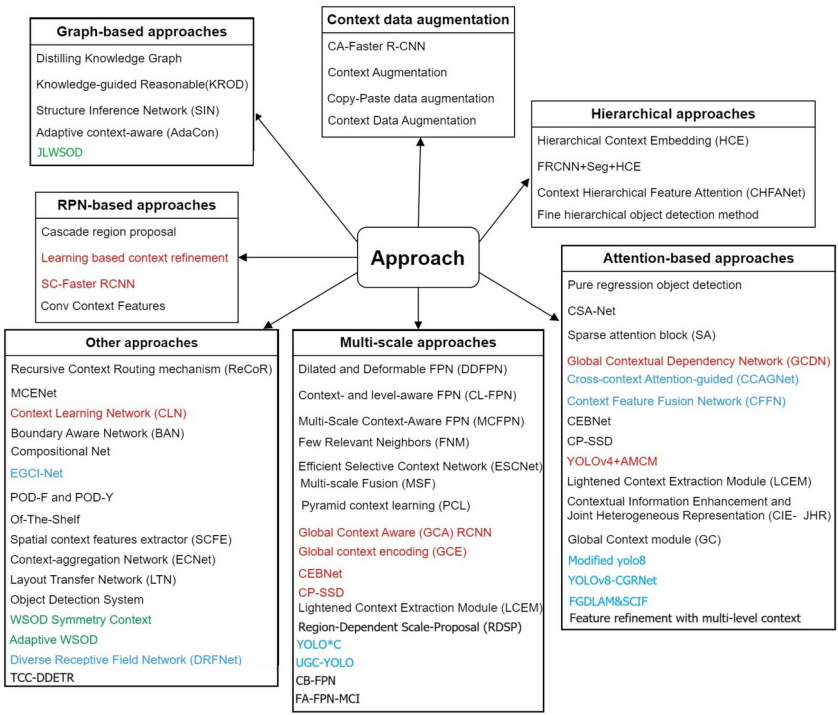

**Fig. 31** Employed approaches for integrating context into general object detection

for better understanding of the scene, as the network can leverage the co-occurrence and spatial positioning of objects, which often provide vital clues for detection. Figure 32 illustrates this, showing how object relationships are captured as graph edges, facilitating scene understanding. For example, a person is next to a skateboard based on spatial context, and a helmet and skateboard commonly appear together based on semantic context.

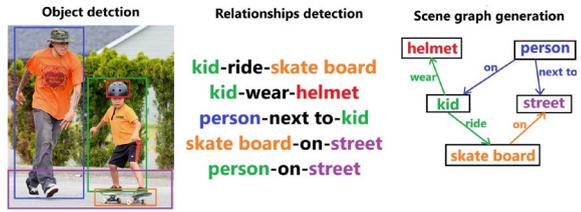

**Fig. 32** Visual relation detection and its scene graph representation

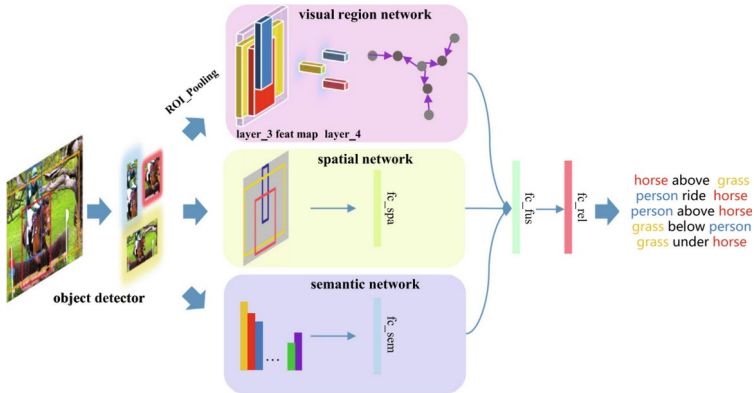

**Fig. 33** Integrating spatial and semantic context into a graph-based model (Zhang et al. 2021)

Figure 33 demonstrates how context can be integrated into a graph-based object detection framework. The object detector first identifies bounding boxes for detected objects. Then, Visual Region Network builds a graph with objects as nodes and relationships as edges, capturing spatial connections like 'above' or 'next to'. Spatial Network encodes spatial relationships, enhancing the model's understanding of object positioning. Semantic Network processes category relationships, identifying associations like 'person rides horse'. By combining these contextual layers, the model refines object detections, as seen in relationships like 'horse above grass'.

In this section, we explore five graph-based approaches, each leveraging context differently to enhance object detection.

(1) Distilling Knowledge Graph (Yang et al. 2023) seamlessly integrates spatial and semantic context into object detection using a knowledge graph and knowledge distillation (KD). By leveraging a teacher-student model, the method constructs both geometry and semantic graphs through a transformer layer, capturing object interactions at both local and global levels. The edges in the graph encode relationships, which are incorporated into the attention matrix to enable graph-level attention, enhancing contextual learning. Among graph-based approaches in this paper tested on COCO dataset, this method achieves the highest AP, APs, APm, and APl.

(2) Knowledge-guided Reasonable Object Detection (KROD) (Ji et al. 2022) focuses on global context and category relationships. It introduces Global Category Knowledge Mining (GKM) that integrates multi-label image-level classification results to provide global category knowledge for the detector. Additionally, the raw detection outputs were inputted by the Category Relationship Knowledge Mining (CRM) module into the object category co-occurrence-based knowledge graph to further refine the initial results.

(3) The Structure Inference Network (SIN) (Liu et al. 2018) enhances object detection by integrating both scene-level and instance-level contextual information within a graphical model framework. In SIN, objects are represented as nodes in a graph, and their relationships form the edges, allowing the model to capture object interactions and scene context. Incorporated into a standard detection framework like Faster R-CNN, SIN utilizes this structured context to iteratively update each object's state, refining predictions based on both local appearance and surrounding context, thus improving accuracy in complex scenes.

(4) Adaptive context-aware object detection (AdaCon) (Neseem and Reda 2021) uses the spatial co-occurrence probabilities of object categories to create an adaptive network. A branch controller selects which sections of the network to execute during runtime based on the spatial context of the input frame. AdaCon is the first detector to present an adaptive algorithm for one-stage object detectors.

(5) JLWSOD (Lai et al. 2024) a method for weakly supervised object detection, integrates two types of contextual information: instance-wise correlation and semantic-wise correlation. The framework comprises three key modules: the Instance-Wise Detection Branch, which enhances object localization by capturing correlations among spatially adjacent instances; the Semantic-Wise Prediction Branch, which addresses semantic ambiguity by modeling relationships between co-occurring object categories; and the Interactive Graph Contrastive Learning (iGCL) module, which facilitates the joint optimization of both contextual information types. This interactive learning mechanism allows for effective propagation of image-level supervisory signals to instance-level predictions.

Each of the reviewed models demonstrates distinct approaches to incorporating context in object detection, revealing both strengths and limitations in various technical aspects. The Distilling Knowledge Graph approach is computationally efficient, benefiting from a knowledge graph that integrates spatial and semantic context, but relies heavily on a teacher-student framework, which can complicate training setups. By contrast, Knowledge-guided Reasonable Object Detection (KROD) provides robust contextual reasoning through modules like Global Knowledge Mining and Category Relationship Mining, effectively suppressing false positives without extensive pre-processing. However, its dependency on prior co-occurrence knowledge graphs limits adaptability to uncommon object relations. Structure Inference Network (SIN) excels by employing a dual-context approach, leveraging scene-level and instance-level relationships to enhance both classification and localization accuracy. This method, however, struggles with rare or unexpected contexts due to its reliance on generalized scene assumptions and is highly dependent on well-structured graphs, which are complex to optimize. On the resource-constrained side, AdaCon stands out with its adaptive efficiency, selectively executing branches based on spatial context, which conserves energy and reduces latency-though at the cost of potentially lower precision on rare object combinations. Finally, JLWSOD offers a novel solution for weakly supervised settings by optimizing instance and semantic correlations simultaneously, enhancing detection performance across crowded scenes. Yet, it faces challenges with small objects and is com-putationally intensive due to its interactive contrastive learning mechanism. Together, these models illustrate the progressive advances in integrating context, highlighting an ongoing trade-off between computational efficiency, adaptability to novel contexts, and detection accuracy across various deployment environments.

### 4.2.2 Hierarchical approaches

Hierarchical approaches involve organizing visual information in a layered or multi-level manner to facilitate efficient processing and analysis, allowing models to consider context at multiple granularities. In object detection, hierarchical frameworks aim to emulate human perception by first capturing broader, scene-level context and then refining this information through intermediate layers down to precise, localized details. This layered processing strategy enhances models' ability to handle complex scenes by systematically incorporating context from global to local levels, which is particularly useful in scenarios with dense or overlapping objects. Through this process, hierarchical models achieve a balance between high-level contextual awareness and detailed object recognition, leading to improved accuracy and robustness in object detection tasks. Figure 34 illustrates this hierarchical process, showing how the focus narrows progressively from the entire scene to specific regions of interest. Each level in the hierarchy represents a step that integrates contextual cues at varying scales, enabling accurate identification of objects within a complex scene.

Here, we explore four hierarchical approaches, including Hierarchical Context Embedding (HCE), Hierarchical Context Embedding module, Context-Aware Hierarchical Feature Attention Network (CHFANet), and Fine hierarchical object detection.

(1) Hierarchical Context Embedding (HCE) (Chen et al. 2020) benefits from hierarchical context by embedding contextual cues at both image-level and instance-level resolutions. Through the image-level categorical embedding module, it integrates whole-image context to support object-level classification, especially for objects dependent on contextual surroundings. The feature fusion and confidence fusion strategies further exploit this hierarchical embedding, combining global and instance-level contextual cues in the final classification stage. This layered context utilization improves the network's ability to discern objects within cluttered backgrounds and identify contextually linked objects, enhancing robustness in varied detection tasks. The HCE exhibits superior performance on region-based detectors tested on the COCO dataset.

(2) Hierarchical Context Embedding module (Qiu et al. 2020) leverages context by recalibrating noisy segmentation features based on hierarchical attention maps that span different spatial distances. Local context features focus on individual parts of objects (e.g., a person's head or a car's wheel), while non-local context encompasses broader scene-level information (e.g., a group of objects or background elements). By embedding this contextual information into the detection features, the model gains an enhanced ability to discern and differentiate objects in intricate settings, ensuring that both local details and global scene characteristics contribute to accurate detection outcomes.

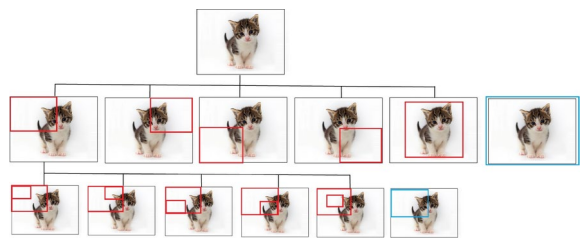

**Fig. 34** Image hierarchy of three levels with non-overlapped quarters

(3) Context-Aware Hierarchical Feature Attention Network (CHFANet) (Xu et al. 2020) benefits from context through its CFE and HFF modules. The CFE module collects context information at different scales using dilated convolutions, creating a context-aware feature map that captures both local and broad contextual cues. This is essential for accurately detecting objects across various scales in an image. The HFF module further strengthens the context utilization by combining high-level semantic information (useful for classification) with low-level spatial information (useful for localization), followed by channel-wise attention to selectively highlight significant features. Through these mechanisms, CHFANet is able to leverage both spatial and semantic context to improve detection accuracy and distinguish between objects more effectively. CHFANet outperforms (Cao et al. 2020) when evaluated on PASCAL 2007 dataset.

(4) Fine hierarchical object detection method (Cao et al. 2020) leverages context primarily through the hierarchical search strategy and the ReNet layer. The top-down search mechanism mimics human perception, analyzing each region in context by progressively zooming in on areas most likely to contain the object. This strategy not only reduces redundant processing but also enables a contextual understanding of each sub-region in relation to the entire image. Additionally, the ReNet layer enhances the agent's ability to interpret spatial context over long distances, aiding in accurate decision-making. By incorporating both local and global contextual cues, the model improves its ability to locate objects that may be partially obscured or surrounded by complex backgrounds.

A technical comparison of these methods reveals distinct strengths and limitations in handling contextual information across different detection scenarios. HCE excels in context-dependent detection, particularly for objects that lack unique visual cues. However, its reliance on global context embedding may lead to misclassification in complex backgrounds, and its multi-layer fusion adds computational load, which can affect efficiency in real-time applications. (Qiu et al. 2020) module takes a more granular approach, integrating local and non-local context with multiscale embeddings, making it more effective in handling cluttered environments. Yet, its dependency on accurate segmentation and the high complexity of attention mechanisms make it challenging for applications with low-quality input data or resource constraints. CHFANet improves multi-scale detection by combining spatial and semantic information through channel-wise attention, which enhances feature selectivity for objects at varied scales. This approach, however, introduces complexity that could slow processing in real-time contexts, and the use of dilated convolutions can introduce artifacts, potentially affecting accuracy in high-resolution images. Lastly, Context-Based Fine Hierarchical Detection offers efficiency by narrowing down search regions through a top-down hierarchical strategy, which is computationally efficient for locating single objects. Yet, its sequential decision-making process can be limiting when multiple objects need detection, and its dependency on a fixed hierarchy reduces flexibility in variable-scale detection tasks. Overall, hierarchical methods show how multi-level context integration can improve object detection, though their complexity and computational demands may limit applicability in real-time or large-scale scenarios.

### 4.2.3 Context data augmentation

While most methods focus on directly incorporating context within model architectures, context data augmentation offers an alternative approach by enriching training data with contextually relevant objects. As illustrated in Fig. 35, context data augmentation can be integrated into the training pipeline to enhance model robustness. This process involves selecting contextually appropriate objects, adjusting their appearance, and blending them into background scenes in a way that aligns with the visual context. By augmenting images with realistic object placements that match scene characteristics, this approach improves the model's ability to recognize and localize objects across diverse environments, ultimately enhancing detection accuracy and generalization.

This category reviews four papers that use context-based data augmentation to enhance object detection by strategically placing synthetic objects within scenes.

(1) Context Augmentation Faster R-CNN (CA-Faster R-CNN) (Leng and Liu 2022) enhances region proposals in two-stage object detectors by initially creating a coarse set and subsequently improving uncertain proposals using appearance and geometry (spatial) information. Furthermore, it uses pair-wise relationships between region proposals to augment global feature information for better recognition outcomes. This method of context augmentation enhances object detection in cluttered scenes, where direct visual cues are insufficient. CA-Faster R-CNN outperforms Context Augmentation (Dvornik et al. 2018) when evaluated on the VOC 2012 dataset.
(2) Context Augmentation (Dvornik et al. 2018) utilizes contextual information by training a CNN to predict likely placements for new objects in augmented images based on surrounding visual cues. This context-aware placement prevents unnatural object positioning, making the augmented data more realistic. The model evaluates the likelihood of an object's presence in a particular bounding box based on the surrounding area, allowing it to integrate objects naturally within scenes. This method particularly benefits detection tasks with ambiguous contexts by ensuring object positioning aligns with common scene configurations, thereby enhancing training efficacy and model accuracy.
(3) Another context-based data augmentation method (Zhang et al. 2021) focuses on copy-paste augmentation, which involves pasting foreground objects onto background images. This method benefits from context by leveraging Context Region Proposal Module (CRPM) to identify regions within a background image where objects can be realistically placed. Class-Location-Size Aware Module (CLSAM) then refines these placements by ensuring that each object's class, location, and size are contextually appropriate for the scene, aligning with common object interactions (e.g., people seated on chairs, bottles on tables). By refining placement based on context, the model creates more photorealistic training data, enhancing the model's ability to generalize to real-world conditions.

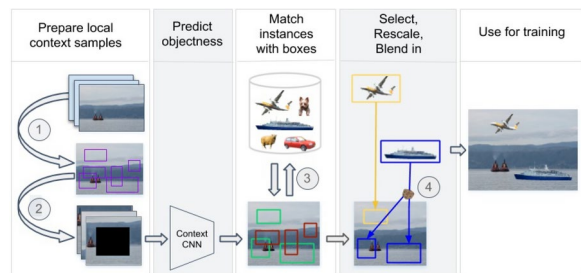

**Fig. 35** Context data augmentation, identifying suitable empty spaces for object insertion based on scene context (Dvornik et al. 2019)

(4) Copy-Paste data augmentation (Li et al. 2022) utilizes context by guiding object placement with both local (instance-level) and global (scene-level) transformations. The local context adapts each instance mask to match real-world scaling and color variations based on its perceived distance from the camera and surrounding conditions. Global context, derived from a multi-task model, directs where objects are placed to maintain logical coherence with traffic environments, such as placing traffic cones only on the ground along lanes and avoiding occlusions with other objects. Together, these contextual insights create realistic data augmentations that strengthen the model's detection accuracy, especially in complex and rare object scenarios.

The methods in context-based data augmentation reveal intriguing trade-offs that highlight potential areas for optimization and improved applicability. For example, while CA-Faster R-CNN's iterative refinement boosts object detection by enhancing region proposal quality, it relies heavily on pairwise relationships, which can limit its generalizability to more complex scenes with ambiguous spatial arrangements. In contrast, methods like (Dvornik et al. 2018) prioritize realism by context-driven object placement, which provides significant accuracy gains in low-data scenarios. However, its reliance on blending techniques, despite their benefits, can sometimes introduce artifacts, reducing consistency in the training data. (Zhang et al. 2021), on the other hand, achieves photorealistic object placement by refining contextual fit through specific class, location, and size considerations, making it highly effective for controlled environments but resource-intensive. (Li et al. 2022) approach is uniquely tailored to traffic contexts, incorporating both local and global transformations that reduce false positives by aligning objects with traffic cues; however, its dependence on domain-specific context restricts its versatility. Therefore, while each method pushes the boundaries of contextual realism in augmentation, future advancements could benefit from hybrid approaches that combine these methods' strengths-such as refining object placement with pairwise and context-aware relationships while remaining computationally efficient and adaptable across varied domains.

### 4.2.4 Multi-scale approaches

In object detection, multi-scale approaches address the challenge of identifying objects at different scales within complex scenes by leveraging techniques like Feature Pyramid Networks (FPN) (Lin, Dollár, et al. 2017). FPN, a widely used architecture in this domain, enhances detection by generating feature maps at multiple scales, where each layer captures a progressively larger or smaller receptive field. As shown in Fig. 36, it involves a bottom-up pathway, which extracts low-level features from input images, and a top-down pathway, which refines these features by combining spatial details and high-level semantics from deeper layers. By merging feature information across levels, FPN helps retain both fine details for small objects and contextual information for larger ones, thereby allowing models to detect objects of various sizes effectively.

Sixteen of the papers reviewed are classified within this category, as outlined below:

(1) Dilated and Deformable Feature Pyramid Network (DDFPN) (Wu et al. 2021) enhances object detection by using context to tailor receptive fields spatially (via Dilated and Deformable Convolution (DDC)) and semantically (via Cross Feature Correlation (CFC) and Co-occurrence Inference (CI) modules), thus accommodating a range of object scales and orientations in complex scenes. It creates more adaptable receptive fields compared to the standard FPN. This multi-level approach enables DDFPN to maintain semantic consistency across object boundaries, supporting the identification of smaller or obscured objects by leveraging cues from their surrounding context.

(2) Context and level-aware FPN (CL-FPN) (Yang et al. 2023) enriches context by capturing spatial and semantic information at different levels. The dilated convolution in Context Enhancement Module (CEM) collects multi-scale context, making the model more adaptable to objects of varying sizes and improving overall detection reliability. Attention-Guided Feature Refinement Module (AFRM) further enhances this by focusing on spatial relationships, which boosts the model's ability to discern context within high-resolution, fine-grained details. Together, CEM and AFRM provide a dual approach to contextual enhancement, strengthening object detection through a comprehensive integration of spatial and semantic cues.

(3) Global Context Aware (GCA) RCNN (Zhang et al. 2021) architecture tackles the problem of losing some contextual information in the process of resizing the object proposal in two-stage object detectors. It improves the spatial relations between the foreground and background by incorporating global context information. The GCA framework incorporates a context-aware mechanism that utilizes a global feature pyramid and attention techniques for the purpose of extracting and refining features, respectively.

(4) Global context encoding (GCE) module (Peng et al. 2022) employs a dual-path approach to fuse top-down, image-level classifications with regional features, making use of high-level contextual cues to improve detection accuracy. This integration enriches the model's understanding of the full scene, allowing it to leverage semantic relationships within the image to identify objects more accurately, especially in complex or cluttered scenes.

(5) CEBNet (Chen et al. 2019) directly enhances low-level feature layers to gather expanded contextual cues, particularly for small objects, through the use of Expansion Receptive Field Block (ERFB). This block captures multi-scale context within a single, efficient structure, making it especially useful for dense and cluttered scenes. Additionally, Feature Attention Block (FAB) ensures feature consistency across scales, enhancing object detection accuracy in diverse contexts by re-weighting features based on channel attention, thus fine-tuning the balance between detailed localization and broad contextual awareness.

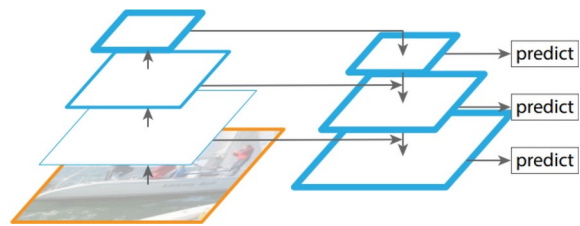

**Fig. 36** Feature pyramid network

(6) CP-SSD (Jiang et al. 2019) captures local and multi-scale contextual information by using dilated convolutions with different rates, allowing for robust context recognition across scales. This method is particularly valuable for scenes with objects that are either small, overlapping, or close to background colors. Additionally, the Semantic Activation Module complements this by learning the interdependence between channels in a self-supervised manner, thus focusing on semantically rich features and strengthening object-background differentiation.

(7) Region-Dependent Scale-Proposal (RDSP) network (Akita and UKita 2023) enhances object detection, especially for small objects, by estimating optimal scale factors based on contextual information derived from the scene structure. The integration of scene context via positional embedding and scene structure embedding allows for a more robust detection process, particularly in challenging conditions with significant object size variations and distance from the camera.

(8) YOLOC's MCTX module (Oreski 2023) adds a multi-task component to the YOLO architecture, enabling it to classify both objects and environmental context in one pass. By evaluating images across multiple scales, YOLOC can interpret spatial and global context cues, such as weather and time of day, directly influencing detection tasks and enhancing accuracy in scenes where these factors significantly alter object appearance and behavior. This explicit modeling of context allows YOLOC to handle complex traffic environments, making it especially valuable in applications like autonomous driving where situational awareness is critical.

(9) CB-FPN (Liu and Cheng 2023) emphasizes capturing spatial and scale-based contextual information. Context Enhancement Module with CSPNet (CEM-CSP) captures diverse receptive fields to retain rich context, effectively bridging scale differences in multi-layer feature fusion. Bidirectional Efficient Feature Pyramid Network (BE-FPN) further enhances context by facilitating bidirectional fusion, ensuring efficient feature propagation across all feature layers, which helps improve object detection by maintaining spatial relationships and context cues across scales. This setup benefits particularly in handling occlusions or objects at multiple scales.

(10) FA-FPN-MCI (Bhalla et al. 2024) leverages multiscale context by using a Twin-Branch Global Context Module (TBGCM) to fuse information across scales and capture global and local features. Style Normalization and Restitution (SNR) module supports domain generalization, aligning feature distributions for consistent performance across varied underwater conditions, while Receptive Field Blocks (RFBs) and deformable convolutions enhance the model's adaptability to varying object scales and complex, cluttered backgrounds in underwater imagery. This approach provides a holistic representation, essential for robust object detection in challenging underwater environments such as color distortion, light attenuation, and complex backgrounds.

(11) Multi-Scale Context-Aware Feature Pyramid Network (MCFPN) (Wang et al. 2022) improves the baseline performance of existing mainstream detectors as an alternative to FPN. This detector has three blocks: The Dilated Residual Block (DRB), Cross-scale Context Aggregation Block (CCAB), and Adaptive Context Aggregation Block (ACAB). DRB mitigates context loss by incorporating context at the topmost level and layering residual blocks with varying dilation rates to produce a more accurate representation. By fusing context information from adjacent levels in an efficient manner, CCAB enables interactive fusion to suppress noise and improve features. By calculating channel and spatial weights, ACAB bridges semantic gaps and utilizes Spatial-guided Aggregation Block (SAB) and Channel-guided Aggregation Block (CAB) to construct a balanced global context. MCFPN adaptively incorporates spatial and channel-specific context features, enhancing object detection accuracy across a range of visual tasks and object scales.

(12) Few Relevant Neighbors (FNM) (Barnea and Ben-Shahar 2019) focuses on local spatial context between objects (higher-order relations) by tackling the challenge of learning when objects collide, particularly during context-detector conflicts. A belief propagation mechanism has been utilized to integrate spatial relations between objects. This mechanism is used to calculate context-based probabilities for objects, dynamically selecting the most informative collection of context variables for each location. Furthermore, for utilizing scale context, it uses scale-invariant representations, which reduce the requirement for varied instances at multiple sizes and simplify training. FNM not only has the greatest AP among approaches in multi-scale category but also outperforms all other general object detection methods on the COCO dataset.

(13) MSF (Wang et al. 2018) leverages multi-scale fusion to enhance context awareness, synthesizing information across multiple resolutions and spatial regions to improve object detection accuracy. By combining context from surrounding regions and incorporating it directly into object detection layers, MSF effectively addresses scenarios with small or partially occluded objects, enhancing both precision and recall in cluttered or complex scenes.

(14) Efficient Selective Context Network (ESCNet) (Nie et al. 2020) utilizes multi-scale context by enhancing feature pyramids and implementing selective attention to refine critical details. The ECM enriches shallow layers with multi-scale information, which is essential for accurate small object detection. The TAM's selective attention further leverages context by filtering and amplifying relevant features, enhancing the network's ability to focus on critical parts of the image while minimizing noise. Through these components, ESCNet successfully integrates spatial, channel, and global contexts, resulting in improved localization and classification, particularly in scenes with complex backgrounds and small, detailed objects.

(15) Pyramid context learning (PCL) (Ding et al. 2020) employs a structured multi-level context extraction process where aggregation operator collects features at various spatial scales, and distribution operator adaptively weights these features based on their contextual importance. This ensures that both global and local context is harnessed, allowing the model to detect complex object arrangements by providing context-aware features across scales. Channel context learning further contributes by refining feature maps through capturing correlations among channels, enhancing the model's ability to focus on critical aspects of the object, which improves accuracy in diverse and cluttered environments.

(16) The UGC-YOLO (Yang et al. 2023) enhances underwater object detection by integrating global context information with the YOLOv3 architecture. It employs deformable convolution to adaptively capture features of various aquatic organisms and differentiates between overlapping objects or those that blend into the background, while also utilizing a pyramid pooling module to aggregate semantic information at different scales.

To provide a clearer evaluation, the following comparison highlights the key strengths and limitations of these multi-scale approaches, focusing on their adaptability, efficiency, and task-specific applications.

**Adaptability to context variability:** Dilated and Deformable FPN (DDFPN) and CL-FPN excel in capturing diverse spatial and semantic features with dilated convolutions and attention modules, benefiting complex scenes with varying object sizes. However, they come with increased memory and computational requirements, which limit their real-time usability. By comparison, CEBNet and CP-SSD offer a lightweight solution with enhanced low-level features, though they may not be as effective for very small objects or in densely packed scenes, where more complex methods like MCFPN and Global Context Aware (GCA) RCNN perform better.

**Computational efficiency:** GCE and ESCNet integrate context effectively without heavy computational demands, ideal for applications needing fast inference. GCE's efficiency drops at high IoU thresholds, while ESCNet still faces background confusion in cluttered scenes. CP-SSD and MSF balance speed and context retention, with CP-SSD's complexity making it robust for object differentiation and MSF's simpler fusion benefiting smaller objects while limiting its reach for larger objects.

**Task-specific strengths:** UGC-YOLO and FA-FPN-MCI, designed for underwater and domain-generalized detection, capture fine details and handle specific environmental challenges. Similarly, CEBNet and YOLOC cater to urban scenes, with CEBNet enhancing small object detection in complex backgrounds and YOLOC modeling context for traffic-specific scenarios, though large-scale improvements are minimal.

In summary, while models like DDFPN and CL-FPN stand out for their contextual richness, their efficiency may hinder real-time application. Streamlined architectures such as FNM and CP-SSD offer better real-time performance but may lack nuanced context handling needed for certain challenging scenarios. Multi-scale approaches continue to demonstrate clear advancements in integrating context across scales, underscoring a progression toward models that can balance efficiency with increasingly complex contextual needs.

### 4.2.5 RPN-based approaches

Region Proposal Network (RPN) (Ren et al. 2016) approaches suggest regions likely containing objects, allowing for more precise object detection in later stages. Figure 37 illustrates a typical RPN pipeline, where the network generates several region proposals based on feature maps extracted from CNN layers. These proposals are then refined through processes like RoI pooling and classification, ultimately identifying the regions most likely to contain objects. RPN-based context approaches have incorporated contextual information to RPN to enhance the accuracy and relevance of these proposals.

In this section, we review four RPN-based context approaches.

(1) Cascade region proposal (Zhong et al. 2020) benefits context by modeling both local and global context. This helps in refining region proposals based on surrounding features, leading to more accurate object detection. The model's global context branch also leverages entire image features, which supports object classification by considering broader contextual cues beyond just the object's immediate area.
(2) Learning based context refinement (Chen et al. 2018) enriches region proposals by iteratively integrating visual, spatial, and semantic context from neighboring regions. This comprehensive use of context improves both proposal localization and classification by ensuring that each object's context is thoroughly examined, making it effective for scenes with dense object interactions. The adaptive weighting strategy prioritizes more relevant context, reducing the impact of irrelevant surroundings, thereby enhancing the model's ability to discriminate between objects and their backgrounds. This approach has achieved superior results compared to other PRN-based models in testing on the COCO dataset.
(3) SC-Faster RCNN (Xiao et al. 2020) model leverages contextual information by embedding a feature extraction module for context after the conv5_3 layer. This module, combined with skip pooling, helps the network better understand the spatial and semantic context around objects. As a result, it can differentiate between similar-looking objects and the background, especially in cluttered or complex scenes, making it effective for detecting partially hidden objects and recognizing objects based on surrounding cues. SC-Faster R-CNN surpasses proposal-based object detection (Kaya and Alatan 2018) in detecting small objects.
(4) The model (Kaya and Alatan 2018) leverages context by adding a dedicated context feature extractor stage, which operates in parallel to the object feature extraction layers in Faster R-CNN. By pooling features from a surrounding "context ring" around the object and merging them with object features via the wrap-around operation, the model captures spatially consistent context information. This spatially aware combination allows the model to discern object boundaries more effectively, especially in cases where context is a key factor in recognizing the extent or class of an object.

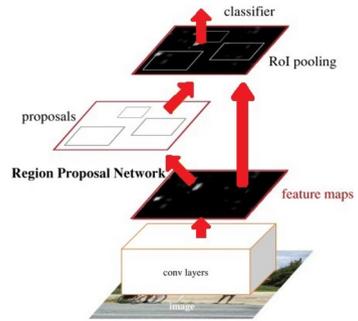

**Fig. 37** Using RPN to propose regions containing objects

To provide a more analytical perspective, we compare these approaches to highlight their respective advantages, limitations, and contextual integration strategies. The cascade RPN model emphasizes global context modeling to improve proposal quality without significant computational overhead, making it efficient but limited in handling small objects. Conversely, the learning-based context refinement approach excels in crowded scenes by iteratively refining spatial and semantic context, enhancing boundary precision. However, it introduces additional computational costs and relies heavily on initial proposal quality, making it less effective in sparse contexts. The SC-Faster R-CNN builds upon these advancements by incorporating skip pooling and guided anchors to improve small and occluded object detection, although it faces challenges with camouflaged or heavily deformed objects. Meanwhile, (Kaya and Alatan 2018) approach offers enhanced boundary determination by preserving spatial relationships between object and context features, proving effective for context-rich classes but adding model complexity and showing limitations in heavily occluded settings. Overall, while each model improves contextual feature extraction, they vary in computational efficiency and suitability for specific object detection challenges, illustrating a progressive refinement of contextual integration tailored to distinct scene complexities and object types.

### 4.2.6 Attention-based approaches

Attention-based approaches leverage mechanisms that focus on relevant features while suppressing irrelevant background information, improving detection performance by emphasizing the most critical context. Figure 38 illustrates in (a) "Without Attention," the model assigns equal importance to surrounding regions, leading to a misclassification where a chair is incorrectly identified as a sofa. In (b) "With Attention," the model applies attention weights, prioritizing relevant contextual features, which enables it to correctly identify the object as a chair. This highlights how attention mechanisms improve object detection by helping models concentrate on the most informative parts of the scene, thus enhancing accuracy.

Attention mechanisms have been used in some approaches in the form of different modules to integrate contextual information. Fourteen papers are reviewed in this category.

(1) Pure regression object detection (Fan et al. 2022) utilizes context by embedding a lightweight Contextual Attention Block (CAB) that enhances the representation of regression points through global contextual information. The CAB allows the model to capture extensive context around objects, addressing the limitations of traditional bounding box-based representations by considering both the spatial extent of objects and their surrounding context. This integration of context information leads to improved localization and classification of objects, particularly in challenging environments where context plays a key role in distinguishing similar objects.
(2) Sparse attention block (SA) (Chen et al. 2022) selectively focuses on high-response areas, such as object edges, to capture meaningful spatial context in long-range dependencies. By aggregating information from only the most relevant positions, SA minimizes the computational and memory burden associated with dense pixel interactions.

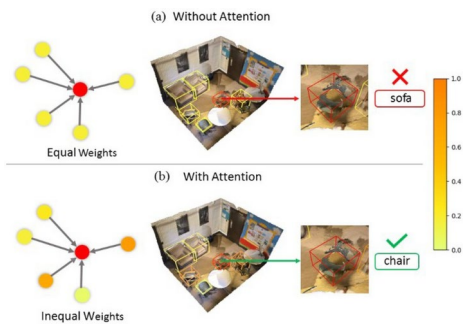

**Fig. 38** Impact of attention mechanisms on object detection accuracy (Lan et al. 2022)

This focused attention reduces background noise and redundancy, ensuring that the included context supports accurate object detection without overwhelming resources, especially in scenarios where distinguishing objects from their surroundings requires effective long-range dependency modeling.

(3) Global Context (GC) module (Lee et al. 2021) leverages self-attention to enhance object detection by blending global and local context in the feature maps. By computing relationships among all spatial elements, it allows each element to incorporate information from distant regions, aiding in the identification of ambiguous objects based on their spatial and semantic relationships. This tailored integration of self-attention supports object distinction in complex scenes.

(4) CSA-Net (Liang et al. 2022) is superior to all other attention-based techniques on the COCO dataset. CSA-Net leverages context through a self-attention mechanism in the ResNet-SA (He et al. 2016), which emphasizes object regions while downplaying background noise. The RFFE module further enhances context utilization by integrating global and local features across multiple receptive fields, providing richer information for detecting both small and large objects. The SFFP network complements this by enabling multi-scale feature fusion, ensuring that features from different layers contribute to accurate object recognition. Together, these components allow CSA-Net to utilize spatial context effectively, particularly in scenes with varied object sizes and complex backgrounds.

(5) Global Contextual Dependency Network (GCDN) (Li et al. 2022) improves two-stage object detectors by enhancing global contextual information. It combines global and local information to strengthen Region of Interest (RoI) feature representation. The Context Representation Module (CRM) provides multi-scale context, capturing relationships across scales, while the Context Dependency Modules (CDMs) use attention to refine these representations. This integration of spatial and global context improves object classification, especially for objects in cluttered backgrounds or with minimal visual features.

(6) In Cross-context Attention-guided Network (CCAGNet) (Miao et al. 2022), to emphasize object-synergy regions and suppress non-object-synergy regions, CCAGNet integrates three attention mechanisms: Cross-context Attention Mechanism (CCAM), Semantic Fusion Attention Mechanism (SFAM), and Receptive Field Attention Mechanism (RFAM). The CCAM assists the model in focusing on significant parts of an image by highlighting relevant regions and dismissing less important ones. The SFAM optimizes upsampling by emphasizing valuable information and minimizing noise in feature representation. The RFAM increases the model's awareness of the context of an image, helping it to better comprehend the relations between distinct features. This network reduces false positives in cluttered or ambiguous environments.

(7) Context-based Feature Fusion Network (CFFN) (Xia et al. 2020)overcomes performance restrictions caused by a lack of global context and the presence of background noise in low-level features. Context Extraction Module (CEM) and Context Refinement Module (CRM), are suggested to augment the network's capacity to acquire abstract global context information and boost the discriminative capability of low-level feature representations through the utilization of an attention mechanism.

(8) AMCM+YOLOv4 (Ma et al. 2022) enhances the understanding of spatial relationships and dependencies within occluded and multi-scale objects. In this framework,

YOLOv4 is enhanced by using attention-guided multi-scale context information module (AMCM) to improve the perception of object context information. The approach begins by performing feature extraction via the CSPDarkNet53 network. Subsequently, it employs the AMCM technique to enhance feature discrimination by applying attention weighting.

(9) Lightened Context Extraction Module (LCEM) (Jiaxuan et al. 2022) employs superimposed dilation convolutions inside the Feature Pyramid Network (FPN) to provide efficient feature fusion across various scales. In addition, this module focuses on the attention mechanism in the Attention-guided Context Feature Pyramid Network (ACFP), specifically stressing the enhancement of feature map integration via dilation convolutions. It is well-suited for real-time applications that require context-rich object detection.

(10) CIE-JHR's combination of attention and convolution layers (Zhao et al. 2022) enhances context extraction, addressing challenges in identifying small objects by incorporating global–local spatial relationships. This enriched feature representation supports applications where accurate object detection amidst clutter or obscured backgrounds is critical, as in the detection of transmission line components.

(11) Improved YOLOv8 (Saha et al. 2024) leverages contextual cues from its dataset, applying attention mechanisms and the C2f module to adapt detection to localized license plate characteristics. These mechanisms help manage the model's focus, enabling it to distinguish among different vehicle types and environmental variations specific to Bangladesh, such as varying license plate colors and formats in diverse lighting conditions. This integration strengthens the model's adaptability and precision in Automatic License Plate Recognition (ALPR) systems.

(12) YOLOv8-CGRNet (Niu et al. 2023) is a lightweight object detection network optimized for mobile devices. By integrating YOLOv8 with Context GuidedNet (CGNet) and Res2Net, it enhances feature learning while maintaining low computational complexity. The model effectively captures local features and spatial dependencies, improving contextual understanding. Additionally, it employs a pyramid network and a dynamic focusing mechanism to handle low-quality examples. This is valuable for scenarios with complex backgrounds or object variances, where accurate context integration can significantly improve detection robustness.

(13) The presented approach in (Ma and Wang 2023), is a multi-scale model that focuses on fusing high-resolution spatial features with deep semantic features. To extract global context, a cross-pooling captures long-range dependencies across the image, which enhances the model's ability to detect objects within complex backgrounds or scenes. Local context can be extracted via a cascaded deformable context module that integrates spatial information from surrounding regions, assisting in detecting objects amidst clutter and varied scales.

(14) The Fine-Grained Dual Level Attention Mechanism joint Spatial Context Information Fusion (FGDLAM and SCIF) (Deng et al. 2024) enhances object detection by refining traditional attention mechanisms, subdividing the feature space into multiple subspaces for precise channel weight extraction while employing a dual-weight strategy to capture relationships both within channels and across spatial regions. Additionally, it integrates a context information extraction module to leverage local and global contextual information, significantly improving object recognition and localization.

A more analytical comparison of these attention-based approaches highlights clear trade-offs between accuracy, computational efficiency, and context utilization. Lightweight models like Pure Regression Object Detection and Sparse Attention Block (SA) are efficient and suitable for real-time applications, focusing on key spatial areas to reduce noise. However, their selective use of spatial context risks losing detail in cluttered scenes, where more comprehensive context may be necessary. Global Context (GC) and Global Contextual Dependency Network (GCDN) improve small and occluded object detection by incorporating extensive global context, though at the cost of increased memory and processing demands, limiting their real-time usability. Balanced models like CSA-Net and CFFN blend local and global context, offering robust multi-scale detection. However, their fusion mechanisms increase complexity, making them less ideal for fast applications. Meanwhile, FGDLAM&SCIF enhances spatial-channel correlations through fine-grained attention, capturing nuanced context across scales, though it adds computational overhead and potential isolation issues that require added processing. In summary, while each approach enhances context-driven detection, they differ in balancing accuracy, speed, and adaptability. These trade-offs reflect a progression in the field, with each method advancing tailored solutions for specific object detection challenges.

### 4.2.7 Other approaches

In the "Other Approaches" category, these models each enhance object detection by leveraging unique forms of context, each with distinct strengths and trade-offs.

(1) Context Learning Network (CLN) (Leng et al. 2018) captures the pairwise relations between objects and the global context. The network is divided into two subnetworks: a Multi-Layer Perceptron (MLP) and a Convolutional Neural Network (ConvNet). The MLP is first used to capture pairwise relations. The ConvNet then gathers and concatenates pairwise relations to learn more about the global context.
(2) One-stage Diverse Receptive Field Network (DRFNet) (Xie et al. 2020) utilizes multi-branch diverse receptive field modules (DRF modules) and a parallel framework to collect contextual information at various sizes.
(3) EGCI-Net (Guo et al. 2020) integrates enhanced global context information through global activation blocks into its backbone. This integration aims to decrease the reliance on local information and increase the global context to tackles the constraints identified in DSOD (Shen et al. 2017). Moreover, a pyramid feature pool module produces multi-scale global context features, guiding the detection process.
(4) In Recursive Context Routing mechanism (ReCoR) (Chen et al. 2021), spatial modeling captures spatial relationships over longer distances via a recursive structure, and channel-wise modeling, which encodes relationships between features for a better understanding of context. The technique dynamically models contexts by integrating spatial relations with channel-wise representations of local items.
(5) MCENet (Wang and Ma 2022) employs rectangle pooling kernels (RPU) to extract and use image level long-range relationships. To capture multi-scale contextual information, dilated convolutions with varying dilation rates are used in addition to image level context. Finally, RPU and dilated convolutions (DC) are merged into a context enhancement module (CEM), which can be used to increase detection accuracy in different models.

(6) Boundary Aware Network (BAN) (Kim et al. 2018) emphasizes the importance of boundary context, which is a subset of spatial context. BAN defines three types of boundary contexts (side, vertex, in/out-boundary) and employs 10 sub-networks to represent the relationship between these contexts. The detection head of BAN is an ensemble of these sub-networks, selectively focusing on different contributions based on the detection sub-problem to improve object detection accuracy.

(7) CompositionalNet (Wang et al. 2020) comprises dissecting image representation into context and object components during training, which is accomplished using context segmentation. This network efficiently manages the impact of context, boosting robustness in identifying heavily occluded objects.

(8) Layout Transfer Network (LTN) (Wang et al. 2019) uses a retrieve-and-transform technique to forecast likely objects' positions and sizes. This technique integrates both bottom-up and top-down visual processing into Faster RCNN for combined reasoning of object detection and scene layout estimation.

(9) Edge-aware Context-aggregation Network (ECNet) (Xiao et al. 2021) detects transparent and reflective objects such as glass products. ED module extracts boundary features from images, serving as a global attention mechanism. DFF module extracts context discontinuity from input depth maps that eases textual feature extraction by fusing into the RGB backbone at each level. Finally, MFE module utilizes multi-receptive-field features to expose the discontinuity in texture.

(10) Of-The-Shelf (Bardool et al. 2019) uses an of-the-shelf Mask R-CNN to generate feature maps and detect objects. In the next step, a Fully Convolutional Network (FCN) (Shelhamer et al. 2017) is employed as a contextual learner to comprehend semantic relationships between objects using contextual feature maps.

(11) In POD-F and POD-Y methods (Ma et al. 2023), a learnable Gabor convolution layer and a Spatial Attention (SA) mechanism are presented for low-level features to collect edge and contour information while improving spatial relationships. For high-level features a Global Context Feature Extraction (GCFE) module is used to extract multi-scale global contextual information, and a Dual Scale Feature Aggregation (DSFA) module to fuse features from different scales.

(12) A trainable spatial context features extractor (SCFE) (Wang et al. 2018) inspired by recurrent neural networks (RNN) was presented for fast object detection by augmenting convolutional neural networks (Wang et al. 2018). The SCFE, as opposed to conventional CNN-based approaches that prioritize local geometric and texture features, extracts spatial context information directly from the scene.

(13) To address the lack of contextual information surrounding R-CNN object proposals, an object detection system (Chu and Cai 2018) was introduced that integrates local appearance and contextual information. It makes use of a fully connected CRF formed on object proposals, with contextual restrictions incorporated as edges. The system incorporates both local interactions between objects and global scene information, employing a logistic regression model to comprehend the scene.

(14) In (Zeng et al. 2021), the approach addresses two challenges: local optima and object ambiguity. Multiple Instance Learning (MIL) with self-training is utilized by the proposed method in order to generate pseudo-labels and enhance object localization throughout training. In order to surpass local optima, a context awareness block is incorporated to focus on background and context. The approach employs a spatial pyramid pooling (SPP) network layer to boost generalization and detection performance while avoiding local optima.

(15) Paper (Gu et al. 2022) addresses WSOD with image-level annotations, highlighting the frequent difficulty of localizing discriminative parts rather than the full object. To improve object localization accuracy, the proposed technique employs a Symmetry Context Module (SCM) and context proposal mining strategies. The SCM uses contextual information from precomputed region proposals to encourage the model to prioritize proposals that include the whole object. Furthermore, to gather context appearance in different spatial areas surrounding, two context proposal mining strategies, naive and Gaussian-based are implemented.

(16) The Transformer-based Context Condensation (TCC) (Chen et al. 2023) improves multi-level feature fusion (MFF) in feature pyramids. It decomposes context information into locally concentrated and globally summarized representations, then uses a Transformer decoder to refine MFF results by exploring relationships between local features and condensed contexts. This method boosts detection accuracy while reducing computational costs.

Other approaches offer diverse strategies to integrate context for object detection, each with unique advantages and challenges. Models like CLN, EGCI-Net, and ReCoR improve detection by emphasizing global context, which enhances accuracy in complex scenes but can increase computational load and reduce adaptability in simpler applications. DRFNet and MCENet focus on multi-scale context extraction, excelling in size-variant object detection but often at the cost of processing speed due to their added structures. BAN and CompositionalNet prioritize boundary and compositional contexts, aiding accuracy for small or occluded objects, though they may miss finer contextual details in cluttered scenes. LTN and ECNet handle specific detection needs, like scene layout forecasting and transparent objects, but can struggle with unpredictable or low-contrast environments. POD-F, POD-Y, and TCC improve object clarity in complex scenes with attention and multi-scale context fusion, though these techniques add computational demands. Lastly, SCFE, CRF-based Object Detection, and Adaptive MIL enhance spatial-semantic coherence for small object detection, albeit with limitations in real-time applications. In sum, these methods advance object detection by addressing unique context challenges, balancing trade-offs between efficiency, accuracy, and adaptability across detection tasks.

### 4.2.8 Results on general object detection

Based on Fig. 31, approaches highlighted in red have been specifically designed to improve two-stage object detectors. The most effective two-stage approach tested on COCO dataset is GCA RCNN (Zhang et al. 2021), which mitigates the loss of contextual information during the resizing of object proposals by integrating a context-aware mechanism that extracts and refines features using a global feature pyramid and attention techniques, respectively. By preserving spatial relationships between the foreground and background, GCA RCNN enhances detection performance across varied object scales and positions. Its context-aware mechanism refines object features with high accuracy, making it effec-tive in COCO's complex scenes. Furthermore, YOLOv4-AMCM demonstrated superior performance on the PASCAL VOC07 dataset by incorporating an attention-based module (AMCM), which emphasizes feature discrimination by applying context-sensitive attention weighting. AMCM allows the model to focus on the most relevant spatial relationships, making YOLOv4-AMCM robust in scenarios with complex object overlap. By improving the perception of object context, YOLOv4-AMCM surpasses other models in the VOC07 dataset by enhancing detection accuracy in cluttered and occlusion-heavy environments. In Fig. 31, approaches highlighted in blue have been designed merely for one-stage models. CCAGNet (Miao et al. 2022), which integrates contextual information using three attention mechanisms, including CCAM, SFAM, and RFAM, is the top network in the one-stage category tested on Pascal 2007 dataset. It effectively focuses on relevant regions while ignoring less important areas. CCAM enhances focus on object-synergy areas, while SFAM reduces noise in feature representation by emphasizing valuable features during upsampling. RFAM further improves the model's contextual awareness by understanding relationships between distinct features. Together, these attention mechanisms help CCAGNet reduce false positives, making it one of the best one-stage detectors for VOC07 by leveraging attention-guided context to avoid distractions in dense scenes. Moreover, this network surpasses other one-stage methods in detecting small objects. FGDLAM and SCIF is another one-stage model that surpasses other one-stage models tested on the COCO dataset in terms of mAP due to the inclusion of dual-level attention mechanism, which divides the feature space into subspaces for fine-grained weight extraction and captures relationships within and across spatial regions. This dual-weight strategy enables the model to capture both local and global context effectively, crucial in COCO's diverse and high-density scenes (Table 4).

Throughout the paper, bolded models in the tables denote the highest mAP achieved, emphasizing the best performance among all compared methods in the one specific dataset.

Based on Fig. 40, ResNet and VGG16 as backbones, Faster RCNN and RetinaNet architectures, and spatial, scale, and semantic context types are mostly used in general object detection. Faster R-CNN, Cascade Mask RCNN, SSD, and YOLOv4 architectures have been used in the mentioned top-performing networks with the highest mAP.

As shown in Tables 5 and 6, on the COCO dataset, FNM has the highest AP, FGD-LAM and SCIF is ranked second, and the Feature Refinement is positioned as the third-best model. FNM excels in the COCO dataset due to its belief propagation mechanism, which manages spatial relations between objects in a dynamic and context-sensitive manner. This mechanism calculates context-based probabilities, dynamically selecting relevant context variables for each location. The use of scale-invariant representations is another key module that reduces the need for multi-scale instances, simplifying the training process while retain-ing accuracy across various object scales. These innovations allow FNM to surpass other methods by effectively leveraging both spatial and scale context in highly variable scenes. Feature Refinement surpasses other models on VOC07 and for medium-to-large objects (APm and APl) due to its cross-pooling and cascaded deformable context modules. The cross-pooling module is specifically designed to capture long-range dependencies, enabling the model to leverage global context efficiently. The deformable context module further refines local context by incorporating spatial details from surrounding regions, making it adept at identifying medium-to-large objects even when they are partially obscured or surrounded by clutter. These context-aware modules allow Feature Refinement to outperform others by precisely balancing global and local context in dense or visually noisy environments. GCE has the highest accuracy for recognizing small objects (APs), a challenging task due to their tendency to be overlooked in cluttered scenes. Its dual-path approach leverages top-down, image-level classifications combined with regional features, which enables the model to utilize high-level contextual cues to locate small objects accurately. This global context encoding enriches the model's understanding of the full scene, allowing it to capture subtle semantic relationships and recognize small objects even when surrounded by larger, more prominent objects. Among all methods evaluated on the PASCAL VOC 2007 and 2012 datasets, Cascade Region Proposal demonstrates the highest level of accuracy for PASCAL 2012 because of its multi-stage refinement process that integrates both local and global context. The model's global context branch extracts entire image features, which enables it to refine region proposals based on surrounding context, leading to more accurate localization. This broader view beyond the immediate object vicinity is particularly advantageous in VOC12, where understanding the scene context as a whole supports better classification. The progressive refinement steps in Cascade Region Proposal ensure precise object boundaries and improve detection accuracy, especially in scenes where objects might otherwise be misclassified due to similar features. Conversely, for PASCAL 2007, Feature Refinement emerges as the most optimal model. The performance of other methods on different datasets has been demonstrated in Table 7. The best performing methods for different tasks are shown in Table 4.

Table 4 Best context-based GOD methods

| Description | Method |
| --- | --- |
| Best method for COCO | FNM |
| Best method for VOC07 | Feature Refinement |
| Best method for VOC12 | Cascade Region Proposal |
| Best method for medium and large objects | Feature Refinement |
| Best method for small objects | GCE |
| Best one-stage detector | CCAGNet (Pascal07), FGDLAM and SCIF (COCO) |
| Best two-stage detector | GCA RCNN (COCO dataset) |

In summary, the combination of context-aware modules across two-stage and one-stage detectors, especially through the use of attention and multi-scale approaches, consistently boosts object detection performance across different datasets. Approaches like GCA-RCNN and FGDLAM show how integrating global and local context improves accuracy across object sizes, while models like Feature Refinement excel in domain-specific tasks. This highlights the critical role context plays in advancing the field of object detection.

Table 8 comprehensively outlines the challenges addressed by general object detection approaches, encompassing issues such as scale variations, object occlusion, background complexity, and environmental factors. These challenges highlight how context-based models are capable of addressing obstacles that context-free approaches often struggle to overcome, showcasing the advantages of integrating contextual information for improved object detection performance.

### 4.3  Small object detection (SOD)

Small object detection focuses specifically on detecting and localizing objects that are small in size. Small objects fill areas less than and equal to $32 \times 32$ pixels (Lin et al. 2014). As illustrated in Fig. 39, the presence of fewer pixels in small objects within an image hinders the network's capacity to extract significant features from them (Tong et al. 2020). Contextual information provides additional details about objects and their surroundings, aiding the network in detecting small objects more effectively. In this section, context-based small object detection approaches are reviewed (Fig. 40).

(1) FA-SSD (Lim et al. 2021) is the result of incorporating two modules: the Context by Feature Fusion module (F-SSD) and the Attention Mechanism (A-SSD). F-SSD utilizes the concatenation of multi-scale features from adjacent pixels to extract context information, thereby enabling a more comprehensive depiction of small objects. A-SSD contains an attention mechanism early in the layer, allowing for focused detection by eliminating extraneous background information.
(2) FPN with CEM and FPM (CEFP2N) (Xiao, Guo, et al. 2023) is a new feature pyramid composite neural network structure. It has two main modules: Context Enhancement Module (CEM) and Feature Purification Module (FPM). CEM enhances context information using multi-scale dilated convolution features, whereas FPM uses feature purification procedures to remove conflicting information in multi-scale feature fusion.
(3) Internal-External Network (IENet) (Leng et al. 2021) utilizes both appearance and context information. This network contains three modules: Bidirectional Feature Fusion Module (Bi-FFM), Context Reasoning Module (CRM), and Context Feature Augmentation Module (CFAM). Bi-FFM collects internal features of objects, proposal quality is enhanced via context reasoning in CRM, and classification is achieved by CFAM through the learning of pair-wise relations between region proposals produced by CRM.
(4) Improved YOLOv5 (Zhang et al. 2022) integrates Coordinate Attention (CA) and a Context Feature Enhancement Module (CFEM) into the YOLOv5 network. CA incorporates positional information into channel attention, whereas CFEM extracts rich context information from multiple receptive fields.
(5) CGA-YOLO (Hang et al. 2022) utilizes a Swin Transformer-based context information extraction module. In the feature fusion network, it employs a Global Attention Mechanism (GAM) comprising Channel Attention Module (CAM) and Spatial Attention Module (SAM) to enhance feature representation, especially for small object detection.
(6) TYOLOv5 (Corsel et al. 2023) is a spatio-temporal model that utilizes temporal context extracted from video sequences to enhance the recognition of small moving objects while maintaining the accuracy of detecting stationary objects. The main contributions consist of integrating temporal data augmentation, introducing a spatio-temporal detector that operates on a single stream, and proposing a two-stream architecture that utilizes frame differencing to capture explicit motion information.
(7) Contextual-YOLOV3 (Luo et al. 2019) incorporates a Contextual Relationship Matrix (CRM) into YOLOv3's classification probability to enhance classification probability. Furthermore, a Context-Based Filtering Algorithm replaces the traditional non-maximum suppression algorithm for optimal window selection.
(8) Context-Aware Block Net (CAB Net) (Cui et al. 2020) addresses losing spatial information and low accuracy due to gradual downsampling by implementing a Context-Aware Block (CAB) that incorporates pyramidal dilated convolutions while maintaining spatial relationships. By preserving both detailed and high-level semantic features,

**Table 5** General object detection (COCO dataset)

| Detector name | Context type/level | Backbone/architecture | Mechanism/module | mAP | APs | APm | APl |
|---|---|---|---|---|---|---|---|
| **FNM** Barnea and Ben-Shahar (2019) | **Spatial,Scale/Local** | **Faster R-CNN** | **Scale-invariant Representations** | **66.9** | – | – | – |
| FGDLAM and SCIF Deng et al. (2024) | Spatial/Local,Global | CSPv5/YOLOx | FGDLAM joint SCIF | 54.2 | – | – | – |
| Feature refinement Ma and Wang (2023) | Spatial/Local,Global | EfficientNet-B6 | Cross-Pooling,Cascaded Context | 53.0 | 35.4 | 57.1 | 65.3 |
| ReCoR Chen et al. (2021) | Spatial/Global | ResNeXt/Cascade Mask RCNN | Recursive Context Routing | 51.6 | - | – | – |
| CSA-Net Liang et al. (2022) | Scale,Spatial,Semantic/Local,Global | ResNet | Attention,SFFP,RFFE | 46.8 | 27.2 | 50.2 | 61.3 |
| MCENet(FCOS+CEM) Wang and Ma (2022) | Scale,Spatial/Local,Global | ResNeXt | RPU,CEM | 46.6 | 28.9 | 49.5 | 58.3 |
| HCE Chen et al. (2020) | Spatial,Semantic/Global | ResNet/Cascade RCNN | Early-and-Late | 46.5 | 27.4 | 49.9 | 59.4 |
| Distilling Graph Yang et al. (2023) | Spatial,Semantic/Local,Global | ResNet | KD,Graph attention | 46.3 | 27.8 | 50.1 | 57.1 |
| DDFPN Wu et al. (2021) | Spatial,Semantic/Local,Global | ResNet/Cascade R-CNN | DDC,CFC,CI | 45.3 | 26.7 | 48.3 | 57.9 |
| TCC-DDETR Chen et al. (2023) | Spatial/Local,Global | ResNet/DETR | TCC | 45.0 | – | – | – |
| Pure Regression Fan et al. (2022) | Scale,Spatial/Global | ResNet | CAB,TFM | 44.9 | 29.6 | 47.4 | 45.5 |
| PCL Ding et al. (2020) | Local,Global | VGG16/Faster R-CNN | PCL module | 44.4 | 28.6 | 45.3 | 54.4 |
| MCFPN Wang et al. (2022) | Scale,Semantic/global | ResNet/Faster RCNN | DRB,CCAB,ACAB | 44.3 | 26.5 | 47.7 | 55.6 |
| CL-FPN Yang et al. (2023) | Scale,Semantic,Spatial/Local,Global | ResNeXt/Faster RCNN | CEM,AFRM,MLF | 43.3 | 25.7 | 46.6 | 53.8 |
| FRCNN+Seg+HCE Qiu et al. (2020) | Scale/Local,Global | ResNet/Mask RCNN | Hierarchical Context,GED | 43.0 | 24.5 | 45.5 | 55.1 |
| MGFPN Li et al. (2022) | Scale/Global | ResNext/RetinaNet | GCNet,MSGC | 42.8 | – | – | – |
| KROD Ji et al. (2022) | Semantic/Prior Knowledge,Global | ResNeXt/RetinaNet | GCN | 42.8 | 25.3 | 45.9 | 53.8 |
| GCA RCNN Zhang et al. (2021) | Spatial,Scale/Global | ResNet/RCNN | Global Pyramid,Attention | 42.1 | 24.4 | 45.2 | 53.2 |
| Context Refinement Chen et al. (2018) | Semantic,Spatial/Local | ResNet/C-Mask RCNN | Adaptive Weighting | 42.0 | 23.4 | 44.7 | 53.8 |
| SA Chen et al. (2022) | Spatial,Semantic/Global | ResNet/Faster R-CNN | SA | 41.3 | 23.7 | 44.8 | 51.7 |
| GCE Peng et al. (2022) | Semantic,Scale/Global | ResNet/Faster+Cascade | GCE | 41.2 | 44.7 | – | 55.2 |
| GCDN Li et al. (2022) | Scale,Spatial/Global | ResNet/Mask R-CNN | Attention | 41.0 | 24.0 | 45.1 | 53.7 |
| CB-FPN Liu and Cheng (2023) | Spatial,Semantic/Local,Global | ResNet-50/Faster R-CNN | CEM-CSP, BE-FPN | 39.2 | 22.8 | 42.8 | 51.0 |
| GC module Lee et al. (2021) | Spatial/Global | ResNet101/RetinaNet | Global Self Attention | 38.4 | – | – | – |
| Deformable R-FCN-BAN Kim et al. (2018) | Spatial/Local | ResNet/R-FCN | BAN | 36.9 | 15.8 | 40.0 | 53.6 |

Table 5 (continued)

| Detector name | Context type/level | Backbone/architecture | Mechanism/module | mAP | APs | APm | APl |
|---|---|---|---|---|---|---|---|
| ESCNet Nie et al. (2020) | Scale,Spatial/Global | ResNet101/SSD | TAM,CGA,SA,ECM | 36.9 | 15.8 | 41.5 | 53.8 |
| Cascade Region Proposal Zhong et al. (2020) | Spatial/Local,Global | ResNet/Faster R-CNN | New RPN | 36.8 | – | – | – |
| JLWSOD Lai et al. (2024) | Semantic/Local | ResNet-50 | iGCL,Instance Branch, Semantic Branch | 35.9 | – | – | – |
| CCAGNet Miao et al. (2022) | Spatial,Semantic,Scale/Local,Global | VGG16/SSD | CCAM,RFAM,SFAM | 35.6 | 27.9 | 52.6 | 64.5 |
| CompositionalNet Wang et al. (2020) | Spatial/Local | CompositionalNets | Part-Based Voting | 35.0 | – | – | – |
| CLN Leng et al. (2018) | Semantic/Local,Global | VGG16/Faster-RCNN | MLP, ConvNet | 34.8 | 18.4 | 41.0 | 54.6 |
| CFFN Xia et al. (2020) | Scale,Spatial,Semantic/Global | ResNet-101/FSSD | Attention,CRM,CEM | 34.4 | 18.0 | 39.0 | 51.1 |
| DRFNet Xie et al. (2020) | Spatial/Local | VGG-16 | Context Aggregation, DRF | 33.5 | 15.2 | 37.4 | 47.6 |
| EGCI-Net Guo et al. (2020) | Spatial/Global | DS/64-192-48-1 | Global blocks,PFPM, PPM | 31.9 | – | – | – |
| AdaCon Neseem and Reda 2021 | Spatial,Semantic/Prior,Local,Global | Darknet/YOLOv3 | Adaptive,Graph | 30.1 | – | – | – |
| SIN Liu et al. (2018) | Spatial,Semantic/Local, Global | VGG-16/Faster R-CNN | Message Passing,Scene,Edge | 23.2 | 7.3 | 24.5 | 36.3 |
| Multi-scale Fusion Wang et al. (2018) | Spatial,Semantic,Scale/Global,Local | VGG16 | - | 23.2 | – | – | – |
| CEBNet Chen et al. (2019) | Scale,Spatial/Local | Modified VGG16/SSD | ERFB,FAB,CEBs | - | 16.3 | 37.5 | 48.5 |

Table 6 General object detection (PASCAL VOC 2007 and 2012)

| Detector name | Context type/level | Backbone/Architecture | Mechanism/Module | Dataset | mAP |
|---|---|---|---|---|---|
| **Cascade region proposal** (Zhong et al. (2020)) | **Spatial/Local,Global** | **ResNet/Faster R-CNN** | **New RPN** | **VOC12** | **87.9** |
| CA-Faster R-CNN Leng and Liu (2022) | Spatial/Global | VGG16/Faster R-CNN | Context Augmentation | VOC12 | 81.7 |
| ESCNet Nie et al. (2020) | Scale,Spatial/Global | ResNet101/SSD | TAM,CGA,SA,ECM | VOC12 | 80.9 |
| CLN Network Leng et al. (2018) | Semantic/Global,Local | VGG16/Faster-RCNN | MLP, ConvNet | VOC12 | 80.7 |
| DRFNet Xie et al. (2020) | Spatial/Local | VGG-16 | Context Aggregation,DRF | VOC12 | 80.4 |
| EGCI-Net Guo et al. (2020) | Spatial/Global | DS/64-192-48-1 | Global blocks,PFPM,PPM,MFP | VOC12 | 78.5 |
| SIN Liu et al. (2018) | Spatial,Semantic/Local, Global | VGG-16/Faster R-CNN | Message Passing,Scene,Edge | VOC12 | 73.1 |
| Multi-scale Fusion Wang et al. (2018) | Spatial,Semantics,Scale/Local,Global | VGG16 | - | VOC12 | 72.0 |
| Lightweight Global Unal and Kovashka (2021) | Semantic,Spatial | VGG16/SIN,Faster RCNN | GCNs,GRU mechanism | VOC12 | 67.2 |
| Context Augmentation Dvornik et al. (2018) | Spatial/Local,Prior Knowledge | ResNet50 | Data Augmentation | VOC12 | 65.9 |
| WSOD Symmetry Context Gu et al. (2022) | Spatial/Local | VGG-16 | Naive-Gaussian context mining,SCM | VOC12 | 48.3 |
| Adaptive WSOD Zeng et al. (2021) | Spatial/Global | VGG-16/Faster-RCNN | MIL, Context Awareness Block, SPP | VOC12 | 47.8 |
| **Feature refinement** Ma and Wang (2023) | **Spatial/Local,Global** | **EfficientNet-B6** | **Cross-Pooling,Cascaded Context** | **VOC07** | **86.2** |
| FGDLAM and SCIF Deng et al. (2024) | Spatial/Local,Global | CSPv5/YOLOx | FGDLAM joint SCIF | VOC07 | 85.60 |
| ReCoR Chen et al. (2021) | Spatial/Global | ResNet/Cascade MaskRCNN | Recursive Context Routing | VOC07 | 83.9 |
| LCEM Jiaxuan et al. (2022) | Scale,Spatial/Global | ResNet/Faster R-CNN | AC-FPN,LCEM,CGA,LSIoU | VOC07 | 83.7 |
| CCAGNet Miao et al. (2022) | Spatial,Semantic,Scale/Local,Global | VGG16/SSD | CCAM,RFAM,SFAM | VOC07 | 83.7 |
| FRCNN+Seg+HCE Qiu et al. (2020) | Scale/Local,Global | ResNet/Mask RCNN | Hierarchical Context,GED | VOC07 | 83.5 |
| PCL Ding et al. (2020) | Local,Global | VGG16/RetinaNet | PCL module | VOC07 | 83.4 |
| YOLOv4+AMCM Ma et al. (2022) | Scale,Semantic/Local,Global | CSPDarkNet53/ YOLOv4 | AMCM | VOC07 | 83.2 |
| CA-Faster R-CNN Leng and Liu 2022) | Spatial/Global | VGG16/Faster RCNN | Context Augmentation | VOC07 | 83.1 |
| CHFANet Xu et al. (2020) | Scale,Spatial/Local,Global | VGG16/SSD | Channel-Wise Attention,CFE,HFF | VOC07 | 82.6 |
| CEBNet Chen et al. (2019) | Scale,Spatial/Local | Modified VGG16/SSD | ERFB,FAB,CEBs | VOC07 | 82.5 |
| DRFNet Xie et al. (2020) | Spatial/Local | VGG-16 | Context Aggregation, DRF | VOC07 | 82.3 |
| Context Refinement Chen et al. (2018) | Semantic,Spatial/Local | ResNet/C-Mask RCNN | Adaptive Weighting | VOC07 | 82.2 |
| ESCNet Nie et al. (2020) | Scale,Spatial/Global | ResNet101/SSD | TAM,CGA,SA,ECM | VOC07 | 82.1 |

**Table 6** (continued)

| Detector name | Context type/level | Backbone/Architecture | Mechanism/Module | Dataset | mAP |
|---|---|---|---|---|---|
| CLN Leng et al. (2018) | Semantic/Local,Global | VGG16/Faster-RCNN | MLP, ConvNet | VOC07 | 82.1 |
| YOLOv8-CGRNet Niu et al. (2023) | Spatial/Local,Global | CSPNet/YOLO8 | CGNet,SimPPFCSPC,DyHead | VOC07 | 81.9 |
| CFFN Xia et al. (2020) | Scale,Spatial,Semantic/Global | ResNet-101/FSSD | Attention,CRM,CEM | VOC07 | 81.4 |
| EGCI-Net Guo et al. (2020) | Spatial/Global | DS/64-192-48-1 | Global blocks,PFPM, PPM | VOC07 | 80.2 |
| CP-SSD Jiang et al. (2019) | Spatial,Scale,Semantic/Local,Global | VGG16/SSD | Context Module, Semantic Module | VOC07 | 77.8 |
| SC-Faster R-CNN Xiao et al. (2020) | Semantic,Scale,Spatial/Local,Global | VGG16/Faster R-CNN | GA-RPN, Skip pooling | VOC07 | 77.6 |
| Deformable R-FCN-BAN Kim et al. (2018) | Spatial/Local | ResNet/R-FCN | BAN | VOC07 | 76.7 |
| SIN Liu et al. (2018) | Spatial,Semantic/Local, Global | VGG-16/Faster RCNN | Message Passing,Scene,Edge | VOC07 | 76.0 |
| Multi-scale Fusion Wang et al. (2018) | Spatial,Semantics,Scale/Local,Global | VGG16 | - | VOC07 | 75.9 |
| Object Detection System Chu and Cai 2018) | Semantic,Spatial/Local,Global | VGG/Faster R-CNN | CRF | VOC07 | 73.5 |
| Context Data Augmentation Zhang et al. (2021) | Spatial/Global | VGG16/Faster R-CNN | CRPM,CLSAM | VOC07 | 70.6 |
| Conv Context Features Kaya and Alatan (2018) | Spatial/Local | VGG-16/Faster R-CNN | Context Coordinates, Wrap Operation | VOC07 | 64.67 |
| FNM Barnea and Ben-Shahar (2019) | Spatial,Scale/Local | Faster R-CNN | Scale-invariant Representations | VOC07 | 62.13 |
| JLWSOD Lai et al. (2024) | Semantic/Local | ResNet-50 | iGCL,Instance Branch, Semantic Branch | VOC07 | 58.5 |
| Distilling Graph Yang et al. (2023) | Spatial,Semantic/Local,Global | ResNet | KD,Graph attention | VOC07 | 56.9 |
| Adaptive WSOD Zeng et al. (2021) | Spatial/Global | VGG-16/Faster-RCNN | MIL,Context Awareness Block, SPP | VOC07 | 53.0 |
| Fine hierarchical Cao et al. (2020) | Spatial/Local | VGG-16/MDP | RL,ReNet Layer | VOC07 | 53.0 |
| WSOD Symmetry Context Gu et al. (2022) | Spatial/Local | VGG-16 | Naive-Gaussian context mining,SCM | VOC07 | 52.4 |

Table 7 General object detection (other datasets)

| Detector name | Context type/level | Backbone/architecture | Mechanism/module | Dataset | mAP50 |
|---|---|---|---|---|---|
| CompositionalNet Wang et al. (2020) | Spatial/Local | CompositionalNets | Part-Based Voting | PASCAL3D | 41 |
| RDSP Akita and UKita (2023) | Scale,Spatial/Local,Global | CenterNet | Positional and scene structure embedding | CityScapes | 41.8 |
| GCA RCNN Zhang et al. (2021) | Spatial,Scale/Global | ResNet-101/RCNN | Global feature pyramid, attention | CityScapes | 37.4 |
| YOLO*C Oreski (2023) | Spatial/Global | CSPNet/YOLO7 | MCTX | BDD100k | 61.6 |
| RDSP Akita and UKita (2023) | Scale,Spatial/Local,Global | CenterNet | Positional and scene structure embedding | BDD100k | 29.6 |
| Improved YOLOv8 Saha et al. (2024) | Semantic,Spatial/Local | CSPNet/YOLO8 | C2f, decoupled head design | ALPR | 96.3 |
| UGC-YOLO Yang et al. (2023) | Semantic/Global | DarkNet-dgp/YOLOv3 | Global Context Block,PPM,Attention | Underwater | 79.03 |
| FA-FPN-MCI Bhalla et al. (2024) | Scale,Spatial,Environmental/Local,Global | ResNet-50 | SNR, RFB,TBGCM,quality embedding | S-UODAC2020 | 23.6 |
| SCFE Wang et al. (2018) | Spatial/Global | Custom/Inspired by RNN | - | KITTI | 93.7 |
| LTN Wang et al. (2019) | Spatial,Local,Global | RESNET/Spatial Transformer,Faster RCNN | Feature Fusion,STN | KITTI | 88.85 |
| YOLOv4+AMCM Ma et al. (2022) | Scale,Semantic,Local,Global | CSPDarkNet53/ YOLOv4 | AMCM | KITTI | 83.2 |
| LTN Wang et al. (2019) | Spatial,Local,Global | RESNET/Spatial Transformer,Faster RCNN | Feature Fusion,STN | MIO-TCD | 82.75 |
| LTN Wang et al. (2019) | Spatial,Local,Global | RESNET/Spatial Transformer,Faster RCNN | Feature Fusion,STN | Bosch | 31.64 |
| Cascade Region Proposal Zhong et al. (2020) | Spatial,Local,Global | ResNet/Faster RCNN | New RPN | ILSVRC | 65.3 |
| ECNet Xiao et al. (2021) | Scale,Spatial,Local,Global,Prior | ResNet-50 | EDM,DFFM,MFEM | COSD | 50.8 |
| ECNet Xiao et al. (2021) | Scale,Spatial,Local,Global,Prior | ResNet-50 | EDM,DFFM,MFEM | MSD | 58.1 |
| ECNet Xiao et al. (2021) | Scale,Spatial,Local,Global,Prior | ResNet-50 | EDM,DFFM,MFEM | GDD | 62.2 |
| Distilling Graph Yang et al. (2023) | Spatial,Semantic/Local,Global | ResNet | KD,Graph Attention | LVIS | 21.2 |
| Copy-Paste Augmentation Li et al. (2022) | Semantic,Spatial,Scale/Local,Global | Darknet/Yolov5 | Adaptive Transformations | NM10k | 55.5 |
| CIE-JHR Zhao et al. (2022) | Spatial,Scale/Local,Global | Swin Transformer | Attention,DVR | Aerial Vehicle | 67.1 |

**Table 7** (continued)

| Detector name | Context type/level | Backbone/architecture | Mechanism/module | Dataset | mAP50 |
|---|---|---|---|---|---|
| Of-The-Shelf Bardool et al. (2019) | Spatial,Semantic/Local,Global | VGG-16/Mask RCNN | Scoring and reasoning | Toy dataset | 79.27 |
| POD-F Ma et al. (2023) | Scale,Semantic,Spatial/Global | ResNet/Faster R-CNN | SA,GCFE,DSFA | SIXray | 86.1 |
| POD-F Ma et al. (2023) | Scale,Semantic,Spatial/Global | ResNet/Faster R-CNN | SA,GCFE,DSFA | OPIXray | 84.9 |
| POD-F Ma et al. (2023) | Scale,Semantic,Spatial/Global | ResNet/Faster R-CNN | SA,GCFE,DSFA | WIXray | 70.9 |
| POD-Y Ma et al. (2023) | Scale,Semantic,Spatial/Global | CSPDarknet/YOLOV5L | SA,GCFE,DSFA | POD-Y | 90.4 |
| POD-Y Ma et al. (2023) | Scale,Semantic,Spatial/Global | CSPDarknet/YOLOV5L | SA,GCFE,DSFA | OPIXray | 90.9 |
| POD-Y Ma et al. (2023) | Scale,Semantic,Spatial/Global | CSPDarknet/YOLOV5L | SA,GCFE,DSFA | WIXray | 71.9 |

CAB Net enhances the accuracy of detection while preventing an increase in model complexity.

(9) In (Fang and Shi 2018), a context information fusion technique is implemented that selectively combines classification information with region proposal windows. This method improves the efficiency of region proposal classification without adding errors in bounding box regression, unlike standard approaches that rely on both classification and bounding box information.

(10) Semantic Context-Aware Network (SCAN) (Guan et al. 2018) uses pyramid pooling to merge larger feature maps with context information. Two modules are utilized to combine them: a Location Fusion Module (LFM) for fine-grained semantic features, and a Context Fusion Module (CFM) for context-aware features.

(11) MCS-YOLO v4 (Ji et al. 2023) introduces a novel detection scale of 104x104 to gather more detailed information about small objects. The Expanded Field of Sensation Block (EFB) collects contextual information around small objects, hence improving feature richness.

(12) Contextual Information Fusion (Chen et al. 2021) employs feature fusion and multi-scale output predictions to integrate contextual information into the network for enhancing resilience in detecting small objects.

(13) Vanishing-Point-Guided Context-Aware Network (VCANet) (Chen et al. 2021) employs a vanishing point prediction block and a context-aware center detection block to collect and extract semantic information. This model has superior accuracy in recognizing small objects on roads compared to generic object detection methods.

(14) Discriminative Learning and Graph-Cut (DLGC) framework (Xi et al. 2020) leverages semantic similarity among predicted objects' candidates. The framework entails the creation of a pairwise constraint to depict semantic similarity, implementing discriminative learning to assess potential similarity, and deploying a graph-cut method to group candidates according to their similarity. Once a graph model has been constructed and candidates have been divided into different groups, a voting mechanism is utilized to ascertain the categorization of candidates within each group. The voting technique improves the accuracy of object detectors by utilizing semantic information and cohesive relationships among neighboring candidates.

(15) CEASC (Du et al. 2023), ptimizes the detection head in drone images. It utilizes a context-enhanced group normalization (CE-GN) layer that incorporates global contextual features to improve accuracy while operating on sparsely sampled regions through adaptive sparse convolution. It also implements an adaptive multi-layer masking (AMM) strategy to dynamically adjust the mask ratio for foreground coverage, ensuring a balance between computational efficiency and detection performance.

(16) Dynamic Local and Global Context Exploration (DCE) (Zhang et al. 2023) dynamically explores local and global context features. DCE includes Dynamic Surrounding Search (DSS), Semantic Object Relation Enhancement (SORE), and Global Feature Supplement (GFS) to enhance detection performance.

(17) The Eagle-YOLO (Ma et al. 2024) integrates a Lightweight Kernel Attention (LKA) mechanism and contextual feature fusion to enhance detection accuracy in complex scenes. Building on the YOLOv5 framework, it employs a Backbone module (CSP-Darknet53) for feature extraction, a Neck module that incorporates a Bidirectional Feature Pyramid Network (Bi-FPN) with LKA to focus on target regions, and a Head

**Table 8** Addressed challenges in general object detection

| Detector name | Highlighters (problems addressed) |
|---|---|
| Distilling Knowledge Graph Yang et al. (2023) | Semantic gap in previous crossmodal approaches |
| KROD Ji et al. (2022) | Limitations of treating each region independently, and lacking crucial global context |
| SIN Liu et al. (2018) | Challenge of incorporating scene context and object relationships within a single image |
| HCE Chen et al. (2020) | Limited context in two-stage detectors that rely on region-wise features extracted by RoIPool or RoIAlign |
| Hierarchical Context Embedding Qiu et al. (2020) | Challenge of integrating pixel-level segmentation into object detection |
| AdaCon Neseem and Reda (2021) | Computational and energy challenges of deploying CNNs for object detection on resource-constrained edge devices |
| GCA RCNN Zhang et al. (2021) | Addressing losing contextual information in the process of resizing the object proposals in two-stage detectors |
| Context refinement Chen et al. (2018) | Poorly localized suggested areas in two-stage detectors |
| GCDN Li et al. (2022) | Lack of global contextual information in two-stage object detectors |
| GCE module Peng et al. (2022) | Constraints of bottom-up manners in two-stage object detectors |
| CLN Leng et al. (2018) | Limitations of classifying candidate proposals using their interior features in two-stage detectors |
| CCAGNet (Miao et al. 2022) | Coping with high complexity and slow speed anchor-free frameworks in real scenarios in one-stage detectors |
| DRFNet Xie et al. (2020) | Computational and context quality limitations of context-utilizing one-stage detectors |
| CFFN Xia et al. (2020) | Restrictions caused by a lack of global context and the presence of background noise in low-level features in one-stage detectors |
| CL-FPN Yang et al. (2023) | Spatial and semantic context issues in large receptive fields and high-resolution data in multi-scale data |
| MCFPN Wang et al. (2022) | Ignoring the context information gap across different levels in FPN in multi-scale data |
| FNM Barnea and Ben-Shahar (2019) | Collision of objects in multi-scale data |
| MSF Wang et al. (2018) | Maximizing the use of spatial information in multi-scale data |
| ESCNet Nie et al. (2020) | Issue of multi-scale object detection in shallow pyramid layers |
| PCL Ding et al. (2020) | Challenge of considering the local and global contexts separately in multi-scale data |
| Cascade region proposal Zhong et al. (2020) | Low accuracy of region proposal and a deep object recognition |
| Pure regression object detection Fan et al. (2022) | Degradation of classification due to relying on bounding boxes that don't consider context between objects |
| SA Chen et al. (2022) | Lacking the capture of long-range dependencies in conventional non-local blocks |
| GC Lee et al. (2021) | Ambiguous objects |
| CSA-Net Liang et al. (2022) | Losing information for small and medium objects during deep feature extraction |
| ReCoR Chen et al. (2021) | Limited expressive capacity and dynamics to encode contextual relationships |
| MCENet Wang and Ma (2022) | Dealing with square kernels in convolution and pooling operations that ignores long-range correlation between pixels |
| BAN Kim et al. (2018) | Exploiting the visual contexts including boundary information and surroundings |

**Table 8** (continued)

| Detector name | Highlighters (problems addressed) |
| --- | --- |
| CompositionalNet Wang et al. (2020), SC-Faster RCNN Xiao et al. (2020) | Detecting occluded objects |
| CA-Faster R-CNN Leng and Liu (2022) | Poor region proposals and inaccurate proposal recognition for tiny and obstructed objects in two-stage detectors |
| Context Augmentation Dvornik et al. (2018) | Managing limitations related to few labeled examples |
| Context-based data augmentation Zhang et al. (2021) | Managing context loss, noise, and the scale of various objects |
| CHFANet Xu et al. (2020) | Tackling limitations of fusing pyramidal features extracted from ConvNets |
| Fine hierarchical Cao et al. (2020) | Loss of the target region due to object sizes that differ from those in the source images |
| CEBNet Chen et al. (2019) | Managing high computational burden in multi-scale and small objects |
| Adaptive WSOD Zeng et al. (2021) | Addressing local optima and object ambiguity |
| LCEM Jiaxuan et al. (2022) | Small objects with weak appearances |
| Object detection system Chu and Cai (2018) | Lack of contextual information surrounding R-CNN object proposals |
| Conv Context Features Kaya and Alatan (2018) | Determining boundaries of objects |
| SCFE Wang et al. (2018) | Lack of using spatial context in CNNs networks that combine the local geometric and texture features |
| LTN Wang et al. (2019) | Managing intra-class variations |
| ECNet Xiao et al. (2021) | Transparent and reflective objects such as mirrors and glass products |
| Copy-Paste data augmentation Li et al. (2022) | Rare objects in autonomous driving |
| CIE-JHR Zhao et al. (2022) | Small and obscured objects |
| POD-F and POD-Y Ma et al. (2023) | Addressing Prohibited Object Detection (POD) in X-ray images, covering issues such as occlusion |
| RDSP Akita and UKita (2023) | False-positive detections in SR-based object detection |
| YOLOv8-CGRNet Niu et al. (2023) | Addressing accuracy on mobile devices with limited memory and computational resources |
| YOLOC Oreski (2023) | Objects in complex traffic |
| Multi-scale model Ma and Wang (2023) | Addressing accuracy in cluttered backgrounds |
| TCC Chen et al. (2023) | Addressing computational complexity |
| UGC-YOLO Yang et al. (2023),FA-FPN-MCI Bhalla et al. (2024) | Addressing complex and degraded underwater environments |

module that includes an additional detection layer for small targets. Additionally, the model introduces the Eagle-IoU loss function, which addresses gradient instability and convergence issues during training.

(18) The paper (Cheng et al. 2023) introduces a multi-level feature fusion module to improve the extraction of detailed information, addressing the challenges of missed and false detections. A regional attention module is employed to focus on small object features while minimizing the influence of background noise. Additionally, the anchor boxes are refined to better accommodate small objects.

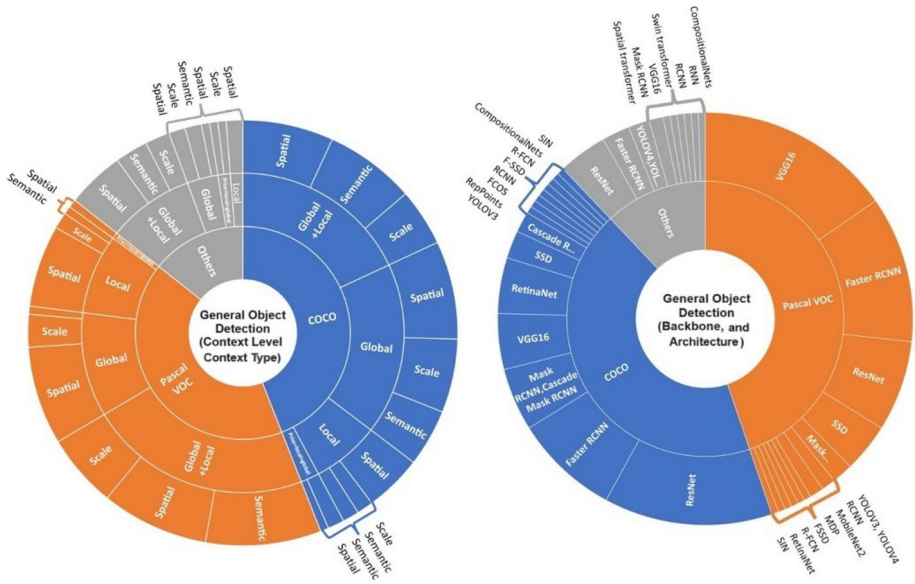

**Fig. 39** Small birds with 20 pixels are much more difficult to detect compared to those with 150 or 120 pixels

**Fig. 40** Overview of datasets, context levels, context types, and architectures employed in general object detection approaches. The size of each section indicates the contribution of that section

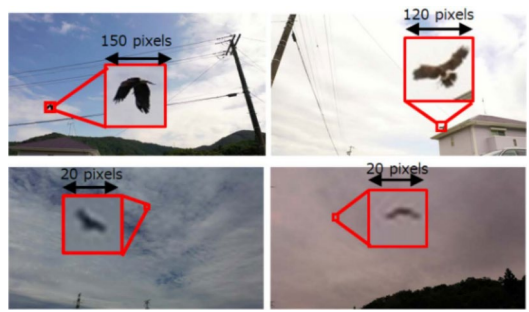

(19) The paper (Kim et al. 2024) proposes a novel drone-view object detection framework that integrates environmental factors like weather, illumination, and visibility to enhance detection robustness against adverse conditions. Utilizing a multimodal large language model (MLLM), the authors generate a diverse weather content feature set, which describes the environmental conditions associated with drone-view images. The framework then adaptively selects the most relevant weather content features and combines them with learnable object queries to improve contextual understanding in the detection process.

(20) The proposed approach in (Jing et al. 2024) presents leverages a composite backbone network, context information, and multi-scale learning to enhance detection accuracy. The approach includes a Composite Dilated Convolution and Attention Module (CDAM) to efficiently integrate context information while reducing noise interference.

Additionally, a Feature Elimination Module (FEM) is introduced to suppress features of medium and large objects, allowing for better detection of small objects.

(21) SG-YOLO (Deng et al. 2024) enhances underwater target detection by integrating a Feature Fusion Module (FFM) to balance semantic information and mitigate feature loss, a Global Context Decoupling (GCD) Head to filter out irrelevant background details while processing classification and regression tasks separately, and a Content-Aware Reassembly Module (CFRM) that adapts upsampling with position-specific kernels to improve localization accuracy for small targets.

(22) The Bi-AFN++CA (Zhang and Chen 2024) integrates a Bi-directional Adaptive Fusion Network that facilitates simultaneous information flow in both top-down and bottom-up pathways, combined with a Context Extraction Module to capture rich contextual spatial-channel information. This dual approach mitigates the challenges of detecting small objects in cluttered environments by refining feature representation and incorporating contextual cues.

(23) The paper (He et al. 2024) introduces three lightweight plug-and-play modules-CIE-Pool, SE-CBAM, and Adaptive Feature Processing (AFP)-to enhance the detection accuracy of small objects within YOLO-based algorithms. The CIE-Pool module enriches feature extraction by incorporating contextual information, while the SE-CBAM module enhances spatial attention for better localization of small objects. The AFP further optimizes feature representation by filtering noise from shallow layers.

### 4.3.1 Results on small object detection

A comprehensive analysis of small object detectors is shown in Table 9. Moreover, context levels, context types, and CNN architectures in small object detection approaches are presented in Fig. 41. As shown in Fig. 41, the combination of local and global context in this category has garnered more attention, resulting in improved networks for detecting small objects. Spatial context, scale context, and semantic context have also been used much more than other contextual information in small object detectors. Since a wide variety of datasets have been used in this category, it is not logical to compare the mAPs of their predictions. Noteworthy is the fact that various versions of the YOLO network have been used more frequently than other architectures. The combination of YOLO with contextual information has led to acceptable results in detecting small objects. In small object detection, the reviewed models show diverse methods to address challenges of feature clarity, contextual information, and computational efficiency. FA-SSD, CEFP2N, and IENet focus on enhancing feature representation by incorporating multi-scale and attention-based modules. FA-SSD uses feature fusion and attention for precise context capture, benefiting small objects but with an increase in computational demand. CEFP2N's purification process effectively reduces noise, enhancing clarity, though the multi-step approach may impact speed. IENet leverages internal and external context through reasoning modules, which improves detection but can be complex to implement. Improved YOLOv5, CGA-YOLO, and Contextual-YOLOv3 each integrate context into YOLO architectures to retain small object details. Improved YOLOv5's coordinate attention boosts positional accuracy, though it remains sensitive to noise in complex backgrounds. CGA-YOLO's Swin Transformer enhances context extraction, offering robust small-object detection with high computational needs. Contextual-YOLOv3, meanwhile, prioritizes classification accuracy but may miss

175 Page 60 of 89 M. Jamali et al.**Table 9** Small object detection

| Detector name | Context type/level | Backbone/architecture | Mechanism/module | Dataset | mAP |
|---|---|---|---|---|---|
| **CEFP2N** Xiao, Guo, et al. (2023) | **Spatial,Semantic,Scale/Local,Global** | **Darknet53** | **Copy-reduce-paste,CEM,FPM** | **VOC2007** | **83.6** |
| FA-SSD Lim et al. (2021) | Scale,Spatial,Semantic/Local,Global | ResNet50/SSD | A-SSD, F-SSD | VOC2007 | 78.3 |
| **Improved Faster RCNN** Cheng et al. (2023) | **Spatial,Semantic/Local,Global** | **VGG16/Faster RCNN** | **Feature Fusion,Regional Attention,Anchor Frame** | **MSCOCO** | **53.8** |
| IENet Leng et al. (2021) | Spatial,Semantic,Scale/Local,Global | ResNet-101 | Bi-FFM,CRM,CFAM | MSCOCO | 51.2 |
| Composite Backbone Jing et al. (2024) | Spatial,Semantic/Local | CSPDarknet53/YOLOv7 | CDAM,FEM | MSCOCO | 47.2 |
| Bi-AFN++CA Zhang and Chen (2024) | Spatial/Local,Global | ResNet50/Faster RCNN | SRM, DEM, CEM | MSCOCO | 30.6 |
| IENet Leng et al. (2021) | Spatial,Semantic,Scale/Local,Global | ResNet-101 | Bi-FFM,CRM,CFAM | WIDER FACE | 93.46 |
| **Weather-aware drone** Kim et al. (2024) | **Environmental/Local,Global** | **ResNet-50/DDQ-DETR** | **Weather-aware object queries** | **VisDrone 2019** | **66.4** |
| CGA-YOLO Hang et al. (2022) | Spatial,Semantic/Global | CSPDarknet53/YOLOv5 | Window-based Self-attention,GAM,CAM,SAM | VisDrone 2019 | 45.25 |
| Improved YOLOv8 He et al. (2024) | Spatial,Scale/Local,Global | CSP-DarkNet/YOLOv8 | CIE-Pool, SE-CBAM, AFP | VisDrone 2019 | 37.1 |
| Eagle-YOLO Ma et al. (2024) | Semantic,Spatial/Local,Global | CSPDarknet53/YOLOv5 | LKA mechanism | VisDrone 2019 | 32.91 |
| Improved YOLOv5 Zhang et al. (2022) | Semantic,Spatial/Local | CSPDarknet53/YOLOv5 | CA,CFEM | VisDrone 2019 | 30.15 |
| Bi-AFN++CA Zhang and Chen (2024) | Spatial/Local,Global | ResNet50/Faster RCNN | SRM, DEM, CEM | VisDrone 2019 | 23.8 |
| DCE Zhang et al. (2023) | Semantic/Local,Global | ResNet50/detectoRS | SORE, GFS, DSS | VisDrone 2019 | 22.9 |
| CEASC Du et al. (2023) | Spatial,Scale/Global | ResNet50/Retinanet | CE-GN and AMM | VisDrone 2019 | 20.8 |
| TYOLOv5 Corsel et al. (2023) | Spatio-Temporal/Local,Global | CSPDarknet53/YOLOv5 | Temporal Augmentation, Spatiotemporal Detector | WPAFB WAMI | 76.9 |

Springer

**Table 9** (continued)

| Detector name | Context type/level | Backbone/ architecture | Mechanism/module | Dataset | mAP |
|---|---|---|---|---|---|
| **Contextual-YOLOV3** Luo et al. (2019) | **Spatial,Semantic/ Local,Global** | **Darknet53/ YOLOV3** | **CRM, Context-Based Filtering Algorithm** | **Tsinghua-Tencent** | **94.0** |
| CAB Net Cui et al. (2020) | Spatial,Semantic/ Local,Global | VGG16 | Context-Aware Block(CAB) | Tsinghua-Tencent, | 78.0 |
| Contextual-YOLOV3 Luo et al. (2019) | Spatial,Semantic/ Local,Global | Darknet53/ YOLOV3 | CRM, Context-Based Filtering | National Grid | 93 |
| CAB Net Cui et al. (2020) | Spatial,Semantic/ Local,Global | VGG16 | Context-Aware Block(CAB) | Airport | 77.4 |
| Faster RCNN+FPN Fang and Shi (2018) | Spatial/Local | RESNET/ Faster RCNN | Context Information Fusion | Subset of COCO | 23.21 |
| SCAN Guan et al. (2018) | Scale,Semantic,Spatial | Inception/ PVANET | LFM,CFM | KITTI | 81.96 |
| MCS-YOLO v4 Ji et al. (2023) | Scale,Spatial/Local | CSPDarknet53/ YOLOv4 | Channel and Spatial Attention,EFB | Cars in UCAS | 84.18 |
| MCS-YOLO v4 Ji et al. (2023) | Scale,Spatial/Local | CSPDarknet53/ YOLOv4 | Channel and Spatial Attention,EFB | RSOD | 84.63 |
| Contextual Fusion Chen et al. (2021) | Spatial,Scale,Semantic/ Local,Global | VGG16/ SSD | Non-Maximum Suppression, Feature Fusion | NWPU VHR-10 | 88.5 |
| VCANet Chen et al. (2021) | Semantic,Scale,Spatial/ Local,Global | ResNet | VPT Block,Context Center Block | TJ-LDRO | 68.5 |
| DLGC Xi et al. (2020) | Semantic/Local,Global | Resnet/Faster RCNN | Graph-Cut,Voting Method | RSIs (DOTA) | 61.73 |
| DCE Zhang et al. (2023) | Semantic/Local,Global | ResNet50/ detectoRS | SORE, GFS, DSS | DOTA | 28.4 |
| SG-YOLO Deng et al. (2024) | Spatial,Semantic/ Local,Global | CSPDarknet53/ YOLOv5-M | FFM,GCD head, CFRM | UTDAC2020 | 87.7 |

localization precision in dense scenes. Models like DCE, Eagle-YOLO, and VCANet excel by fusing local and global context with targeted modules, which refines object boundaries and reduces false positives. Eagle-YOLO's lightweight attention mechanism balances accuracy with speed, but detection can be impacted under extreme conditions. VCANet, which employs a vanishing-point-based context block, excels in road scenarios but is less effective in varied settings. Lastly, MCS-YOLOv4, SG-YOLO, and the Bi-AFN++CA model focus on balancing semantic information and spatial details, reducing feature loss and enhancing localization in challenging environments. SG-YOLO's adaptive reassembly mitigates noise effectively, while Bi-AFN++CA's dual-pathway approach captures rich context, excelling in cluttered scenes though at the cost of increased model complexity.

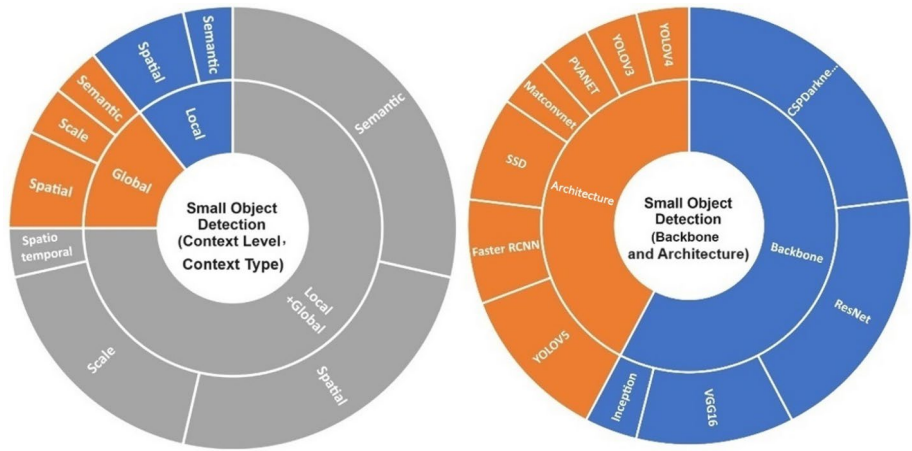

**Fig. 41** Overview of context levels, context types, and architectures employed in small object detection approaches. The size of each section indicates the contribution of that section

### 4.4 Video object detection (VOD)

Video object detection involves detecting objects using video data as compared to conventional object detection using static images (Zhu et al. 2020). In video object detection, objects may move and change their appearance across frames. As a result, objects should be detected and tracked across consecutive frames. Contextual information can aid video object detection by leveraging spatial and temporal cues, enhancing object recognition and tracking consistency across frames. In this section, context-based video object detection approaches are reviewed.

(1) Context Faster R-CNN (Beery et al. 2020) was designed for static monitoring cameras that have low and irregular sample frequency. By utilizing temporal context extracted from unlabeled frames, the model improves the performance of object detection. This model employs an attention-based technique to index a Long Term Memory bank (Mlong) built on per-camera to collect long-term contextual information from previous unlabeled frames. Short Term Memory (Mshort) is also added to include short-term context from neighboring frames. This approach solves issues such as partially observed objects, low quality, and background distractors.

(2) In contrast to other methods that combine features at once, Progressive Temporal-Spatial Enhanced Transformer (PTSEFormer) (Wang et al. 2022), employs a progressive strategy to optimize feature utilization by integrating both temporal and spatial information. This method utilizes a Temporal Feature Aggregation Module (TFAM) to handle temporal context and a Spatial Transition Awareness Module (STAM) to handle spatial context.

(3) Flow and LSTM (Zhang and Kim 2019) is a causal recurrent flow-based method for real-time online detection. In contrast to conventional approaches that need a large number of preceding and succeeding frames to detect objects, making them impracticable for real-time online detection, the suggested method reads only the current frame and one prior frame from the memory buffer at each time step. This model employs short-term temporal context via optical flow-based feature warping from the previous frame and long-term temporal context via a temporal convolutional LSTM, and combines both long-term and short-term temporal context information.

(4) Temporal Context Enhanced Network (TCENet) (He et al. 2020) mitigates appearance degradation in video frames by proposing Temporal Context Enhanced Aggregation (TCEA), which aggregates features from adjacent frames to model temporal context. This network comprises a DeformAlign module that ensures precise spatial alignment at the pixel level throughout time, as well as a Temporal Stride Predictor that intelligently chooses video frames for aggregation. TCENet handles severe appearance degradation such unusual poses and occlusions better than traditional spatial feature enhanced aggregation approaches.

(5) Video Object Detection Using Object's Motion Context and Spatio-Temporal Feature Aggregation (VOD-MT) (Kim et al. 2021) was designed for one-stage detectors. It takes advantage of both motion context and spatio-temporal features collected across multiple frames. This approach computes correlation maps between neighboring frames, and encodes them for motion context using LSTM. Subsequently, a gated attention network is utilized to collect spatial feature maps. This method can be useful to detect objects in videos with motion blur or defocusing.

(6) Context and Structure Mining Network (CSMN) (Han et al. 2021) addresses issues in video frames such as occlusion, motion blur, and uncommon postures using a spatial-temporal Context Information Encoding module (stCIE) and a Structure-based Proposed Feature Aggregation module (SPFA). The stCIE encodes spatial-temporal contextual information into object features by evaluating each object pixel's spatial and temporal relationship with surrounding pixels. On the other hand, the SPFA divides proposals into several patches, effectively dealing with problems such as misalignment of poses and occlusion.

(7) Motion Context Network (MC-Net) (Jin et al. 2020) improves weakly supervised object detection in videos by utilizing motion context. In order to overcome the difficulties associated with accurately locating objects in the absence of box-level annotations and complex motion patterns, MC-Net implements a Motion Context Module (MCM) that utilizes neighborhood motion correlation to derive motion context features, which are then effectively combined with appearance information. Furthermore, a Temporal Aggregation Module (TAM) addresses problems associated with degraded object appearances by consolidating features over consecutive frames.

(8) Non-local prior based spatiotemporal attention model (Lu et al. 2020) aims to utilize global spatio-temporal context and non-local dependencies in order to improve video object detection. The technique enhances accuracy in challenging video sequences, such as intricate backgrounds, motion blur, and partial occlusion, by employing non-local blocks and 3D convolutions into the YOLOv2 network. The method incorporates self-attention mechanisms through non-local blocks that capture long-range dependencies in both spatial and temporal forms, improving features by evaluating the similarity between any two points in the image. On the other hand, 3D convolutions gather temporal information by merging frames.

(9) Adaptive omni-attention model (Yu et al. 2022) addresses challenges such as lighting variations, smaller objects, and motion blur in video frames. The model includes inter-frame and intra-frame attention modules, in addition to a feature fusion module. It uses inter-frame contextual information to enhance identification in low-quality frames and leverages intra-frame attention to reduce false positive detections in background regions. This strategy makes optimal use of attention in the temporal, spatial, and channel domains, enhancing detection accuracy while reducing training costs.

(10) Temporal Aggregation with Context Focusing (TACF) (Han et al. 2023) framework enhances few-shot video object detection by integrating temporal information from adjacent frames with contextual details from support images. It comprises two main modules: the Context Focusing (CF) module, which computes similarity scores between support images and adjacent frame features to focus on relevant target object features, and the Temporal Aggregation (TA) module, which aggregates features from adjacent frames based on their similarity to the Region of Interest (ROI) features. This approach effectively addresses challenges like occlusion and motion blur.

### 4.4.1 Results on video object detection

A comprehensive analysis of video object detection approaches is shown in Table 10. Moreover, context levels, context types, and CNN architectures in video object detection approaches are presented in Fig. 42. Temporal consistency, scale variation, occlusions, and motion blur are some challenges that have been addressed by video object detection approaches. The ImageNet VID dataset is frequently employed to evaluate VOD models, and PTSEFormer built upon the DETR framework based on temporal-spatial context, outperforms other approaches on this dataset. Moreover, VOD-MT (Kim et al. 2021) and MC-Net (Jin et al. 2020) introduce a new concept called "motion context" in their approaches, which, in addition to spatial and temporal aspects, also addresses the motion characteristics of objects. This allows for the identification of objects in videos with defocusing or motion blur through motion cues. In terms of context type, context level, and CNN architecture, spatio-temporal context, local+global level, ResNet, and R-FCN are the most commonly used in video object detection approaches. For video object detection, contextual information plays a critical role in handling challenges like motion blur, occlusion, and appearance variations across frames. Context Faster R-CNN leverages long- and short-term memory mechanisms for static monitoring, utilizing temporal context from prior frames, which helps address partially observed objects and background noise. PTSEFormer adopts a progressive strategy for temporal and spatial context integration, offering refined contextual aggregation across frames. In contrast, Flow and LSTM improve real-time capabilities by utilizing only the current and previous frames, combining optical flow-based short-term context with LSTM-based long-term context for efficient online detection. TCENet addresses appearance degradation by aligning pixel-level features across time, which effectively counters occlusions and unusual poses, while VOD-MT focuses on spatio-temporal features and motion context, enhancing detection in videos with motion blur. CSMN introduces structured feature aggregation to improve detection accuracy amid occlusions and complex postures, combining spatial-temporal context with proposal patches. MC-Net and the Non-local prior model also emphasize motion context, with MC-Net specifically optimizing for weakly-supervised settings, and the Non-local prior model using non-local dependencies for global spatio-temporal attention. Meanwhile, the Adaptive omni-attention model addresses quality variations with intra- and inter-frame attention modules, enhancing accuracy in challenging

Table 10 Video object detection

| Detector name | Context type/level | Backbone/architecture | Mechanism/module | Dataset | mAP |
|---|---|---|---|---|---|
| Context Faster R-CNN Beery et al. (2020) | Long Term Temporal/ Local,Global | Resnet/Faster R-CNN | Attention, Mlong,Mshort | Caltech Camera Traps | 76.3 |
| **PTSE-Former Wang et al. (2022)** | **Temporal-Spatial(Short Temporal)/Global** | **ResNet/Deformable DETR** | **Attention,TFAM,STAM** | **ImageNet VID** | **88.1** |
| CSMN Han et al. (2021) | Spatial-Temporal/ Local,Global | ResNeXt-101 | stCIE, SPFA | ImageNet VID | 86.2 |
| TCENet He et al. (2020) | Temporal,Spatial/Global | ResNet/R-FCN | DeformAlign,Temporal Predictor | ImageNet VID | 81 |
| Flow and LSTM Zhang and Kim (2019) | Long Short Temporal/ Local,Global | ResNet/ FlowNet,LSTM,R-FCN | Optical flow,Temporal LSTM | ImageNet VID | 75.5 |
| VOD-MT Kim et al. (2021) | Spatio-Temporal,Motion/ Local,Global | VGG-16/SSD300 | Gated Attention Network, LSTM | ImageNet VID | 73.2 |
| MC-Net Jin et al. (2020) | Motion,Temporal/ Local,Global | custom | MCM, TAM | ImageNet VID | 37.4 |
| Spatio-temporal Lu et al. (2020) | Spatio-Temporal/Global | Darknet/Yolov2 | Non-local Blocks, 3D Conv | OTB50 | 84.31 |
| Spatio-temporal Lu et al. (2020) | Spatio-Temporal/Global | Darknet/Yolov2 | Non-local Blocks, 3D Conv | OCS driving recorder | 58.31 |
| Adaptive Attention Yu et al. (2022) | Temporal,Spatial/ Local,Global | DLA-34/CenterNet | inter and intra attentions | UA-DETRAC | 87.52 |
| TACF Han et al. (2023) | Temporal,Spatial/ Local,Global | ResNet-50/Faster R-CNN | CF, TA | FSVOD | 26.3 |

**Table 10** (continued)

| Detector name | Context type/level | Backbone/architecture | Mechanism/module | Dataset | mAP |
|---|---|---|---|---|---|
| TACF Han et al. (2023) | Temporal,Spatial/ Local,Global | ResNet-50/Faster R-CNN | CF, TA | FSYTV | 15.9 |

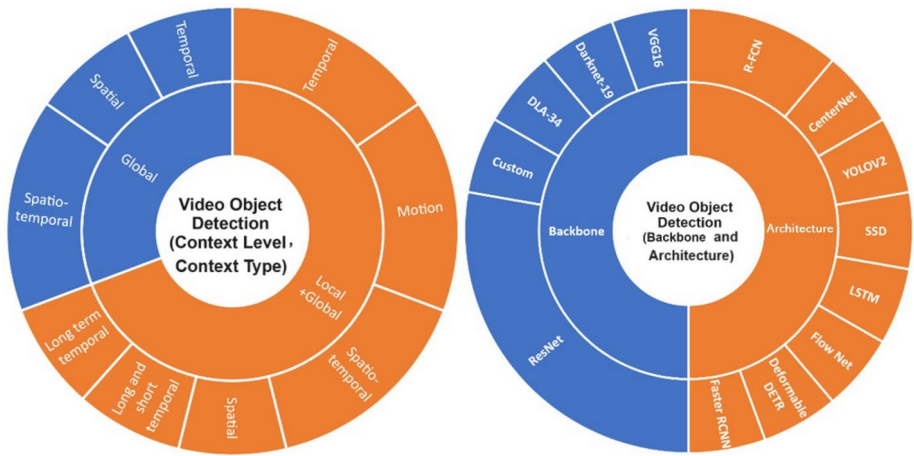

**Fig. 42** Overview of context levels, context types, and architectures employed in video object detection approaches. The size of each section indicates the contribution of that section in the reviewed articles

video sequences, while TACF framework improves few-shot detection by integrating context from support images, effectively countering occlusion and motion blur through focused temporal aggregation. Each method's integration of temporal and spatial context advances video object detection by enhancing feature richness, tracking consistency, and robustness to scene complexity.

### 4.5 Zero-shot, one-shot, and few-shot object detection

Zero-shot object detection (ZSD) refers to the task of detecting objects from categories that have not been seen during the training phase (Bansal et al. 2018). The model is trained to detect objects from known categories, and then it can generalize to detect objects from unseen categories based on auxiliary information or semantic embeddings. Only one published work, Multi-label Context (MLC) framework (Wei and Ma 2022), comprises four main components: the Multi-Label Head, Contextual RoI Feature Generation, Background Dynamic Generator (BDG), and Zero-Shot Head. First, the Multi-Label Head extracts object-level concepts from holistic, image-level context, helping the model learn relationships across the entire image. Then, Contextual RoI Features are generated by combining instance-level and global information, providing a complementary representation to conventional RoI features. The BDG dynamically updates background word vectors, reducing confusion between background and unseen objects. Finally, the Zero-Shot Head uses both the fused context and BDG to locate and classify seen and unseen objects, utilizing knowledge from previously seen classes to support zero-shot detection. This integration of global and local context strengthens the framework's ability to generalize to unseen classes.

In one-shot object detection (OSOD), a model is trained to detect objects in an image after being exposed to just one example of each object during the training phase (Osokin et al. 2020). This approach aims to enable the model to generalize and accurately detect previously unseen objects with minimal training instances. Attention mechanism mutual Global Context (mGC) block ( Jia et al. 2021) and Adaptive context and scale-aware feature aggregation module (ACS) with a feature alignment metric block (FAM) (Zhang et al. 2022) are context-based methods for one-shot object detection. Both have been tested on the COCO and PASCAL datasets, and the findings demonstrate that mGC performs superior in recognizing seen and unseen objects on both datasets. The mGC block ( Jia et al. 2021) focuses on important regions of an image by utilizing contextual information. This block improves the quality of features obtained from the image, specifically enhancing the region proposal network (RPN), which identifies the potential locations of objects. Once feature maps have been extracted and contextual information has been added to them using the mGC block, the process continues with the RPN, which produces enhanced Regions of Interest (ROIs). Ultimately, ROIs are categorized by a Metric-based Detector, which has the ability to handle both seen and unseen classes without requiring any further training. ACS+FAM (Zhang et al. 2022) safeguards crucial details during the OSOD process by integrating both global and local contextual information. The ACS module enables a strong knowledge of the object's surroundings and size fluctuations by combining context enrichment with conditioned multi-scale interaction. Meanwhile, the FAM block tackles issues related to spatial misalignment by employing feature alignment with the assistance of a spatial transformer network (STN) (Jaderberg et al. 2015), enhancing the overall accuracy and resilience of the OSOD system. CSSI model (Yang et al. 2024) proposes a novel method for one-shot object detection without requiring fine-tuning. It introduces the ATP module, which utilizes a transformer encoder to enhance long-range spatial interactions across different scales by aggregating features in a size-aware manner. Additionally, the GCC module extracts semantic-consistent spatial correlations by analyzing a complete 4D correlation tensor, complemented by inter-channel interactions through the CCL branch.

Few-shot object detection (FSOD) aims to extract semantic knowledge from limited object instances of novel categories within a target domain (Li et al. 2023). This method can be used in scenarios where limited labeled data is available for training an object detection model. Context has received more attention in FSOD, and four approaches, including Semantic Relation Reasoning Few-Shot Detector (SRR-FSD) (Zhu et al. 2021), Context-Transformer (Yang et al. 2020), and Dense Relation Distillation with Context-aware Aggregation (DCNet) (Hu et al. 2021), and Instance Context Network (ICNet) (Ran et al. 2023) are introduced. SRR-FSD (Zhu et al. 2021) combines semantic relations with visual information to improve the stability and robustness of FSOD across different shot variations. The detector utilizes semantic embeddings, which are obtained from textual data, to represent classes of objects. The model incorporates a relation reasoning module that considers the relations between different classes based on their semantic embeddings. The fundamental idea is that in situations where visual data is limited, the detection of new objects is facilitated by knowledge of the relationships between specific classes. Context-Transformer (Yang et al. 2020) surmounts the challenge of limited data diversity encountered in con-ventional transfer learning for FSOD by leveraging source-domain object knowledge and extracting context from the target domain's limited training set. Context-Transformer consists of two submodules: affinity discovery and context aggregation. Affinity discovery generates contextual fields for a certain target image using default prior boxes. Then, it utilizes the relations between these boxes and contextual fields. Moreover, context aggregation uses these relationships as a guide and inserts critical contexts into each box. The approach allows Context-Transformer to generate context-aware representations for each preceding box, enabling the detector to differentiate few-shot confusion using significant contextual cues. DCNet (Hu et al. 2021) has two main modules: Dense Relation Distillation Module (DRD) and Context-aware Feature Aggregation Module (CFA). The DRD module utilizes support features through a pixel-wise matching technique. Furthermore, in order to address the issue of scale variation, the CFA module collects features from various resolutions adaptively during RoI pooling. Another approach, Instance Context Network (ICNet) (Ran et al. 2023), integrates an instance-level context extraction module that utilizes a self-attention mechanism to improve feature representation. The method first extracts context information from the instances, which helps in enhancing the features of regions of interest. These enhanced features are then embedded into a transfer learning detection module, allowing for more effective differentiation between object classes.

### 4.5.1 Results on zero-shot, one-shot, and few-shot object detection

A comprehensive analysis of ZSD, OSOD, and FSOD approaches is shown in Table 11. The enhancement of networks in detecting objects in cases with not enough training data, known as corner cases (Heidecker et al. 2024), demonstrates the advantageous impact of context in such situations. The findings demonstrate that incorporating global semantic context has a positive role in detecting unseen objects in zero-shot object detection. In one-shot object detection, focusing on important regions of an image to detect both seen and unseen classes, and gathering a strong knowledge of the object's surroundings and size fluctuations are benefits of leveraging context. In few-shot object detection, improving the stability and robustness across different shot variations and tackling limited data diversity and scale variation are examples are advantages gained from using context. Nevertheless, given the scarcity of research in this domain, there exists an opportunity to further use contextual information.

### 4.6 Camouflaged object detection (COD)

Camouflaged object detection (COD) aims to identify objects, as shown in Fig. 43, that are seamlessly embedded in their surroundings, making them difficult to distinguish (Fan et al. 2020). The focus is on developing algorithms that can effectively identify such objects even in challenging backgrounds, where the object's appearance may match or mimic the surrounding environment.

Based on inclusion and exclusion criteria, six papers were found on the application of context in camouflaged object detection. They have been tested on different datasets, including the CAMO, Chameleon, COD10K, and NC4K. Context-aware Cross-level Fusion Network (C2F-Net) (Sun et al. 2021) has three main components: Multi-Scale Channel Attention (MSCA), Attention-induced Cross-level Fusion Module (ACFM), and Dual-branch Global Context Module (DGCM). MSCA effectively captures information at multiple scales by taking into account both global and local contexts; ACFM incorporates features from multiple levels with a particular emphasis on high-level features by utilizing attention mechanisms guided by MSCA; and DGCM makes additional use of global context data contained within the fused features. These modules work together to improve the model to detect camouflaged objects by improving feature representations at various sizes via attention-guided fusion and global context. Boundary-guided Context-aware Network (BCNet) (Xiao, Chen, et al. 2023) deals with inaccurate contours in COD. BCNet incorporates a High-resolution Feature Enhancement Module (HFEM) to extract multi-scale data while keeping precise cues, surpassing the limitations of prior approaches. Furthermore, in order to enhance segmentation with more precise contours, a Boundary-guided Feature Interaction Module (BFIM) is specifically engineered to uncover complementary information between camouflaged objects and their boundaries. These modules contribute to BCNet's performance in producing high-quality segmentation maps, demonstrating increases in accuracy as well as computational efficiency. Another approach titled CamoFocus (Khan et al. 2024) introduces two main modules: the Feature Split and Modulation (FSM) module, which separates and modulates foreground and background features using a supervisory mask, and the Context Refinement Module (CRM), which refines these features through cross-scale interactions. This approach improves semantic representation while maintaining lower computational complexity. Pixel-Centric Context Perception Network (PCPNet) (Song et al. 2023) uses a CNN-based encoder with a Vital Component Generation (VCG) module to extract rich spatial and semantic features. It employs a parameter-free Pixel Importance Estimation (PIE) function to prioritize pixels in complex backgrounds, guiding the network during decoding. A Local Continuity Refinement Module (LCRM) further refines detection results for improved accuracy. In another paper (Wen et al. 2024), the method combines the Swin Transformer (Swin-T) and EfficientNet-B7 models to enhance efficiency in feature extraction and segmentation. It introduces three key modules: the Masked-Edge Attention Module for efficient edge detection using Fourier transform, the Joint Dense Skip Attention Module for aggregating multi-level feature information, and the Object Attention Module to minimize discrepancies between encoder and decoder outputs. This method focuses on extracting both shallow and deep semantic information to improve the detection of camouflaged objects. Another approach, Discriminative Context-Aware Network (DiCANet) (Ike et al. 2024) enhances camouflaged object detection through a two-stage approach. It features an Adaptive Restoration Block (ARB) that intelligently weights feature channels and pixels, prioritizing informative data while suppressing noise, utilizing channel and pixel attention mechanisms. Following this, a Cascaded Detection Module (CDB) refines object predictions by enlarging the receptive field, producing accurate saliency maps with clear boundaries.

Table 11 Zero-shot, one-shot, and few-shot object detection

| Detector name | Task | Context type/level | Backbone/Architecture | Mechanism/Module | Approach of Learning | Dataset | mAP |
|---|---|---|---|---|---|---|---|
| MLC Wei and Ma (2022) | ZSD | Semantic/Global | Faster RCNN | BDG | multi-label learning | COCO | seen+unseen 15.7 |
| mGC Jia et al. (2021) | OSOD | Semantic/Global | ResNet/Fast RCNN | mGC block | Metric learning | COCO | seen: 44.8, unseen: 22.6 |
| mGC Jia et al. (2021) | OSOD | Semantic/Global | ResNet/Fast RCNN | mGC block | Metric learning | VOC07 | seen: 72.4, unseen: 72.7 |
| ACS+FAM Zhang et al. (2022) | OSOD | Spatial,Scale,Semantic/Local,Global | ResNet/STN | ACS,FAM+STN | attention | COCO | Seen:29.7,Unseen:12.6 |
| ACS+FAM Zhang et al. (2022) | OSOD | Spatial,Scale,Semantic/Local,Global | ResNet/STN | ACS,FAM+STN | attention | VOC07 | Seen:71.1,Unseen:72.3 |
| CSSI Yang et al. (2024) | OSOD | Spatial,Semantic/Local,Global | ResNet50 | ATP,GCC,CCL | Contrastive Learning | COCO | Seen:54.9,Unseen:27.7 |
| SRR-FSD Zhu et al. (2021) | FSOD | Semantic/Local,Global | ResNet/Faster RCNN | Relation reasoning | Metric Learning | COCO | 9.8 |
| SRR-FSD Zhu et al. (2021) | FSOD | Semantic/Local,Global | ResNet/Faster RCNN | Relation reasoning | Metric Learning | VOC07 | 56.8 |
| Context-Transformer Yang et al. (2020) | FSOD | Scale,Spatial,Semantic/Local | SSD | Affinity, Context Aggregation | Transfer Learning | VOC07+12 | 43.8 |
| DCNet Hu et al. (2021) | FSOD | Scales,Spatial/Local,Global | ResNet/MetaRCNN | DRD,CFA | Meta Learning, Matching | COCO mini | 18.6 |
| ICNet Ran et al. (2023) | FSOD | Semantic,Spatial/Local,Global | ResNet101+FPN/Faster R-CNN | Instance Context,CCA, Context Fusion | Transfer Learning,Attention | PASCAL10 | 53.2 |

### 4.6.1 Results on camouflaged object detection

The analysis of camouflaged object detection approaches is shown in Table 12. The results are sorted for each dataset based on MAE, where a lower MAE indicates better accuracy in locating and detecting camouflaged objects. For the other metrics-F-measure, S-measure, and E-measure-higher values are better, reflecting improved precision, structural similarity, and alignment with the ground truth. The results demonstrate that integrating context can address and improve difficulties in COD arising from diverse appearances of camou-flaged objects, low boundary contrast, and inaccurate contours. Based on the evaluation metrics shown in Table 12, each model exhibits specific strengths. C2F-Net and BCNet generally achieve higher accuracy in detecting camouflaged objects with strong F-measure and S-measure scores, indicating robust performance in handling multi-scale and boundary-aware features. CamoFocus stands out for its computational efficiency while maintaining competitive accuracy, as seen in its balanced performance across datasets. PCPNet effectively prioritizes complex background pixels but has room for improvement in boundary precision. DiCANet excels in generating clear boundaries, reflected by its high E-measure scores, though its complexity may increase computational demands. Each approach highlights unique advancements in COD, addressing challenges such as multi-scale representation, boundary precision, and efficient feature refinement. In summary, CamoFocus performs best for the CAMO, COD10K, and NC4K datasets, while PCPNet has the best performance on the Chameleon dataset based on the sorted MAE values. Considering the scarcity of context-based camouflaged object detectors, it is reasonable to anticipate a greater number of scholarly articles exploring the use of context to improve object detection.

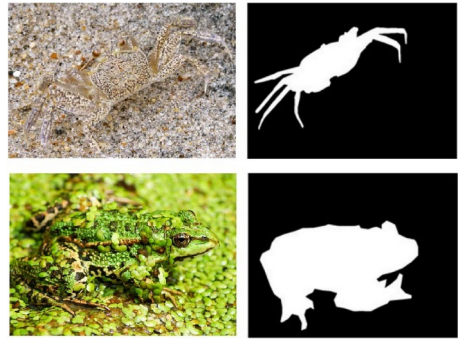

**Fig. 43** Detecting the frog and crab camouflaged in their environments is very challenging

## 5 Conclusion, research gaps, and limitations

### 5.1 Conclusion

In this systematic literature review, we comprehensively surveyed various aspects of context and noteworthy context-based object detection approaches within seven distinct categories, including general object detection, small object detection, video object detection, zero-shot object detection, one-shot object detection, few-shot object detection, and camouflaged object detection. In Sect. 2, We started by defining context aspects, including context in computer vision and human vision, context levels, contextual interactions, higher-order and pairwise relations, and a comprehensive analysis of context types. Then, after collecting papers from three databases and applying inclusion and exclusion criteria based on Sect. 3, in Sect. 4, we conducted a systematic analysis of 117 object detection papers that utilized context in their architectures to improve their performance. All 117 papers were reviewed and compared based on context level, context type, backbone, architecture, methods and modules, dataset, mAP, and other evaluation metrics. Most papers have focused on the following four types of context: spatial, size, semantic, and temporal, either individually or in

Table 12 Camouflaged object detection

| Detector name | Context type/level | Backbone/Architecture | Mechanism/Module | Dataset | MAE(M) | F-measure | S-measure | E-measure |
|---|---|---|---|---|---|---|---|---|
| **CamoFocus** Khan et al. (2024) | **Spatial/Local,Global** | **Res2Net50** | **FSM,CRM** | **CAMO** | **0.043** | **0.842** | **0.873** | **0.926** |
| PCPNet Song et al. (2023) | Spatial,Semantic/Local,Global | ConvNext-tiny | VCG,PIE,LCRM | CAMO | 0.053 | 0.799 | 0.840 | 0.913 |
| BCNet Xiao, Chen, et al. (2023) | Scale,Spatial/Local,Global | Res2Net-50 | HFEM,BFIM | CAMO | 0.068 | 0.761 | 0.830 | 0.886 |
| Model Wen et al. (2024) | Spatial,Semantic/Local,Global | Swin-T,EfficientNet-B7 | Masked-Edge and Joint Dense Skip Attention | CAMO | 0.069 | 0.727 | 0.802 | 0.865 |
| C2F-Net Sun et al. (2021) | Scale,Spatial/Local,Global | Res2Net-50 | MSCA,ACFM,DGCM | CAMO | 0.080 | 0.719 | 0.796 | 0.854 |
| DiCANet Ike et al. (2024) | Spatial,Semantic/Local,Global | ResNet-50 | ARB,CDB | CAMO | 0.091 | 0.647 | 0.747 | 0.828 |
| **PCPNet** Song et al. (2023) | **Spatial,Semantic/Local,Global** | **ConvNext-tiny** | **VCG,PIE,LCRM** | **Chameleon** | **0.021** | **0.866** | **0.902** | **0.958** |
| CamoFocus Khan et al. (2024) | Spatial/Local,Global | Res2Net50 | FSM,CRM | Chameleon | 0.023 | 0.876 | 0.912 | 0.957 |
| Model Wen et al. (2024) | Spatial,Semantic/Local,Global | Swin-T,EfficientNet-B7 | Masked-Edge and Joint Dense Skip Attention | Chameleon | 0.028 | 0.776 | 0.853 | 0.914 |
| BCNet Xiao, Chen, et al. (2023) | Scale,Spatial/Local,Global | Res2Net-50 | HFEM,BFIM | Chameleon | 0.029 | 0.839 | 0.901 | 0.944 |
| C2F-Net Sun et al. (2021) | Scale,Spatial/Local,Global | Res2Net-50 | MSCA,ACFM,DGCM | Chameleon | 0.032 | 0.828 | 0.888 | 0.935 |
| DiCANet Ike et al. (2024) | Spatial,Semantic/Local,Global | ResNet-50 | ARB,CDB | Chameleon | 0.034 | 0.805 | 0.871 | 0.950 |
| **CamoFocus** Khan et al. (2024) | **Spatial/Local,Global** | **Res2Net50** | **FSM,CRM** | **COD10K** | **0.021** | **0.802** | **0.873** | **0.935** |
| PCPNet Song et al. (2023) | Spatial,Semantic/Local,Global | ConvNext-tiny | VCG,PIE,LCRM | COD10K | 0.027 | 0.751 | 0.838 | 0.924 |

**Table 12** (continued)

| Detector name | Context type/level | Backbone/Architecture | Mechanism/Module | Dataset | MAE(M) | F-measure | S-measure | E-measure |
|---|---|---|---|---|---|---|---|---|
| Model Wen et al. (2024) | Spatial,Semantic/Local,Global | Swin-T,EfficientNet-B7 | Masked-Edge and Joint Dense Skip Attention | COD10K | 0.032 | 0.676 | 0.802 | 0.890 |
| BCNet Xiao, Chen, et al. (2023) | Scale,Spatial/Local,Global | Res2Net-50 | HFEM,BFIM | COD10K | 0.033 | 0.704 | 0.827 | 0.894 |
| C2F-Net Sun et al. (2021) | Scale,Spatial/Local,Global | Res2Net-50 | MSCA,ACFM,DGCM | COD10K | 0.036 | 0.686 | 0.813 | 0.890 |
| DiCANet Ike et al. (2024) | Spatial,Semantic/Local,Global | ResNet-50 | ARB,CDB | COD10K | 0.043 | 0.629 | 0.775 | 0.872 |
| **CamoFocus** Khan et al. (2024) | **Spatial/Local,Global** | **Res2Net50** | **FSM,CRM** | **NC4K** | **0.030** | **0.853** | **0.889** | **0.936** |
| PCPNet Song et al. (2023) | Spatial,Semantic/Local,Global | ConvNext-tiny | VCG,PIE,LCRM | NC4K | 0.038 | 0.816 | 0.861 | 0.925 |
| BCNet Xiao, Chen, et al. (2023) | Scale,Spatial/Local,Global | Res2Net-50 | HFEM,BFIM | NC4K | 0.043 | 0.788 | 0.857 | 0.910 |

combination, such as the spatio-temporal context. Moreover, the combination of local and global context has garnered more attention than their individual usage. In order to integrate contextual information, papers have proposed distinct modules and mechanisms that can be added into their architectures. In all seven reviewed categories, context has played a significant role in improving models performance.

In general object detection 4.2, different approaches, including graph-based approaches 4.2.1, hierarchical approaches 4.2.2, multi-scale approaches 4.2.4, RPN-based approaches 4.2.5, attention-based approaches 4.2.6, and other approaches 4.2.7, have integrated context into their architecture through various modules and mechanisms during the network training process. Instead of adding context into context-free architectures, context data augmentation methods 4.2.3 augment available contextual information.

In small object detection 4.3, various versions of the YOLO network have been used more frequently than other architectures. The combination of YOLO with contextual information has led to acceptable results in detecting small objects.

In video object detection 4.4, the investigated methods have effectively utilized context to improve their performance by tackling the problems such as low quality, background distractors, occlusions, motion blur, and degraded object appearance.

In zero-shot, one-shot, and few-shot object detection 4.5, where the number of training samples is very limited, networks attempt to focus on regions with a higher likelihood of presence through the integration of context. They also aim to detect unseen objects that have not been seen during the training process using global context and background information.

The methods presented in camouflaged object detection 4.6, tackle the difficulties in COD stemming from the diverse appearances of camouflaged objects, low boundary contrast, and inaccurate contours by integrating context into their architectures.

We answer the research questions outlined in the introduction Sect. 1 as follows:

- RQ1. Which context types have been predominantly used in different categories of object detection? Figs. 40, 41, and 42 show the distribution of context types and levels in general object detection, small object detection, and video object detection. Based on Fig. 11, more than 14 types of context can be utilized in object detection, but most articles have focused on size, spatial, temporal, and combinations of them. This preference can stem from the alignment of these context types with the unique challenges posed by specific detection categories; for instance, spatial and temporal context naturally aid in tracking movement in video detection or locating small objects within crowded scenes. Additionally, the frequent use of these contexts could reflect their well-established effectiveness in enhancing model performance. Furthermore, in all three categorizations, a combination of local and global context is more commonly utilized than each individually. Tables 11 and 12 also demonstrate a similar trend regarding the distribution of context in zero-shot, one-shot, few-shot, and camouflaged object detection. The limited exploration of other context types suggests potential for future research to investigate broader context types. This could offer new insights and solutions to complex scenarios where underutilized contexts, such as environmental or spectral data, might further enhance object detection accuracy and adaptability.
- RQ2. What approaches are applicable for integrating context in object detection? In general object detection, to integrate context into algorithms, methods have utilized different modules and mechanisms in various approaches such as graph-based approaches, hierarchical approaches, context augmentation, multi-scale approaches, RPN-based approaches, attention-based approaches, etc. Although the majority of papers integrate context into context-free methods, some approaches have been proposed to facilitate object detection through context data augmentation. In other categories, including small object detection, video object detection, zero-shot, one-shot, few-shot object detection, and camouflaged object detection, various proposed methods have also utilized different modules to integrate context into architectures. Given that methods have been tested on different datasets, it cannot be concluded which approach is the best. All approaches have been thoroughly reviewed in detail in Sect. 4.

- RQ3. Why are certain backbone networks and architectures most commonly used in recent context-based object detectors? Based on Figs. 40, 41, 42, and Tables 11 and 12, ResNet, VGG16, and Faster RCNN have been more commonly utilized in object detection approaches. This preference can be attributed to the proven effectiveness of these architectures in balancing computational efficiency with high performance, especially for complex contextual features. ResNet and VGG16, for instance, offer robust feature extraction capabilities, essential for capturing fine-grained context information. Faster R-CNN, a two-stage detector, is favored for its high accuracy and ability to integrate additional contextual modules, making it particularly suitable for scenarios requiring nuanced contextual understanding.

- RQ4. What are the best performing context-based methods on the most widely used datasets, including COCO and PASCAL VOC? What about for one-stage and two-stage object detectors? In the domain of general object detection, the FNM (Barnea and Ben-Shahar 2019) demonstrates superiority on the COCO dataset, while the Feature Refinement (Ma and Wang 2023) excels on VOC07, and the Cascade Region Proposal (Zhong et al. 2020) stands out as the most effective for VOC12. On the COCO dataset, Feature Refinement (Ma and Wang 2023) proves to be the best method for medium and large objects, while GCE (Peng et al. 2022) stands out as the best for small objects. In the realm of one-stage object detection, CCAGNet (Miao et al. 2022) emerges as the top performer on PASCAL07, whereas GCA RCNN (Zhang et al. 2021) demonstrates superiority on the COCO dataset.

- RQ5. To what extent can context improve object detection in scenarios where the number of training samples is very limited, such as in few-shot object detection, or when objects are indistinguishable from the background, as in camouflage object detection? The number of papers using context to enhance performance in zero-shot, one-shot, few-shot, and camouflaged object detection is much lower compared to other categories. However, the superior performance of the proposed approaches in reviewed papers over context-free methods shows that context can have a significant impact on training networks that are constrained by limited training data. Additionally, context can enhance the network's capability in detecting objects that are indiscernible in camouflaged object detection.

## 5.2 Research gaps

Based on the findings from the papers, we have identified and categorized the following research gaps related to context-based object detection:

- Many current object detection models demonstrate strong performance within specific domains; however, the generalization of context-based detectors across domains remains underexplored. For instance, approaches like SG-YOLO and VCANet have shown effective domain-specific adaptations for underwater object detection and small objects on roads but lack the ability to generalize across diverse domains. Future research could focus on evaluating and enhancing cross-domain robustness, potentially incorporating adaptive mechanisms like self-supervised learning or domain adaptation techniques to improve performance across varied contexts.
- Most models rely on static context modules that lack adaptability to various contexts within a single scene or across tasks. Modular, flexible context integration-where context types like spatial, temporal, or multi-modal information can be dynamically added-could enhance model versatility and improve performance in complex, real-world environments, particularly in situations with shifting contextual needs.
- Many models are optimized for specific object sizes, often excelling with either small or large objects but not both. Real-world scenes typically include a range of object sizes, which current models are not fully equipped to handle in a unified, efficient manner. Research into developing multi-scale, adaptive models that can dynamically detect objects across various scales in real-time could greatly benefit high-stakes applications like autonomous navigation.
- Video-based object detection for few-shot or camouflaged scenarios could benefit from utilizing long-range temporal context across frames. Long-term dependencies can provide cues to improve detection in challenging conditions, like rapid movement or occlusion. Integrating robust temporal context mechanisms, such as those in Context Faster R-CNN and TACF, could enhance model accuracy for tracking and detection in video sequences.
- Most studies focus on spatial, temporal, or semantic contexts, with limited exploration of other context types, such as environmental (weather or lighting) and behavioral cues (e.g., predicting actions based on object interactions). Further research into these novel contextual types, especially in applications like surveillance and autonomous vehicles, could offer new insights and improve model performance in complex scenarios.
- Current approaches that uses scale context in object detection often overlook depth as a factor influencing object size. This assumption of fixed size relationships can lead to inaccuracies in layered scenes where objects appear at different scales depending on their distance from the camera. Future research could address this by incorporating depth-aware scale context, allowing models to adapt object size based on depth cues. Such an approach would enhance robustness, particularly in complex environments with varying object distances.
- Although contextual information has enhanced the performance of zero-shot, one-shot, few-shot, and camouflaged object detectors, there is a scarcity of published studies on these topics. This lack could be remedied by devoting additional research efforts to these domains.
- More multimodal networks for integrating different types of context, such as text and audio, could be implemented to improve object detection performance, especially in complex scenes or under challenging conditions. Recently, large vision-language models (VLMs), such as those developed for multimodal applications, have shown potential in leveraging textual and visual information to enhance contextual understanding.

Exploring these VLMs for object detection tasks may offer new pathways to integrate richer contextual cues, potentially improving detection accuracy in scenarios where traditional methods struggle.
- A notable gap in available papers is the limited consideration of uncertainties and statistical rigor in evaluating object detection models. While performance metrics like mAP and APs are frequently reported, few studies include crucial statistical insights, such as error bars, or confidence intervals, to assess the significance and reliability of their results. Addressing this gap by incorporating uncertainty quantification and statistical testing would provide a clearer understanding of model robustness and result validity, enhancing the reliability of findings in object detection research.

### 5.3 Limitations

- While this review employed a boolean search criterion targeting the term "context" within titles, we recognize that this may not capture all relevant papers, especially those that address context implicitly. For instance, multi-modal approaches utilizing LiDAR, thermal, or text-based inputs, such as visual question answering systems, often integrate spatial, semantic, or other contextual information without explicitly mentioning "context." Additionally, multi-task learning frameworks can provide implicit context by sharing feature representations across tasks. Examples of such approaches that may not have been fully captured include TIDE for semantic contextual adaptation Kerssies et al. (2022), MMDetection3D for LiDAR-based context integration Contributors (2020), and MTLNet for multi-task learning in complex environments He et al. (2019). Future reviews could expand search criteria to better capture these implicit uses of context in object detection research.
- This review highlights the impact of context on improving object detection performance across various categories. However, a limitation is the lack of a direct comparison between context-based and context-free approaches in the studies analyzed. Most reviewed papers do not present baseline results for context-free models, making it difficult to quantify the precise contribution of context in isolation. Addressing this gap could provide a clearer understanding of context's value and limitations. Future research would benefit from systematic studies that evaluate and report on the performance differences between context-inclusive and context-free approaches across object detection categories.
- We have concentrated on seven categories of context-based object detection, including general object detection, video object detection, small object detection, camouflaged object detection, zero-shot, one-shot, and few-shot object detection. Other approaches, such as salient object detection, RGB-D object detection, and those listed in the exclusion section of Table 2, could be explored as future research directions, as they have not been extensively covered thus far and hold significant potential for further exploration.

**Funding** Open access funding provided by Malmö University.

**Data availability** No datasets were generated or analysed during the current study.

## Declarations

**Competing interests** The authors declare no competing interests.

**Publisher's Note**  Springer Nature remains neutral with regard to jurisdictional claims in published maps and institutional affiliations.

## Authors and Affiliations

**Mahtab Jamali[1] · Paul Davidsson[1] · Reza Khoshkangini[1] · Martin Georg Ljungqvist[2] · Radu-Casian Mihailescu[1]**

✉ Mahtab Jamali

mahtab.jamali@mau.se

Paul Davidsson
paul.davidsson@mau.se

Reza Khoshkangini
reza.khoshkangini@mau.se

Martin Georg Ljungqvist
martin.ljungqvist@axis.com

Radu-Casian Mihailescu
radu.c.mihailescu@mau.se

[1] Department of Computer Science and Media Technology, Sustainable Digitalisation Research Centre, Malmö University, Malmö, Sweden

[2] Axis Communications AB, Lund, Sweden